\documentclass[10pt]{article} 

\usepackage[utf8]{inputenc}
\usepackage{amsmath}
\usepackage{amsfonts}
\usepackage{amssymb}
\usepackage{listings}
\usepackage{mathrsfs}
\usepackage{times}
\usepackage{float}
\usepackage{color}
\usepackage{setspace}
\usepackage{color,soul}

\usepackage{amsmath,amsfonts,amsthm,eucal}
\usepackage{enumerate}
\usepackage[letterpaper,margin=3cm]{geometry}
\usepackage{float}
\usepackage[title]{appendix}
\usepackage{color}
\usepackage{tabularx}
\usepackage{siunitx}
\usepackage[tableposition=below]{caption}
\usepackage[
pdfstartview=XYZ,
bookmarks=true,
colorlinks=true,
linkcolor=blue,
urlcolor=blue,
citecolor=blue,
pdftex,
bookmarks=true,
linktocpage=true,   
hyperindex=true
]{hyperref}

\usepackage{lipsum}
\usepackage[FIGBOTCAP,TABTOPCAP,bf,tight]{subfigure}

\usepackage{natbib}
\usepackage[pdftex]{graphicx}
\usepackage{epstopdf}
\usepackage{booktabs} 
\usepackage{tikz,mathpazo}
\usetikzlibrary{shapes.geometric, arrows}

\usepackage{caption}

\newcommand{\tensor}[1]{\ensuremath{\boldsymbol{#1}}}
\newcommand{\set}[1]{\ensuremath{\mathbb{#1}}}
\newcommand{\tuple}[1]{\ensuremath{\mathbb{#1}}}

\theoremstyle{remark}

\renewcommand{\vec}[1]{\ensuremath{\boldsymbol{#1}}}
\newcommand*\circled[1]{\tikz[baseline=(char.base)]{
		\node[shape=circle,draw,inner sep=2pt] (char) {#1};}}

\usepackage{algorithm}
\usepackage{algorithmicx}
\usepackage[noend]{algpseudocode}

\theoremstyle{definition}
\newtheorem{definition}{Definition}

\usepackage{longtable}
\usepackage{adjustbox}
\usepackage{framed}

\title{A cooperative game for automated learning of elasto-plasticity knowledge graphs and models with AI-guided experimentation} 

\begin{document}


\author{Kun Wang\thanks{Department of Civil Engineering and Engineering Mechanics, 
 Columbia University, 
 New York, NY 10027.     \textit{kw2534@columbia.edu}  }       \and
        WaiChing Sun\thanks{Department of Civil Engineering and Engineering Mechanics, 
 Columbia University, 
 New York, NY 10027.
  \textit{wsun@columbia.edu}  (corresponding author)      
}
\and
Qiang Du\thanks{Department of Applied Physics and Applied Mathematics, and Data Science Institute, Columbia University, New York, NY 10027.
  \textit{qd2125@columbia.edu}}
}

\maketitle

\begin{abstract}
We introduce a multi-agent meta-modeling game to generate data, knowledge, and models that make predictions on constitutive responses of elasto-plastic materials. We introduce a new concept from graph theory where a modeler agent is tasked with evaluating all the modeling options recast as a directed multigraph and find the optimal path that links the source of the directed graph (e.g. strain history) to the target (e.g. stress) measured by an objective function. Meanwhile, the data agent, which is tasked with generating data from real or virtual experiments (e.g. molecular dynamics, discrete element simulations), interacts with the modeling agent sequentially and uses reinforcement learning to design new experiments to optimize the prediction capacity. Consequently, this treatment enables us to emulate an idealized scientific collaboration as selections of the optimal choices in a decision tree search done automatically via deep reinforcement learning.  
\end{abstract}

\section{Introduction}
\label{intro}
In single-physics solid mechanics problems, the balance of linear momentum is often used to provide constraints 
for the motion of a body in the space-time continuum, while a constitutive law is often supplied to replicate constitutive 
responses at a selected material point of the body. 
Many successful commercial and open-source codes now introduce mechanisms or gateways that simplify 
the incorporation of material 
point constitutive models into predefined solid mechanics solvers (e.g. UMAT in ABAQUS)
\citep{hibbitt2001abaqus, rutqvist2011implementation, sun2013stabilized, sun2015stabilized, 
salinger2016albany, wollny2017hierarchical, na2018computational, choo2018cracking, choo2018coupled}. 
Once a constitutive law is formulated, algorithms are then 
designed to approximate the mathematical model such that a computer can be used to 
run simulations. The algorithms that approximate or enforce the constitutive laws are then 
verified, validated and eventually used in engineering practice \citep{kirchdoerfer2016data, ibanez2018manifold, wang2018multiscale}. 

Conventionally, a constitutive model that replicates the relation between the kinetic 
 and kinematics quantities is derived from a finite set of fundamental principles, assumptions and 
 phenomenological equations \citep{truesdell1959rational, truesdell2004non}.  
For instance, the laws of thermodynamics, material frame indifference, and balance laws are 
universal principles that are widely believed to be 
true for all materials under common circumstances.  
After enforcing those universal 
principles, there often remains a finite set of choices a modeler can make to construct a constitutive model. 
In particular, different types of 
experiments are designed such that a proper set of additional constraints can be generated. 
These constraints may not be fully explained by universal principles but are added to ensure 
the compatibility between observed and simulated mechanical responses. 
In reality, the universal principles alone are insufficient to complete most of the constitutive laws, 
regardless of the spatial scales they are designed for. 
As a result, phenomenological relations are introduced such that all the constraints imposed by
principles and observations can be enforced. 

\subsection{Rationales of phenomenological relations} 
Even though phenomenological relations cannot be fully justified via universal-principle arguments, 
it is understandable that proposers of these phenomenological relations often seek justifications by 
introducing new theories or incorporating microstructural information as constraints. 
For instance, the most commonly used family of soil models, the critical-state plasticity, relies on the existence  
of a critical state line in the state path (i.e. the plot of specific volume against the natural logarithmic of 
effective mean pressure) such that soil in the numerical simulations may develop plastic shear strain without volumetric deformation 
when reaching the critical state and exhibit the plastic dilatancy or contraction at different void ratio and 
overconsolidation ratio
 \citep{schofield1968critical, casagrande1976liquefaction, been1991critical, ling2006unified, 
 sun2013unified, na2017computational}. 
Experimental evidences are then sought to either
justify the claim (cf. \citet{wood1990soil}), or redefine the applicability of the theory in light of 
new evidences (c.f. \citet{mooney1998unique, li2011anisotropic, zhao2013unique}).
The incorporation of fabric tensors in critical state plasticity is another example where 
sub-scale information is incorporated to enhance forward prediction quality \citep{li2011anisotropic, wang2016semi}. 
Other types of information incorporated into the constitutive law  may come from microstructural 
attributes or the kinematics of microstructures. 
 A classical example is 
crystal plasticity where the kinematics of the plastic flow is restricted by the orientations of the 
slip systems \citep{asaro1983crystal, miehe2001comparative, borja2013plasticity, na2018computational}.

Finally, for practical reasons or due to lack of sufficient experimental evidences to prove otherwise, 
assumptions are sometimes made to interpret a phenomenological relation. 
A classical example for this type of phenomenological approach is the effective displacement theory commonly used in traction-separation models where 
one assumes that a scalar kinematic measure, often a weighted norm of normal and tangential 
displacements, can be used to determine a scalar traction measure that leads to the traction vector 
\citep{ortiz1999finite, park2011cohesive}. 
Nevertheless, the distinction between phenomenology that only enhances curve-fitting in calibration 
and the counterpart that leads to more accurate, robust and reliable predictions
is often a blurred line and might be subject to debate \citep{truesdell2004non, wang2016identifying, wang2019updated, wang2019meta}. 
Furthermore, the popularity of a model in the short term 
is also not necessarily purely based on the prediction quality,
but also ties to the difficulty in calibration and interpretation of the model \citep{lange2012makes}, the demand of experimental data \citep{wang2016identifying, olivier2018marginalized}, 
as well as the social, cultural and personal influences 
(cf. \citet{malmgren2010role}), among other factors . 
In the case where a limited subset of data might be chosen to
make a constitutive law or theory sound plausible or consistent with a physical phenomenon, 
the true forward prediction quality of the model might take a toll while the apparent capacity of the model could be exaggerated \citet{munafo2017manifesto}.
The underlying problem is that this issue is very difficult to detect unless all the models are compared objectively 
in the same benchmark study and subjected to a universally agreed validation metric \citep{boyce2014sandia, pack2014sandia}. Hence, a validation procedure that employs blind predictions is critical, \textit{regardless of the type of models used for predictions}. 

\subsection{Data-driven approaches as alternatives}
An alternative to the conventional modeling approach is the data-driven modeling in which constitutive responses are
predicted primarily based on the available data either by black-box neural networks \citep{furukawa1998implicit, ghaboussi1998autoprogressive, lefik2003artificial, wang2017data, wang2018multiscale} or via minimization problems in the phase space \citep{kirchdoerfer2016data}. 
While the latter approach, as outlined in \citet{kirchdoerfer2016data} and \citet{kirchdoerfer2017data},
has shown great promises to 
handle hyperelasticity problems, the extension to plasticity problems
 likely requires either imposing further constraints (e.g. perfect plasticity
\citet{ibanez2018manifold}) or creating a sufficiently large database 
to capture the phase space of the history-dependent responses. 
On the other hand, \citet{lefik2003artificial} has demonstrated that a neural network can generate 
cyclic elasto-plastic responses with some level of success. 
Nevertheless, despite the fact that a multi-layer neural network can be considered as a universal approximator, as pointed out in \citet{hornik1989multilayer}, this does not imply that the training of the neural network is always successful. In fact, 
failure to complete the training is quite common and it might be caused by, for example, 
(1) higher demand of data for the neural network training compared to the material parameter identification in conventional modeling, 
(2) the curse of high dimensionality that leads to inconsistency between calibration and forward prediction performance, 
(3) issues related to under-fitting and over-fitting, and (4) the vanishing gradient issues that make the algorithm 
unable to locate the global minimizer of the loss function \citep{wang2018multiscale}. 
Furthermore, without special treatment to extend the database for training 
the neural network, the resultant models 
often exhibit dependence on coordinate systems. Even though this issue has been addressed recently using the spectral decomposition of tensorial inputs and outputs in recurrent neural networks \citep{wang2018multiscale}, 
this lack of consistency with theory indicates that the domain expertise remains critical for evaluating the quality of the machine learning model and finding remedies 
for issues not immediately apparent for nonspecialists. 

In the aforementioned data-driven approach, the demand for big data remains an ongoing challenge 
\citep{smith2016linking, tang2018virtual, liu2018microstructural}. In particular, 
 machine learning models, especially those in most generic forms (i.e. model-free approaches), may suffer a lack of constraints imposed by material theory, thus increasing the demand for data to generate the constraints.  
Hence, it is important for modelers to be able to estimate the least amount of data 
required to complete the training of a specific model. The introduction of the two-player cooperative game in this paper can provide a practical solution to find the required amount and type of data for path-dependent materials. 


%
%
%
%
%
%
\subsection{The hybridized theoretical/data-driven approach} 
In this paper, our goal is to (1) introduce a meta-modeling method to generate algorithms that \textbf{hybridize} theory, phenomenological relations, 
and universal principles to \textbf{automatically} generate constitutive laws to fulfill a specific objective defined by the loss (objective function)
in a quantitatively optimal manner and (2) incorporate the reinforcement learning technique to select experiments that lead to improvement in prediction capacity. 
We do not limit ourselves to the approach in which the neural network model is either used to replace the entire constitutive law or not being used at all (cf. \citep{ghaboussi1991knowledge, kirchdoerfer2016data, wang2018multiscale}). 
Instead, our goal is to find the optimal way out of all the possible choices to construct a constitutive law for a given material data. 

To reach our goal, we employ 
two techniques of discrete mathematics that are less commonly used in computational mechanics, the directed multigraph and decision tree learning. 
First, the directed multigraph is used to recast the available choices of constitutive laws as 
a family of possible ways to configure a graph of information flow from the upstream (the source or input, such as the relative displacement or strain) 
to the downstream (the target or output, such as the traction or stress). A model is a path (in the terminology of graph theory) of this 
directed multigraph that optimizes an objective function. 
As such, a model is associated with a collection of physical quantities (vertices in the directed graph) linked by either mathematical expressions or machine learning models that connect the upstream to the downstream (edges in the directed graph) (cf. \citet{wang2018multiscale}). 

Within our framework, a black-box neural networks model, for instance, 
is simply a model in which there are no human-interpretable quantities connecting the input and output. Many classical neural network models such as \citet{ghaboussi1991knowledge, lefik2003artificial} and 
 \citet{wang2018multiscale} all belong to this category, as neurons are the only media that propagate the information flow. 
Meanwhile, a classical theory-based constitutive law can be viewed as a directed graph (or a particular path of the directed multigraph) in which all the edges are mappings that can be written as mathematical expressions formulated by human. 
On the other hand, a hybridized model could have a subset of neural network edges while having the rest edges theoretically based. 


Since the optimal configuration of the directed graph for a given problem and the corresponding 
objective function is not known a priori, we introduce mechanisms to hierarchically explore the possible modeling choices using a decision tree. 
A decision tree is simply an explicit representation of all possible scenarios such that the sequence of decisions 
(in our case the modeling choices and data explorations) is evaluated by an agent who then takes account of the possible 
observations (e.g. experimental observations), and state changes (e.g. the changes of validation metrics or loss function values)
to estimate the best choices. 

In this work, our major contributions are threefold. First, we introduce the concept of directed multigraph to enable the hybridization of theory-based and data-driven models to yield optimal forward predictions. Second, we create a model to emulate the process of formulating constitutive laws as an optimization problem for modeling choices, rather than performances. This treatment gives us hierarchical information that helps understand the causal connection among events and mechanisms. 
The importance of the usage of multigraph is that it enables us to form complex idea, knowledge, prediction, inference and 
response with a rather small set of simple elements. 
This kind of application of the principle of combinatorial generalization 
has long been regarded as the key signature of intelligence 
\citet{chomsky1965aspects,humboldt1999language, battaglia2018relational}.
Third, we also introduce a cooperative mechanism to integrate the data exploration into the modeling process. 
In this way, the framework can not only generate constitutive models to make the best predictions among the limited data, but also estimate the most efficient way to select experiments such that the most needed information is included to generate the knowledge closure. 

\subsection{Content organization}
The rest of the paper is organized as follows. We first introduce the meta-modeling cooperative game, including the method to recast a model as directed multigraph, and the generation of decision tree (Section \ref{sec:metamodeling}). 
Following this, we will introduce the detailed design of the data collection/meta-modeling game for modeling the collaboration of the AI data agent and the AI modeler agent (Section \ref{sec:game}). 
In Section \ref{sec:drl_algorithm}, we then review the multi-agent reinforcement learning algorithms that enable us to find the optimal decision for constitutive models, as well as the corresponding optimal actions the data agent 
takes to maximize the prediction quality of the AI-generated model. 
We then present numerical experiments to assess the accuracy and robustness of the blind predictions of the model generated via our meta-modeling algorithm operated on the directed multigraph. 
To check whether our approach is able to deal with a wide spectrum of situations and can be generalized for different materials, the multigraph meta-modeling algorithm is tested with distinctive types of data (e.g. synthetic data from elasto-plastic models and discrete element simulations). To aid the reproducibility of our numerical experiments by the third party, these data will be open source upon the publication of this article. 

\subsection{Notations and terminologies}
For convenience, we  provide a minimal review of the essential terminologies and concepts from graph theory that are used throughout this paper. Their definitions can be found in, for instance, \citet{graham1989concrete,
west2001introduction, bang2008digraphs}.

\begin{definition}
A \textbf{$\vec{n}$-tuple} is a sequence or ordered list that consists of $n$ element where $n$ is a non-negative integer and that (unlike a set) may contain multiple instances of the same element. 
\label{def:tuple}
\end{definition}

\begin{definition}
A \textbf{directed graph} (digraph) is an ordered pair (2-tuple) $\tuple{G}=(\set{V},\set{E})$ where $\set{V}$ is a nonempty set of vertices and $\set{E}$ is a \textit{set} of \textit{ordered} pairs of vertices (directed edges) where each edge in $\set{E}$ connects a pair of source (beginning) and target (end)  vertices in a specific direction. Both vertices connected by an edge in $\set{E}$
must be elements of $\set{V}$ and the edge connecting them must be unique. 
\label{def:digraph}
\end{definition}

\begin{definition}
	A \textbf{directed acyclic graph} is a directed graph where edges do not form any directed cycle. In a directed acyclic graph, there is no path that can start from a vertex and eventually loop back to the same vertex.  
\end{definition}

\begin{definition}
	A \textbf{directed multigraph} with a distinctive edge identity (also called multi digraph) 
	is an ordered 4-tuple $\tuple{G}=(\set{V},\set{E},\vec{s}, \vec{t})$ where $\set{V}$ is a set of vertices, $\set{E}$ is a set of edges that connect source and target vertices, $\vec{s}: \set{E} \rightarrow \set{V}$ is a mapping that maps each edge to its source node, and  $\vec{t}:\set{E} \rightarrow \set{V}$ is a mapping that maps each edge to its target node.
	\label{def:dimgraph}
\end{definition}

\begin{definition}
	An \textbf{underlying graph} $\tuple{U}$ of a directed multigraph $\tuple{G}$ is a multigraph whose edges are without directions.   
	\label{def:underlyinggraph}
\end{definition}

\begin{definition}
A \textbf{subgraph} 
$\tuple{G}'$ of a directed multigraph $\tuple{G}$ is a directed multigraph whose vertex set $\set{V}'$ is a subset of $\set{V}$ ( $\set{V}' \subseteq \set{V}$), and whose edge set $\set{E}'$ is a subset of $\set{E}$ ( $\set{E}' \subseteq \set{E}$). 
\label{def:subgraph}
\end{definition}

\begin{definition}
	A \textbf{labeled directed multigraph} is a directed multigraph with labeled vertices and edges which can be mathematically
	expressed as an 8-tuple $\tuple{G}=(\set{L_{V}}, \set{L_{E}}, \set{V},\set{E},\vec{s},\vec{t}, \vec{n_{V}}, \vec{n_{E}})$ where
	$\set{V}$ and $\set{E}$ are the sets of vertices and edges, $\set{L_{V}}$ and $\set{L_{E}}$ are the sets 
	of labels for the vertices and edges, $s:\set{E} \rightarrow \set{V}$ and $t:\set{E} \rightarrow \set{V}$ are the mappings that map the edges to the source and target vetrices, $n_{V}:\set{V} \rightarrow \set{L_{V}}$ and $n_{E}: \set{E} \rightarrow \set{L_{E}}$ are the mappings that give the vertices and edges the corresponding labels in $\set{L_{V}}$ and $\set{L_{E}}$ accordingly. 
	\label{def:lmdigraph}
\end{definition}


As for notations and symbols, bold-faced letters
denote tensors (including vectors which are rank-one tensors); 
the symbol '$\cdot$' denotes a single contraction of adjacent indices of two tensors (e.g. $\vec{a} \cdot \vec{b} = a_{i}b_{i}$ or $\tensor{c}
\cdot \vec{d} = c_{ij}d_{jk}$ ); the symbol `:' denotes a double
contraction of adjacent indices of tensor of rank two or higher (
e.g. $\tensor{C} : \vec{\epsilon^{e}}$ = $C_{ijkl} \epsilon_{kl}^{e}$
); the symbol `$\otimes$' denotes a juxtaposition of two vectors
(e.g. $\vec{a} \otimes \vec{b} = a_{i}b_{j}$) or two symmetric second
order tensors (e.g. $(\vec{\alpha} \otimes \vec{\beta})_{ijkl} =
\alpha_{ij}\beta_{kl}$). Moreover, $(\tensor{\alpha}\oplus\tensor{\beta})_{ijkl} = \alpha_{jl} \beta_{ik}$ and $(\tensor{\alpha}\ominus\tensor{\beta})_{ijkl} = \alpha_{il} \beta_{jk}$. We also define identity tensors $(\tensor{I})_{ij} = \delta_{ij}$, $(\tensor{I}^4)_{ijkl} = \delta_{ik}\delta_{jl}$, and $(\tensor{I}^4_{\text{sym}})_{ijkl} = \frac{1}{2} (\delta_{ik}\delta_{jl} + \delta_{il}\delta_{kj})$, where $\delta_{ij}$ is the Kronecker delta. As for sign conventions, unless specified otherwise,
we consider the direction of the tensile stress and dilative pressure as positive.

\section{Meta-modeling: deriving material laws from a directed multigraph}  \label{sec:metamodeling}
In this section, we describe the concepts behind the proposed automated meta-modeling procedure and the mechanism 
of the learning process. The key departures of our newly proposed method via the neural network approaches 
for constitutive laws
 (e.g.  \citet{ghaboussi1991knowledge, ghaboussi1998autoprogressive, lefik2002artificial, lefik2003artificial, wang2018multiscale}) 
is the introduction of labeled directed multigraph that represents all possible theories under consideration for modeling a physical process, the acyclic directed graph that represents the most plausible knowledge on 
the relationships among physical quantities, and the data agent which enables users to estimate the 
amount of data required to reach the point where additional information no longer enhances prediction capacity for a given action space.
In this paper, our focus is limited to the class of materials that exhibits elasto-plastic responses while damage can be neglected. We assume that the deformation is infinitesimal and the material is under isothermal condition. The proposed methodology, however, can be extended to other more complex materials.

\subsection{Material modeling algorithm as a directed multigraph}
The architecture of an algorithm is often considered as a directed multigraph \citep{dkabrowski2011software}. 
In essence, a material model can be thought as a procedure that employs organized knowledge
to make predictions such that relationships of components and the universally accepted principles governs the 
outcomes of predictions. For instance, we may consider a traction-separation model as an information flow in a directed graph where 
physical attributes, such as porosity, plastic flow, permeability, are considered as vertices 
and their relationships are considered as edges \citep{wang2018multiscale}. The input and output of the models 
(e.g. relative displacement history and traction) are then considered as the sources and targets of the directed graph. 

However, in some circumstances, a physical relation can be modeled by more than one methods, 
theories or constitutive relations. To reflect the availability of options, a generalized representation of 
the thought process is needed when we try to use artificial intelligence algorithm to replace human
to write constitute models. 
This generalized thought process,  which we refer as metal-modeling (i.e.  modeling the process of writing a model),
can be recast as a labeled directed multigraph. The latter can be used where a pair of connected
vertices are not necessarily connected by one edge but by multiple edges, 
each represents a specific model that connects two physical quantities (e.g. porosity-permeability relationship). 
A formal statement can be written as follows:

\begin{framed}
\textbf{Possible configurations of constitutive laws as a labeled directed multigraph}. Given a data set which measures a set of  physical quantities defined as $\set{V}$ with a corresponding set of labels $\set{L_{V}}$ where $n_{\set{V}}:\set{V} \rightarrow \set{L_{E}}$ is a bijective mapping that maps the vertices to the labels. 
Let $\set{V}_{R} \subset \set{V}$  and $\set{V}_{L} \subset \set{V}$ be the source(s) and target(s) of the directed multigraph. 
All possible ways to write constitutive laws that map the input $V_{R}$ (e.g. strain history)  to output $V_{L}$ (e.g. stress) 
as information flow can be defined by the sets of directed edges where each edge that links two physical quantities $\set{E}$, the mappings $\vec{s}:\set{E} \rightarrow  \set{V}$ and $\vec{t} : \set{E} \rightarrow \set{V}$
that provide the direction of the information flow, and the surjective mapping $\vec{n}_{\set{E}}:\set{E} \rightarrow \set{L_{E}}$ that assigns the edge labels (names) to the edges. 
\label{fr:lawoptions}
\end{framed}

\begin{figure}[h!]\center
		\includegraphics[width=0.9\textwidth]{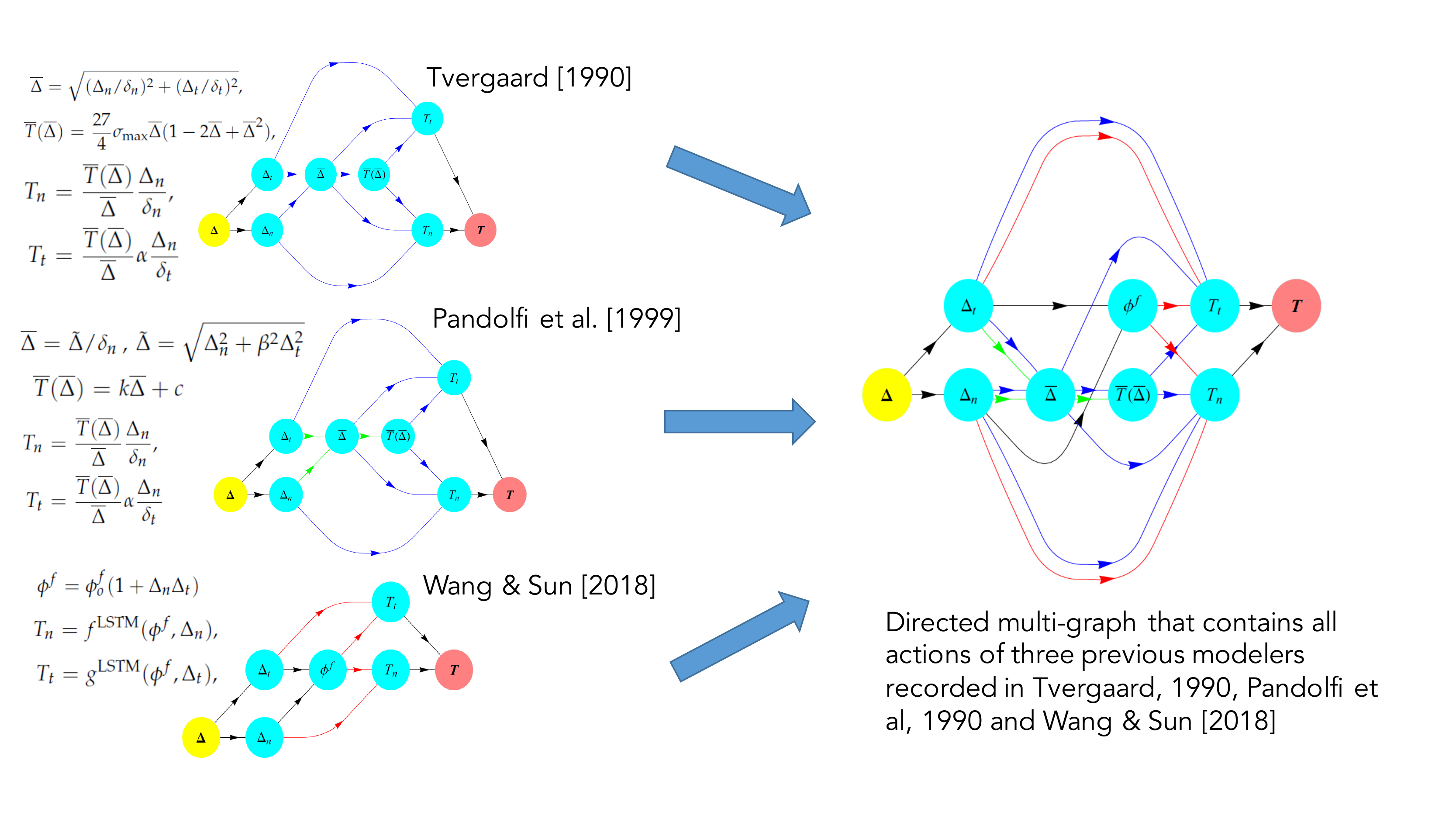}
	\caption{The generation of directed multi-graph by expanding action space using previous models. }
	\label{fig:multigraph}
\end{figure}

\begin{proof}[Example 1] Traction-separation Law. 
Given a pre-defined objective function, assume that the only known theoretical traction-separation model incorporated in the labeled directed multigraph 
are the Tvergaard model  (cf.\citet{tvergaard1990effect}) and the Ortiz-Pandolfi model (cf. \citet{pandolfi1999finite}).
In addition, we also consider using a neural network that incorporates porosity to predict traction-separation relations. 
Define the labeled directed multi-graph that provides all the options available. \\

First, we convert the traction-separation laws into directed graphs where the relative displacement vector is the input 
and the traction is the output. Notice that both \citet{tvergaard1990effect} and \citet{pandolfi1999finite} are effective displacement models where an effective displacement $\overline{\Delta}$ is used as additional input to predict the traction. 
In \citet{tvergaard1990effect}, 
\begin{eqnarray}
T_{n} &=& \frac{\overline{T}(\overline{\Delta})}{\overline{\Delta}}\frac{\Delta_{n}}{\delta_{n}},  \label{eq:traction1} \\
T_{t}  &=& \frac{\overline{T}(\overline{\Delta})}{\overline{\Delta}} \alpha \frac{\Delta_{n}}{\delta_{t}} 
\label{eq:traction2}
\end{eqnarray}
and the effective displacement and effective traction are scalars defined as, 
\begin{eqnarray}
\overline{\Delta} &=& \sqrt{ (\Delta_{n}/ \delta_{n})^{2} + (\Delta_{t} / \delta_{t})^{2}}, \label{eq:traction3} \\
\overline{T}(\overline{\Delta}) &=& \frac{27}{4} \sigma_{\max} \overline{\Delta}(1 - 2 \overline{\Delta} + \overline{\Delta}^{2}),
\label{eq:traction4} 
\end{eqnarray}
where $\delta_{n}$ and $\delta_{t}$ are the characteristic length corresponding to the fracture energy and cohesive strength of the normal and tangential opening modes,  $\alpha$ is a non-dimensional material parameter.  
As pointed out in \cite{park2011cohesive}, the traction-separation model in \citet{pandolfi1999finite} can be expressed in the forms of Eq. \eqref{eq:traction1} and \eqref{eq:traction2} with the alternative definition of effective displacement and traction separation law, i.e., 
\begin{eqnarray}
\overline{\Delta} &=& \tilde{\Delta}  / \delta_{n}   \: , \: \tilde{\Delta} =  \sqrt{ \Delta_{n}^{2} + \beta^{2} \Delta_{t}^{2}}  \label{eq:traction5} \\
\overline{T}(\overline{\Delta})  &=&  k \overline{\Delta} + c  \label{eq:traction6}
\end{eqnarray}
where $k$ is typically negative and $c$ is the effective cohesive strength. Fully we consider a neural network model in which the traction depends on the porosity $\phi^{f}$ \citep{coussy2004poromechanics, sun2013multiscale, wang2016semi}, i.e., 
\begin{eqnarray}
T_{n} &=&  f^{\text{LSTM}}(\phi^{f}, \Delta_{n}), \label{eq:traction7} \\ 
T_{t} &=&  g^{\text{LSTM}}(\phi^{f},  \Delta_{t}),  \label{eq:traction8}
\end{eqnarray}
where the exact expression of the function $f^{\text{LSTM}}$ and $g^{\text{LSTM}}$ are determined by adjusting the weight of the neurons in the recurrent neural network \citep{koeppe2017neural, wang2018multiscale}. Assuming that the solid constituent is incompressible, the porosity reads, 
\begin{equation}
\phi^{f} = \phi^{f}_{o} (1 + \Delta_{n} \Delta_{t})
\label{eq:porosity}
\end{equation}
The multi-graph that combines all the possible choices of the three traction separation laws can therefore be defined by multi-graph statement with the following sets, 
\begin{eqnarray}
\set{V} &=& \{\Delta_{n}, \Delta_{t}, T_{n}, T_{t}, \overline{\Delta}, \overline{T}, \phi^{f}\}  \label{eq:V} \\
\set{E} &=& \set{E}_{1} \cup \set{E}_{2} \cup  \set{E}_{3} \label{eq:E} \\
\set{E}_{1} &=& \{ \Delta_{n} \rightarrow \overline{\Delta}, \Delta_{t} \rightarrow \overline{\Delta}, 
\Delta_{n} \rightarrow \phi^{f}, \Delta_{t} \rightarrow \phi^{f}, 
\Delta_{n} \rightarrow T_{n}, \Delta_{t} \rightarrow T_{t}
\} \label{eq:E1} \\
\set{E}_{2} &=& \{ \overline{\Delta} \rightarrow \overline{T}, 
  \phi^{f} \rightarrow T_{n}, \phi^{f} \rightarrow T_{t},
\Delta_{n} \rightarrow T_{n}
\} \\
\set{E}_{3} &=& \{ 
\overline{T} \rightarrow T_{n}, \overline{T} \rightarrow T_{t}
\} \\
\set{L_{V}} &=& \{\text{normal disp., tan. disp., normal traction, tan. traction, eff. disp., eff. traction, porosity}  \}  \label{eq:LV} \\ 
\set{L_{E}} &=& \{ \text{Eq.} \eqref{eq:traction1},  \text{Eq.} \eqref{eq:traction2}, \text{Eq.} \eqref{eq:traction3}, \text{Eq.} \eqref{eq:traction4}, \text{Eq.} \eqref{eq:traction5}, \text{Eq.} \eqref{eq:traction6}, \text{Eq.} \eqref{eq:traction7}, \text{Eq.} \eqref{eq:traction8}, \text{Eq.} \eqref{eq:porosity} \}  \label{eq:LE}
\end{eqnarray}

Since $\vec{n}_{\set{V}}$ is a bijective mapping, the labeling of the vertices is trivial. The rest of the mappings, i.e. $\vec{s}$, $\vec{t}$ and $\vec{n}_{\set{E}}$ can be visualized in a labeled directed multigraph as shown in Figure \ref{fig:multigraph}. Essentially, the process of creating the directed multigraph is to mathematically represent all the possible options modelers can have when they are tasked to create a constitutive model for a data set. 
\end{proof}


\subsection{Recasting the process of writing constitutive laws as selecting subgraphs in a directed multigraph}
In the first meta-modeling game introduced in this work, we consider a scenario where a set of experimental data 
is given. This experimental data include measurement of different physical quantities, but the inherent relationships 
are unknown to the modeler. 
Furthermore, in the process of writing the constitutive law, the modeler must follow a set of rules coined as universal principles 
(e.g. thermodynamic principles, material frame indifference) \citep{kirchdoerfer2016data, wang2018multiscale}.  
Here, we first assume that an objective of writing the constitutive model is well defined and hence a score system is available for the deep Q-learning.  
We then idealize the process of writing a constitutive law \textit{with a fixed set of data} as a two-step process. 
 First, we consider all the possible ways to write a constitutive law and represent all these possibilities in a labeled directed multigraph. This labeled directed multigraph define the action space of the meta-modeling game.  Second, among all the possible ways to write a constitutive law, i.e., on the labeled directed multigraph, we seek the optimal configuration that will lead to the best outcome measured by an objective function. If the total number of possible configurations is sufficiently small, then the optimal configuration can be sought by building all the possible configurations and comparing their performance afterward. However, this brute force approach becomes infeasible when the total number of configurations is very large as in the case of the game of chess and Go \citep{silver2017mastering, silver2017masteringb}. 
As a result, the proposed 
procedure of finding the optimal configuration of a constitutive law is given as follow. 

\begin{framed}
Instants of constitutive laws are considered as directed graphs.
Given a dataset that contains the time history of measurable physical quantities 
of $n$ types of data stored in the vertices 
labeled by the vertex label $l_{i} \in \set{L_{V}}$ 
 and the labeled direct graph defined by the 8-tuple $\tuple{G}=(\set{L_{V}}, \set{L_{E}}, \set{V},\set{E},\vec{s},\vec{t}, \vec{n_{V}}, \vec{n_{E}})$, and objective function SCORE and constraints to enforce universal principles. 
Find an subgraph $\tuple{G}'$ of $\tuple{G}$ consists of vertices $\vec{V} \in \set{V}^s \subseteq \set{V}$ and edges 
$\vec{E} \in \set{E}^{s} \subseteq \set{E}$ such that 1) $\tuple{G}'$ is a directed acyclic graph, 2) a score metric is maximized under a set of $m$ constraints $f_{i}(l_{1}, l_{2},\ldots, l_{n}) =0, i=1,\ldots, m$ where , i.e., 
\begin{equation}
\begin{aligned}
& \underset{l_{i}}{\text{maximize}}
& & \text{SCORE}(l_{1}, l_{2},\ldots, l_{n}) \\
& \text{subject to}
& & f_i(i_{i}) = 0, \; i = 1, \ldots, m.
\end{aligned}
\end{equation}
\end{framed}

\begin{proof}[Example 2] Game Action for traction-separation Laws. 
Given an 8-tuple $\tuple{G}=(\set{L_{V}}, \set{L_{E}}, \set{V},\set{E},\vec{s},\vec{t}, \vec{n_{V}}, \vec{n_{E}})$ 
with elements defined in \eqref{eq:V}, \eqref{eq:E}, \eqref{eq:LE}, \eqref{eq:LV}. Find the subgraph $\tuple{G}'$
of $\tuple{G}$ such that this subgraph becomes the directed acyclic graph that maximizes the blind prediction accuracy 
defined by an objection function. 
\end{proof}


\section{Two-player meta-modeling game for the discovery of elasto-plastic models through modeling and automated experiments}\label{sec:game}
In this work, we conceptualize the process of writing, calibrating and validating constitutive laws as a 
cooperative two-player game played by one modeler and one experimentalist (data) agent. These two agents, in theory, can be played by either a
human or an artificial intelligence (AI) machine. 
To simplify the problems, we consider only virtual experiments such as discrete element simulations \citep{sun2013multiscale, zohdi2013rapid,  liu2016nonlocal, xin2017discrete, ulven2018capturing, wang2018multiscale} and that the agents are not constrained by the number of virtual experiment tests they might conduct. The control of the experimental cost and the ability to automate the execution of experiments are important topics but are both out of the scope of this work. 

\begin{figure}[h!]\center
		\includegraphics[width=0.9\textwidth]{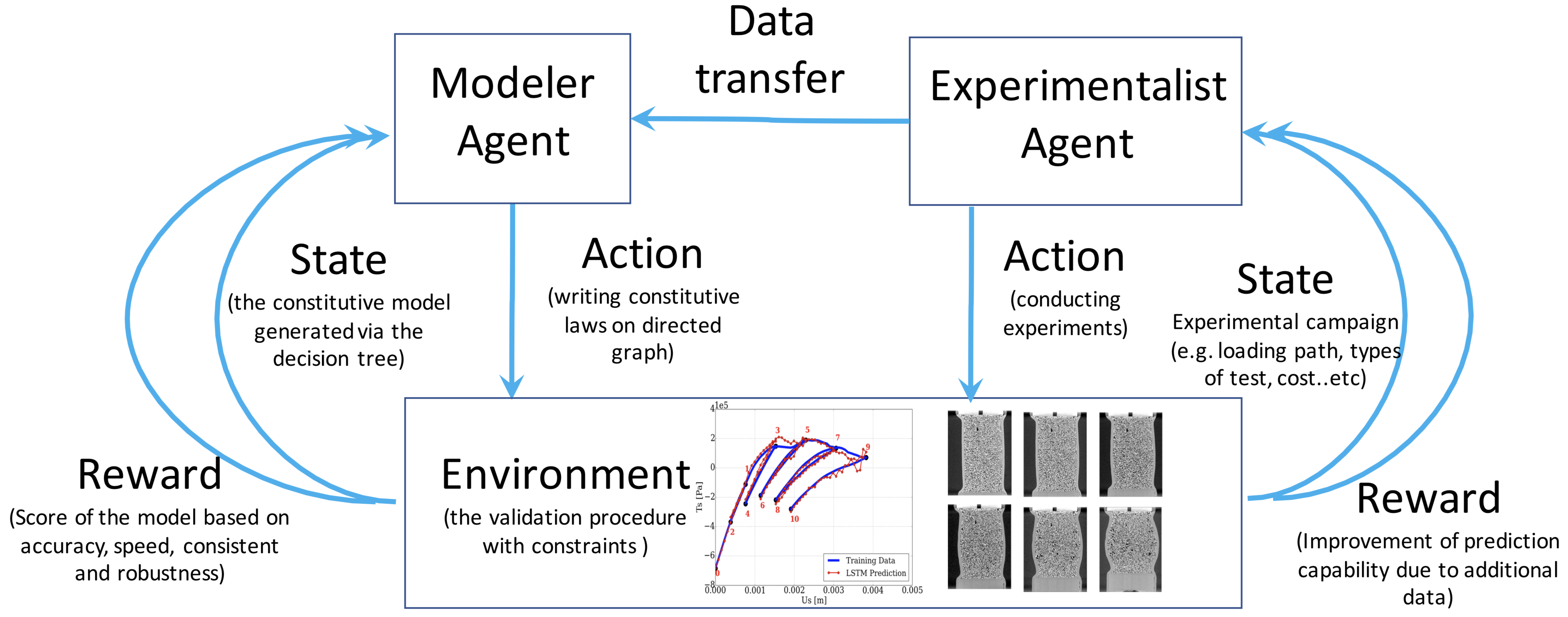}
	\caption{Scheme of the reinforcement learning algorithm in which two agents interact with environment and receives rewards for their corresponding actions (writing models and conducting experiments).}
	\label{fig:twoplayerlearning}
\end{figure}

As such two AI agents must be able to cooperate such that they can 
find the hierarchical relationships among available data and (2) come up with the experiment plan that helps improve the performance of the blind predictions made by the directed graph model, as shown in Figure \ref{fig:twoplayerlearning}. 
This lead to a multi-agent multi-objective problem that can be solved by deep reinforcement learning \citep{tan1993multi, raileanu2018modeling}. 

\subsection{Data collection game for experimentalist agent}
\label{subsec:datagame}
This section presents a design of the data collection game involving the common decision-making activities of experimentalists in testing the mechanical properties of a material. 
The goal of this game is for the experimentalist agent to find the optimal subset of tests for model generation and parameter calibration within a set of candidate tests on the material.
The key ingredients of the game are detailed as follows.

\subsubsection{Game Board for Experimentalist} \label{sec:gbe}
Consider a set of possible mechanical experiments on a material $\textbf{T} = \{ T_1, T_2, T_3, ..., T_n \}$. 
The experiments can be divided into two types: (1) a subset $\textbf{T}_c$ of calibration experiments for material parameter identification in a constitutive model, (2) a subset $\textbf{T}_v$ of validation experiments for testing the forward prediction accuracy of the constitutive model. 
$\textbf{T} = \textbf{T}_c \cup \textbf{T}_v$, $\textbf{T}_c \cap \textbf{T}_v = \emptyset$, $\textbf{T}_c \neq \emptyset$ and $\textbf{T}_v \neq \emptyset$. 
Suppose the experimentalist has a priori preselected the elements in both categories: $\textbf{T}_c = \textbf{T}_c^0 = \{ T_{c1}, T_{c2}, T_{c3}, ..., T_{cn} \}$ and $\textbf{T}_v = \textbf{T}_v^0 = \{ T_{v1}, T_{v2}, T_{v3}, ..., T_{vn} \}$. 
This selection could be based on the availability of laboratory equipment, i.e., $\textbf{T}_c^0$ includes all tests that the experimentalist can perform in the laboratory, while $\textbf{T}_v^0$ includes other tests that can only be acquired from literature or third-party laboratories. 
The experimentalist then chooses the final set of experiments $\textbf{T}_c \subset \textbf{T}_c^0$ which could generate necessary and sufficient data for the modeler agent to develop and calibrate a constitutive model with the highest model score. 
The final validation set $\textbf{T}_v$ contains both experiments in $\textbf{T}_v^0$ and those not selected in $\textbf{T}_c$, i.e., $\textbf{T}_v = \textbf{T}_v^0 \cup (\textbf{T}_c^0 \setminus \textbf{T}_c)$. 
Hence the set $\textbf{T}_c^0$ constitutes the "game board" for the experimentalist agent to play the data collection game. 

\subsubsection{Game State for Experimentalist} \label{sec:gse}
The mathematical description of the current state of the game board is a list of binary indicators $s = [i_{c1}, i_{c2}, i_{c3}, ..., i_{cn}, i_{terminate}]$ representing whether a test $T_{ci} \in \textbf{T}_c^0$ is selected to be one of the calibration tests, and also whether the game is terminated. 
If $T_{ci} \in \textbf{T}_c$, the corresponding indicator $i_{ci}=1$, if $T_{ci} \notin \textbf{T}_c$ $i_{ci}=0$. 
If $i_{terminate} = 1$, the game reaches the end, otherwise the experimentalist can continue. 
The initial state of the game is $i_{ci}=0,\ \forall T_{ci} \in \textbf{T}_c^0$ and $i_{terminate} = 0$. 
A special final state in which $i_{ci}=0,\ \forall T_{ci} \in \textbf{T}_c^0$ and $i_{terminate} = 1$ indicates that no data is available for model generation and calibration, hence the reward for this state is set to $0$.

\subsubsection{Game Action for Experimentalist} \label{sec:gae}
At each state $s$, the experimentalist can select the next calibration test $T_{ci} \in \textbf{T}_c$, by changing the indicator $i_{ci}$ from 0 to 1, or decide to stop the selection immediately, by changing $i_{terminate}$ from 0 to 1.

\subsubsection{Game Rule for Experimentalist} \label{sec:grue}
Generally, there are no specific rules constraining the selection of experiments for model parameter calibration. 
But the game designer could always customize certain rules that prohibit the coexistence of certain experiments in $\textbf{T}_c$. 
The game rule can be reflected by a list of binaries $LegalActions(s) = [ii_{c1}, ii_{c2}, ii_{c3}, ..., ii_{cn}, ii_{terminate}]$, indicating whether an element $i_{ci}$ of the state $s$ can be changed in the next action step. \\
$\bullet$ If $i_{ci} = 0$ in the current state $s$, then $ii_{ci}=1$ in $LegalActions(s)$.\\
$\bullet$ If $i_{ci} = 1$ , then $ii_{ci}=0$.\\
$\bullet$ if $i_{terminate} = 0$ , then $ii_{terminate}=1$. \\
We enforce a game rule that require the two tests $T_{ci}$ and $T_{cj}$ are mutually exclusive in $\textbf{T}_c$. \\
$\bullet$ If $i_{ci} = 1$ , then $ii_{cj} = 0$, and vice versa. \\
The initial legal actions are $ii_{ci}=1,\ \forall T_{ci} \in \textbf{T}_c^0$ and $ii_{terminate} = 1$.

\subsubsection{Game Reward for Experimentalist} \label{sec:gree}
The reward from the game environment to the experimentalist agent should consider the scores of the constitutive models generated by the modeler, given the calibration data and validation data prepared by the experimentalist. 
For each result of the data collection game $\textbf{T}_c$ (hence its pair $\textbf{T}_v = \textbf{T} \setminus \textbf{T}_c$), the modeler could generate a number of different constitutive models with scores $[\text{SCORE}_{i,\ i=1,2,3,...}]_{\textbf{T}_c}$. 
The reward should also consider the total cost of the calibration tests $\textbf{T}_c$. 
This can be measured by a weighted sum $\text{COST}(\textbf{T}_c) = \sum^{\textbf{T}_c^0} w^{cost}_{ci} * i_{ci}$, where $w^{cost}_{ci}$ is the normalized cost for test $T_{ci} \in \textbf{T}_c^0$, 
$\sum^{\textbf{T}_c^0} w^{cost}_{ci} = 1$, 
$w^{cost}_{ci} \in [0,1]$. 

If the experimentalist and the modeler are fully cooperative on generating the constitutive model with the highest score, the reward $r$ is a function of the maximum model scores for all possible $\textbf{T}_c \subset \textbf{T}_c^0$ and the total experimental cost of $\textbf{T}_c$. 
Suppose that since the beginning of the two-payer cooperative game (Figure \ref{fig:twoplayerlearning}), the experimentalist have experienced a number of calibration test sets $\textbf{T}_c$ (they constitute a set $\mathbb{T}_c^{\text{history}}$), and the modeler have generated constitutive models and evaluated their scores for these calibration test sets ($[\text{SCORE}_{i,\ i=1,2,3,...}]_{\textbf{T}_c},\ \forall \textbf{T}_c \in \mathbb{T}_c^{\text{history}}$). 
Then both agent have the knowledge of the highest model score for each $\textbf{T}_c$: $\text{SCORE}_{\textbf{T}_c}^{\max} = \max([\text{SCORE}_{i,\ i=1,2,3,...}]_{\textbf{T}_c})$. 
Thus they know the highest model score in the history of self-played games: $\text{SCORE}^{\max} = \max(\text{SCORE}_{\textbf{T}_c}^{\max}),\ \forall \textbf{T}_c \in \mathbb{T}_c^{\text{history}}$. 
Then the agents can identify a set $\mathbb{T}_c^{\text{max}} \subset \mathbb{T}_c^{\text{history}}$ in which the elements are all calibration test sets that can lead to maximum scores comparable to the highest score, i.e.,  
$\textbf{T}_c \in \mathbb{T}_c^{\text{max}}$, if $ |\text{SCORE}_{\textbf{T}_c}^{\max} - \text{SCORE}^{\max}| \leq \text{TOL}$, where TOL is a small tolerance criteria. 

From the perspective of the experimentalist agent, for a fully cooperative game, $\textbf{T}_c$ (represented by the state $s$) is winning the data collection game if it is an element of the set $\mathbb{T}_c^{\text{max}}$, and if its total cost is the lowest among all elements in $\mathbb{T}_c^{\text{max}}$. Hence the reward is designed as
\begin{equation}
r(s) = \left\{
\begin{aligned}
&1,\ \ \ \text{if}\ \textbf{T}_c \in \mathbb{T}_c^{\text{max}} \text{and}\ \text{COST}(\textbf{T}_c) \leq \text{COST}(\forall \textbf{T}_c^i \in \mathbb{T}_c^{\text{max}})\\
&0,\ \ \ \text{otherwise}
\end{aligned}
\right.,
\end{equation}

\subsubsection{Game Choices for Experimentalist} \label{sec:gce} The elements in the set $\textbf{T} = \{ T_1, T_2, T_3, ..., T_n \}$ could be all possible mechanical experiments on a material. 
For example, for granular materials, the candidates can include the following common types of tests in soil laboratories: 
\begin{enumerate}
	\item Drained conventional triaxial test ($\dot{\epsilon}_{11} \neq 0$, $\dot{\sigma}_{22}=\dot{\sigma}_{33}=\dot{\sigma}_{12}=\dot{\sigma}_{23}=\dot{\sigma}_{13}=0$).
	\item Drained true triaxial test ($\dot{\epsilon}_{11} \neq 0$, $b = \frac{\sigma_{22}-\sigma_{33}}{\sigma_{11}-\sigma_{33}}$, $\dot{\sigma}_{33}=\dot{\sigma}_{12}=\dot{\sigma}_{23}=\dot{\sigma}_{13}=0$).
	\item Undrained triaxial test ($\dot{\epsilon}_{11} \neq 0$, $\dot{\epsilon}_{11}+\dot{\epsilon}_{22}+\dot{\epsilon}_{33}=0$, $\dot{\sigma}_{22}=\dot{\sigma}_{33}$, $\dot{\sigma}_{12}=\dot{\sigma}_{23}=\dot{\sigma}_{13}=0$).
	\item One-dimensional test ($\dot{\epsilon}_{11}\neq0$, $\dot{\epsilon}_{22}=\dot{\epsilon}_{33}=\dot{\epsilon}_{12}=\dot{\epsilon}_{23}=\dot{\epsilon}_{13}=0$).
	\item Simple shear test ($\dot{\epsilon}_{12}>0$, $\dot{\sigma}_{11}=\dot{\sigma}_{22}=\dot{\epsilon}_{33}=\dot{\epsilon}_{23}=\dot{\epsilon}_{13}=0$).
\end{enumerate}
The loading conditions are represented by constraints on the components of the stress rate and strain rate tensors
\begin{equation}
\dot{\tensor{\epsilon}} =
\begin{bmatrix}
\dot{\epsilon}_{11} & \dot{\epsilon}_{12} & \dot{\epsilon}_{13}\\
& \dot{\epsilon}_{22} & \dot{\epsilon}_{23}\\
\text{sym} &  & \dot{\epsilon}_{33}\\
\end{bmatrix},\ 
\dot{\tensor{\sigma}} =
\begin{bmatrix}
\dot{\sigma}_{11} & \dot{\sigma}_{12} & \dot{\sigma}_{13}\\
& \dot{\sigma}_{22} & \dot{\sigma}_{23}\\
\text{sym} &  & \dot{\sigma}_{33}\\
\end{bmatrix}.
\end{equation}

\textit{Remarks on implementation} In the numerical testing of the constitutive models, the above material test conditions are applied via a linearized integration technique for loading constraints of laboratory experiments $\tensor{S} d \tensor{\sigma} + \tensor{E} d \tensor{\epsilon} = d\vec{Y}$, combined with incremental constitutive equations, as proposed in \citep{bardet1991linearized}.

\subsection{Meta-modeling game for modeler agent}
\label{subsec:modelgame}
This section presents a design of the constitutive modeling game involving the common decision-making activities of modelers in developing models to approximate the mechanical properties of a material. 
The goal of this game is for the modeler agent to find the optimal configuration of the directed graph from a predefined directed multigraph (Section \ref{sec:metamodeling})  with its structure inherited from the graphs of the classical infinitesimal strain elasto-plasticity models. 
The key ingredients of the meta-modeling game consist of {\it game agents, game board,  game state, game actions, game Rules, game reward and game choices} such that it constitutes an agent-environment interactive system \citep{bonabeau2002agent, wang2019meta}
which are detailed as follows.

\subsubsection{Game Board for Modeler} \label{sec:gbm}
\begin{figure}[h!]\center
	\includegraphics[width=0.99\textwidth]{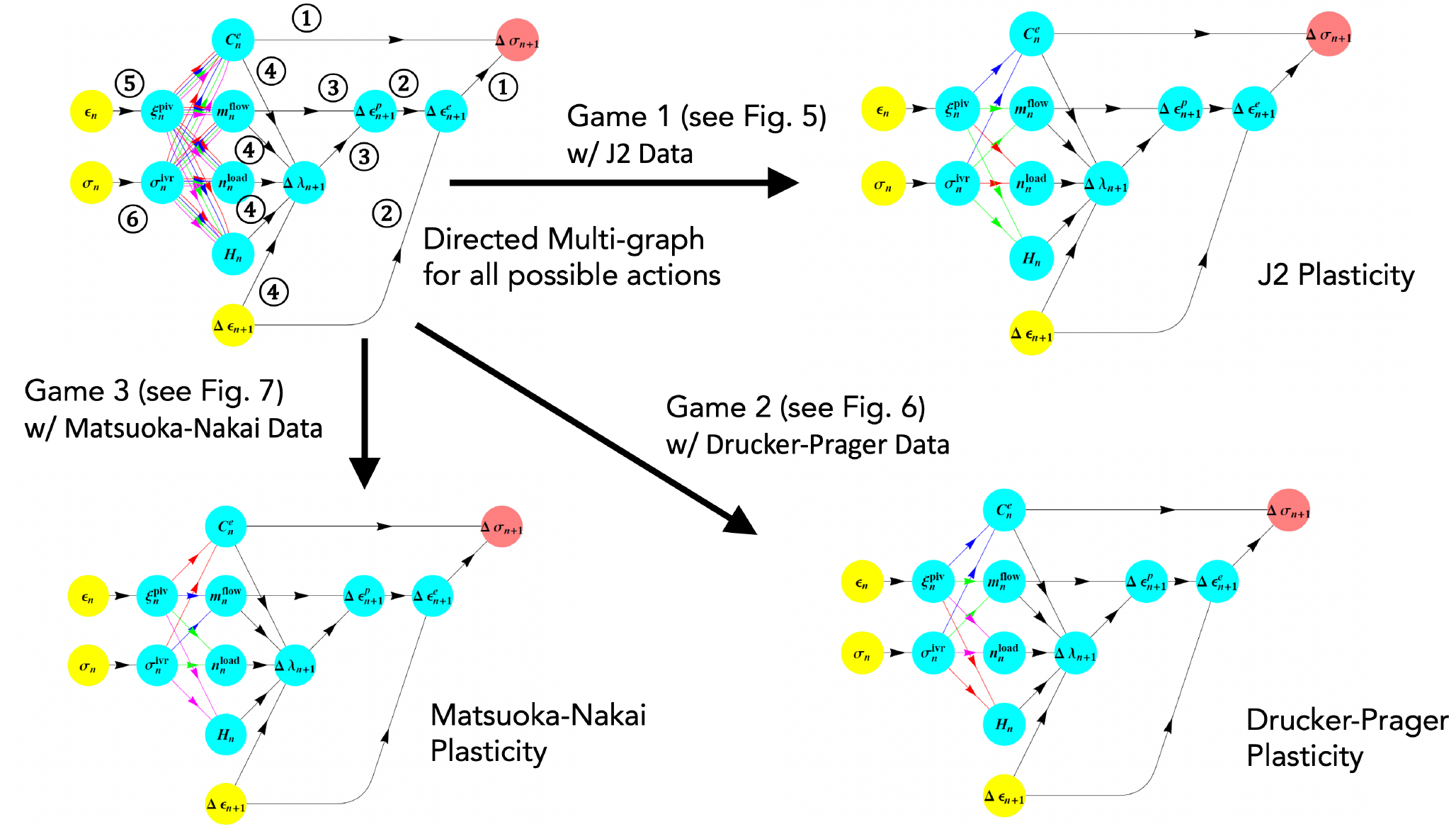}
	\caption{Directed multigraph of an elasto-plasticity model.The yellow nodes of the strain $\tensor{\epsilon}_n$, stress $\tensor{\sigma}_n$ and strain increment $\Delta\tensor{\epsilon}_{n+1}$ refer to the root nodes, the pink node of the stress increment $\Delta\tensor{\sigma}_{n+1}$ refers to the leaf node, and the cyan nodes refer to intermediate nodes. The black arrows refer to "definition" edges. The color arrows refer to "phenomenological relations" edges. In the Meta-modeling game, the modeler AI agent generates the optimal configuration of the model from the labeled directed multi-graph for a given set of data. In the case of reverse engineering, the modeler AI agent should be able to recover the 
	original constitutive laws when given the corresponding types of data.}
	\label{fig:ElastoPlasticityGraph}
\end{figure}

A constitutive model in the generalized elasto-plasticity framework \citep{pastor1990generalized, zienkiewicz1999computational} requires four essential components of "phenomenological relations" : (1) elasticity law (2) loading direction (3) plastic flow direction (4) hardening modulus. 
The process of obtaining a directed graph (the final state of the game) from the game board, i.e., the direct multigraph of the proposed framework is presented in Figure \ref{fig:ElastoPlasticityGraph}. 
The quantities are presented in the incremental form at discrete time steps. 
A quantity $a$ at the current time step $t_n$ is denoted as $a_n = a(t_n)$. 
The next time step is $t_{n+1}$ with the time increment $\Delta t = t_{n+1} - t_n$. 
Then the increment of the quantity $a$ within $\Delta t$ is denoted as $\Delta a_{n+1} = a_{n+1} - a_{n}$. 
The essential "definition" edges in the direct multigraph are written as

\begin{equation}
\begin{aligned}
&\circled{1} & \Delta\tensor{\sigma}_{n+1} &= \tensor{C}^e_n : \Delta\tensor{\epsilon}^e_{n+1}\\
&\circled{2} & \Delta\tensor{\epsilon}^e_{n+1} &= \Delta\tensor{\epsilon}_{n+1} - \Delta\tensor{\epsilon}^p_{n+1}\\
&\circled{3} & \Delta\tensor{\epsilon}^p_{n+1} &= \Delta\lambda_{n+1} \tensor{m}^{flow}_{n}\\
&\circled{4} & \Delta\lambda_{n+1} &= \left\{
\begin{aligned}
&\frac{\tensor{n}^{load}_{n} : \tensor{C}^e_n : \Delta\tensor{\epsilon}_{n+1}}{H_n + \tensor{n}^{load}_{n} : \tensor{C}^e_n : \tensor{m}^{flow}_{n}}\ \ \ & \text{if plastic loading}\\
&\ \ \ \ \ \ 0\ \ \ & \text{if elastic loading}\\
\end{aligned}
\right. ,
\end{aligned}
\label{eq:def_1}
\end{equation}
where $\Delta\lambda_{n+1}$ is the plastic multiplier and $H_n$ is the generalized plastic modulus. 

The "elastic loading" and "plastic loading" states are determined via the projection of the trial elastic stress increment $\Delta\tensor{\sigma}^e_{n+1} = \tensor{C}^e_n : \Delta\tensor{\epsilon}_{n+1}$ on the loading direction $\tensor{n}^{load}_{n}$. 
If there is no assumed yield surface, then
\begin{equation}
\left\{
\begin{aligned}
\Delta\tensor{\sigma}^e_{n+1}:\tensor{n}^{load}_{n} &\neq 0 \rightarrow \text{plastic loading}\\
\Delta\tensor{\sigma}^e_{n+1}:\tensor{n}^{load}_{n} &= 0 \rightarrow \text{elastic loading}\\
\end{aligned}
\right.,
\label{eq:loadcond_1}
\end{equation}
or if there exists a yield surface $f(\tensor{\sigma}, \vec{q}^{piv}_n(\vec{\xi}^{piv}_n))$, then 
\begin{equation}
\left\{
\begin{aligned}
f(\tensor{\sigma}_n + \Delta\tensor{\sigma}^e_{n+1}, \vec{q}^{piv}_n(\vec{\xi}^{piv}_n)) &> 0 \rightarrow \text{plastic loading}\\
f(\tensor{\sigma}_n + \Delta\tensor{\sigma}^e_{n+1}, \vec{q}^{piv}_n(\vec{\xi}^{piv}_n)) &\leq 0 \rightarrow \text{elastic loading}\\
\end{aligned}
\right.,
\label{eq:loadcond_2}
\end{equation}
where $\vec{\xi}^{piv}_n$ is a vector of strain-like plastic internal variables and $\vec{q}^{piv}_n$ is a vector of stress-like plastic internal variables conjugate to $\vec{\xi}^{piv}_n$. 
$\vec{\xi}^{piv}_n$ may include the following internal state variables accumulated during the deformations from the initial time $t_0$ to the current time $t_n$,
\begin{equation}
\circled{5}
\left\{
\begin{aligned}
\lambda_n &= \int_{0}^{t_n}\dot{\lambda} dt\\
\bar{\epsilon}^p_n &= \int_{0}^{t_n}||\dot{\tensor{\epsilon}^p}|| dt\\
\bar{\epsilon}^p_{v_n} &= \int_{0}^{t_n}\text{tr}(\dot{\tensor{\epsilon}^p}) dt\\
\bar{\epsilon}^p_{s_n} &= \int_{0}^{t_n}||\dot{\tensor{\epsilon}^p} - \frac{1}{3} \text{tr}(\dot{\tensor{\epsilon}^p})||dt\\
e_{n} &= e_{0} + \int_{0}^{t_n}\dot{e} dt = e_{0} + \int_{0}^{t_n} (1+e) \dot{\epsilon}_v dt\\
\end{aligned}
\right. ,
\label{eq:piv_def}
\end{equation}
where $\bar{\epsilon}^p$, $\bar{\epsilon}^p_{v}$ and $\bar{\epsilon}^p_{s}$ are accumulated total, volumetric and deviatoric plastic strains, respectively. 
$e$ is the void ratio for granular materials, defined as the ratio between volume of the void and the solid constituent. 
We assume that the yield function is isotropic and therefore can be expressed in terms of stress invariants \citep{borja2013plasticity}. 
As a result,  the phenomenological relations can be represented as functions of a stress invariants $\vec{\sigma}^{ivr}_n$, which may include 
\begin{equation}
\circled{6}
\left\{
\begin{aligned}
p_n &= \frac{\text{tr}(\tensor{\sigma}_n)}{3}\\
q_n &= \sqrt{3 J_2} = \sqrt{\frac{3}{2}} ||\tensor{s}_n||\\
\theta_n &= \frac{1}{3} \sin^{-1}(-\frac{3\sqrt{3}}{2} \frac{J_3}{J_2^{3/2}}),\ -\frac{\pi}{6} \leq \theta \leq \frac{\pi}{6}
\end{aligned}
\right.
\end{equation}
where $J_2 = \frac{1}{2} \text{trace}(\tensor{s}_n^2)$, $J_3 = \frac{1}{3} \text{trace}(\tensor{s}_n^3)$, $\tensor{s}_n = \tensor{\sigma}_n-p_n\tensor{I}$ and $\theta_{n}$ is the Lode's angle, the smallest angle between the line of pure shear and the projection of 
stress tensor in the deviatoric plane \citep{malcher2009numerical}.
The constitutive relation between the loading direction $\tensor{n}^{load}$ and the state variables $\vec{\xi}^{piv}_n$, $\vec{\sigma}^{ivr}_n$ can be defined either by formulating a yield surface $f$ such that,
\begin{equation}
\tensor{n}^{load} = \frac{\partial f}{\partial \tensor{\sigma}} ||\frac{\partial f}{\partial \tensor{\sigma}}||^{-1}, 
\end{equation}
or, in the case yield surface is absence, directly inferred from observations as those in the generalized plasticity framework (cf. \citet{lubliner1996generalized, pastor1990generalized, ling2006unified}), 
\begin{equation}
\tensor{n}^{load} = n^{load}_v \tensor{n}_v + n^{load}_s \tensor{n}_s.
\end{equation}
where
\begin{equation}
\left\{
\begin{aligned}
\tensor{n}_v &= \frac{\partial p}{\partial \tensor{\sigma}} = \frac{1}{3}\tensor{I}\\
\tensor{n}_s &= \frac{\partial q}{\partial \tensor{\sigma}} = \frac{\sqrt{3}}{2\sqrt{J_2}}\tensor{S}.\\
\end{aligned}
\right.
\end{equation}

Similarly, the constitutive relation between the plastic flow direction $\tensor{m}^{flow}$ and the state variables $\vec{\xi}^{piv}_n$, $\vec{\sigma}^{ivr}_n$ can be defined either by formulating a plastic potential surface $g$ such that, 
\begin{equation}
\tensor{m}^{flow} = \frac{\partial g}{\partial \tensor{\sigma}} ||\frac{\partial g}{\partial \tensor{\sigma}}||^{-1}.
\end{equation}
or directly inferred from observations as those in the generalized plasticity framework (cf. \citet{lubliner1996generalized, pastor1990generalized, ling2006unified})
\begin{equation}
\tensor{m}^{flow} = m^{flow}_v \tensor{n}_v + m^{flow}_s \tensor{n}_s.
\end{equation}

\subsubsection{Game State for Modeler} \label{sec:gsm}
The mathematical description of the current state of the game board is a list of binary indicators $s = [i_{e1}, i_{e2}, i_{e3}, ..., i_{en}]$ representing whether a labeled edge $E_{ei}$ in the labeled edge set $\set{L_{E}}$ of the directed multigraph $\tuple{G}$ is selected in the final generated directed graph $\tuple{G}'$. 
If $E_{ei}$ is included in $\tuple{G}'$, the corresponding indicator $i_{ei}=1$, otherwise $i_{ei}=0$. 
The initial state of the game is $i_{ei}=0,\ \forall E_{ei} \in \set{L_{E}}$. 

\subsubsection{Game Action for Modeler} \label{sec:gam}
At each state $s$, the modeler can select the next labeled edge $E_{ei} \in \set{L_{E}}$, by changing the indicator $i_{ei}$ from 0 to 1.

\subsubsection{Game Rule for Modeler} \label{sec:grum}
The modeling choices for the four essential components in an elasto-plasticity model are not fully compatible with each other. For example, a J2 yield surface only has the yield stress as the stress-like plastic internal variable, while a strain hardening law for a Drucker–Prager yield surface has both frictional and cohesion hardening laws. These restrictions on compatible edge choices are specified 
by a list of binaries $LegalActions(s) = [ii_{1}, ii_{2}, ii_{3}, ..., ii_{n}]$ of legal choices for each state. 
Another set of game rules consist of universal principles on the constitutive models. For example, thermodynamic consistency states that the rate of mechanical dissipation must be non-negative, for isothermal process $\mathcal{D} = \tensor{\sigma}:\dot{\tensor{\epsilon}} - \frac{d \psi}{d t} \geq 0$. This game rule is incorporated in the game by the definition of the model score. If the final model in an episode violates this rule, the final model score is set to be 0. This low score is then used as training material for the mastermind modeler agent such that it 
reduces the policy probabilities of the choices that violate universal principles as shown in Figure \ref{fig:selfplay_learn}. As the training of the constitutive law can only be completed if the score of  the best candidate model is sufficiently high, this prevents the meta-modeling algorithm from generating any model that violates the first principles. 

\subsubsection{Game Reward for Modeler} \label{sec:grem}
A score system must be introduced to evaluate the generated directed graphs for constitutive models 
such that the accuracy and credibility in replicating the mechanical behavior of real-world materials can be assessed. 
This score system may also serve as the objective function that defines the rewards for the deep reinforcement learning agent to improve the generated digraphs and resultant constitutive laws. 
In this work, we define 
the score as a positive real-valued function of the range $[0,1]$ which depends on the measures $A_i$ $(i=1,2,3,...,n)$ of $n$ important features of a constitutive model,
\begin{equation}
\text{SCORE} = F(A_1, A_2, A_3, ..., A_n),
\end{equation}
where $0\leq A_i \leq 1$. 
Some features are introduced to measure the performance of a model such as the accuracy and computation speed. 
Other features are introduced to enforce constraints to ensure 
the admissibility of a constitutive model, such as the frame indifference and the thermodynamic consistency. 
Suppose there are $n_{\text{pfm}}$ measures of performance features $A^{\text{pfm}}_i$ and $n_{\text{crit}}$ measures of critical features $A^{\text{crit}}_i$ in the measure system of constitutive models, the score takes the form,
\begin{equation}
\text{SCORE} = (\prod_{j=1}^{n_{\text{crit}}} A^{\text{crit}}_j) \cdot (\sum_{i=1}^{n_{\text{pfm}}} w_i A^{\text{pfm}}_i),
\end{equation}
where $w_i \in [0,1]$ is the weight associated with the measure $A^{\text{pfm}}_i$, and $\sum_{i=1}^{n_{\text{pfm}}} w_i = 1$. 

For example, for measures of accuracy $A_{\text{accuracy}}$ of calibrations and forward predictions, we introduce a cross-validation procedure in which the dataset used for training  the models (e.g. identifying material 
parameters (e.g. \citet{wang2016identifying, liu2016determining}) or adjusting weights of neurons in recurrent neural networks (e.g. \citet{lefik2002artificial, wang2018multiscale}) is mutually exclusive to the testing dataset used to evaluate the quality of blind predictions. 
Both calibration and blind prediction results are compared against the target data. 
The mean squared error (MSE) commonly used in statistics and also as objective function in machine learning is chosen as the error measure for each data sample $i$ in this study, i.e., 
\begin{equation}
\text{MSE}_{i} = \frac{1}{N_{\text{feature}}} \sum_{j=1}^{N_{\text{feature}}} [\mathcal{S}_{j} (Y_{i_{j}}^{\text{data}}) -\mathcal{S}_{j} (Y_{i_{j}}^{\text{model}})]^2,
\label{eq:mse_data_i}
\end{equation}
where $Y_{i_{j}}^{\text{data}}$ and $Y_{i_{j}}^{\text{model}}$ are the values of the $j$th feature of the $i$th data sample, from target data value and predictions from constitutive models, respectively. 
$N_{\text{feature}}$ is the number of output features. 
$\mathcal{S}_{j}$ is a scaling operator (standardization, min-max scaling, ...) for the output feature $\{Y_{i_{j}}\},\ i \in [1,N_{\text{data}}]$. 

The empirical cumulative distribution functions (eCDFs) are computed for MSE of the entire dataset $\{\text{MSE}_{i}\},\ i \in [1,N_{\text{data}}]$, for MSE of the training dataset $\{\text{MSE}_{i}\},\ i \in [1,N_{\text{traindata}}]$ and for MSE of the test dataset $\{\text{MSE}_{i}\},\ i \in [1,N_{\text{testdata}}]$, with the eCDF defined as \citep{kendall1946advanced},
\begin{equation}
F_{N}(\text{MSE}) = \left \{
\begin{aligned}
&0, &\text{MSE} < \text{MSE}_1,\\
&\frac{r}{N}, &\text{MSE}_{r} \leq \text{MSE} < \text{MSE}_{r+1},\ r = 1,...,N-1,\\
&1, &\text{MSE}_{N} \leq \text{MSE},
\end{aligned}
\right .
\label{eq:ecdf_mse}
\end{equation}
where $N = N_{\text{data}}$, or $N_{\text{traindata}}$, or $N_{\text{testdata}}$, and all $\{\text{MSE}_{i}\}$ are arranged in increasing order. 
A measure of accuracy is proposed based on the above statistics,
\begin{equation}
A_{\text{accuracy}} = \max(\frac{ \log [\max(\varepsilon_{P\%}, \varepsilon_{\text{crit}})] }{\log \varepsilon_{\text{crit}}}, 0),
\label{eq:acc_indicator}
\end{equation}
where $\varepsilon_{P\%}$ is the $P$th percentile (the MSE value corresponding to $P\%$ in the eCDF plot) of the eCDF on the entire, training or test dataset. 
$\varepsilon_{\text{crit}} \ll 1$ is the critical MSE chosen by users such that a model can be considered as "satisfactorily accurate" when $\varepsilon_{P\%} \leq \varepsilon_{\text{crit}}$.

Once a complete constitutive model is generated, the model score is evaluated. 
The final reward is defined as: if the current score is higher than the average score of models from a group of already played games by the agent, then the current game wins and $r_T=1$, otherwise, the current game loses and $r_T=-1$. 
The average score can be initialized to 0 for the first game.

\subsubsection{Game Choices for Modeler} \label{sec:gcm}
This section specifies the candidate edges in the directed multi-graph of elasto-plasticity models (Fig. \ref{fig:ElastoPlasticityGraph}) for the modeler agent to choose during deep reinforcement learning. 
The edges are categorized into four groups representing the four essential constitutive relations in the model. The edges $\vec{\sigma}^{ivr}_n \rightarrow \tensor{C}^e_n$ and $\vec{\xi}^{piv}_n \rightarrow \tensor{C}^e_n$ represent the elasticity law. 
The edges $\vec{\sigma}^{ivr}_n \rightarrow \tensor{n}^{load}_{n}$ and $\vec{\xi}^{piv}_n \rightarrow \tensor{n}^{load}_{n}$ represent the definition of the loading direction. 
The edges $\vec{\sigma}^{ivr}_n  \rightarrow \tensor{m}^{flow}_{n}$ and $\vec{\xi}^{piv}_n \rightarrow \tensor{m}^{flow}_{n}$ represent the definition of the plastic flow direction.
The edges $\vec{\sigma}^{ivr}_n \rightarrow H_{n}$ and $\vec{\xi}^{piv}_n \rightarrow H_{n}$ represent the hardening law.
Each edge allows multiple choices extracted from the phenomenological relations developed in the computational plasticity literature. 
In this paper, for simplicity of illustration of the meta-modeling game framework, the edge choices are not exhaustive. 
The following lists only contain common representative choices for geomaterials. 
But the designer of the meta-modeling game is always free to add more edge choices to expand the action space. 

The edges for elasticity law ($\vec{\sigma}^{ivr}_n \rightarrow \tensor{C}^e_n$ and $\vec{\xi}^{piv}_n \rightarrow \tensor{C}^e_n$) represent the definition and evolution of the elastic stiffness tensor 
\begin{equation}
\tensor{C}^e_n = K \tensor{I}\otimes\tensor{I} + 2 G (\tensor{I}^4_{sym} - \frac{\tensor{I}\otimes\tensor{I}}{3}),
\end{equation}
where $K$ is the elastic bulk modulus and $G$ is the elastic shear modulus. 

Three common formulations of the elastic stiffness tensor for granular materials are available for model choice:
\begin{description}
\item[(E1)] Linear elasticity
\begin{equation}
\left\{
\begin{aligned}
K&=K_0\\
G&=G_0
\end{aligned}
\right.,
\end{equation}
where $K_0$ and $G_0$ are constants. 
\item[(E2)] Nonlinear elasticity with dependence on the mean pressure $p$ \citep{manzari1997critical}
\begin{equation}
\left\{
\begin{aligned}
K&= K_0(\frac{p}{p_{at}})^a\\
G&= G_0(\frac{p}{p_{at}})^a
\end{aligned}
\right.,
\label{eq:elast_2}
\end{equation}
where $p_{at}$ is the atmospheric pressure ($\approx$ -100 kPa) and $a$ is a material constant. 
\item[(E3)] Nonlinear elasticity with dependence on the mean pressure $p$ and the void ratio $e$ \citep{dafalias2004simple}
\begin{equation}
\left\{
\begin{aligned}
K&= \frac{2(1+\nu)}{3(1-2\nu)} G\\
G&= G_0 p_{at} \frac{(2.97-e)^2}{1+e} (\frac{p}{p_{at}})^{1/2}
\end{aligned}
\right.,
\end{equation}
where $\nu$ is the constant Poisson's ratio.
\end{description}

The edges ($\vec{\sigma}^{ivr}_n \rightarrow \tensor{n}^{load}_{n}$ and $\vec{\xi}^{piv}_n \rightarrow \tensor{n}^{load}_{n}$) represent the definition and evolution of the loading direction.  
$\tensor{n}^{load}_{n}$ can be either derived from an assumed yield surface $f \leq 0$ or defined explicitly in the space of stress invariants $\vec{\sigma}^{ivr}_n$.

The following common formulations of loading direction for granular materials are considered for model choices: 
\begin{description}
\item[(L1)] Yield surface of J2 plasticity $f = q - \sigma_y$ and linear hardening law
\begin{equation}
\sigma_y = \sigma_{y0} + H_0 \bar{\epsilon}^p,
\end{equation}
where $\sigma_{y0},H_0$ are material parameters. 
\item[(L2)] Yield surface of J2 plasticity $f = q - \sigma_y$ and $\sigma_y$ is the solution of the power law equation 
\begin{equation}
\frac{\sigma_y}{\sigma_{y0}} = (\frac{\sigma_y}{\sigma_{y0}} + \frac{3G}{\sigma_{y0}}\bar{\epsilon}^p)^n,
\end{equation}
where $\sigma_{y0},n$ are material parameters, $G$ is the elastic shear modulus.
\item[(L3)] Yield surface of J2 plasticity $f = q - \sigma_y$ and Voce hardening law
\begin{equation}
\sigma_y = \sigma_{y0} + H_0 \bar{\epsilon}^p + H_{\infty}(1-\exp(-b\bar{\epsilon}^p)),
\end{equation}
where $\sigma_{y0}, H_0, H_{\infty},b$ are material parameters. 
\item[(L4)] Yield surface of Drucker–Prager plasticity $f = q + \alpha p$ and $\alpha$ evolves according to
\begin{equation}
\alpha = a_0 + a_1\bar{\epsilon}^p \exp(a_2 p - a_3 \bar{\epsilon}^p),
\label{eq:n_DP_0}
\end{equation}
where $a_0, a_1, a_2, a_3$ are material parameters \citep{tu2009return}. 
\item[(L5)] Yield surface of Drucker–Prager plasticity $f = q + \alpha p$ and $\alpha$ evolves according to
\begin{equation}
\alpha = a_0 + 2 a_1 \frac{\sqrt{k \bar{\epsilon}^p}}{k+\bar{\epsilon}^p},
\label{eq:n_DP_1}
\end{equation}
where $a_0, a_1, k$ are material parameters \citep{borja2013plasticity}. 
\item[(L6)] Yield surface of three-invariant Matsuoka–Nakai model \citep{borja2003numerical}
\begin{equation}
\left\{
\begin{aligned}
f &= (k_1 I_3)^{1/3} - (I_1 I_2)^{1/3}\\
k_1 &= c_0 + \kappa_1 (\frac{p_{at}}{I_1})^m\\
\kappa_1 &= a_1 \bar{\epsilon}^p \exp(a_2 I_1) \exp(-a_3 \bar{\epsilon}^p)
\end{aligned}
\right.,
\label{eq:n_MN_0}
\end{equation}
where $c_0, a_1, a_2, a_3, m$ are material parameters. 
\item[(L7)] Yield surface of Nor-Sand \citep{jefferies1993nor, andrade2006capturing}
\begin{equation}
\left\{
\begin{aligned}
f &= \zeta q + \eta p\\
\zeta &= \frac{(1+\rho)+(1-\rho)\cos3(\theta+\pi/6)}{2 \rho}\\
\eta &= \left\{\begin{aligned}
&M[1+\log(p_i/p)]\ &\text{if}\ N=0\\
&(M/N)[1-(1-N)(p/p_i)^{N/(1-N)}]\ &\text{if}\ N>0\\
\end{aligned}\right.\\
\dot{p_i} &= -\sqrt{\frac{2}{3}} h(p_i-p^*_i) ||\dot{\tensor{e}^p}||,\ \dot{\tensor{e}^p} = \dot{\tensor{\epsilon}^p} - \frac{1}{3} tr(\dot{\tensor{\epsilon}^p}) \tensor{I}\\
\frac{p^*_i}{p} &= \left\{\begin{aligned}
&\exp(\bar{\alpha}\psi_i/M)\ &\text{if}\ \bar{N}=N=0\\
&(1-\bar{\alpha} \psi_i N/M)^{(N-1)/N}]\ &\text{if}\ 0 \leq \bar{N} \leq N \neq 0\\
\end{aligned}\right.\\
\bar{\alpha} &= -3.5 \frac{1-\bar{N}}{1-N}\\
\psi_i &= e - e_{c0} + \tilde{\lambda} (p_i/p_{at})^{a}
\end{aligned}
\right.,
\end{equation}
where $\rho,N,\bar{N},M,h,e_{c0},\tilde{\lambda},a$ are material parameters. 
\item[(L8)] Yield surface in the shape of a small cone \citep{dafalias2004simple}
\begin{equation}
\left\{
\begin{aligned}
f &= ||\tensor{S} - p \tensor{\alpha}|| - \sqrt{2/3}pm\\
\dot{\tensor{\alpha}} &= \dot{\lambda} (2/3) h (\tensor{\alpha}_{\theta}^b - \tensor{\alpha})\\
\tensor{\alpha}_{\theta}^b &= \sqrt{2/3} [\frac{1}{\zeta} M \exp(-n^b \psi)-m] \tensor{n}\\
\zeta &= \frac{(1+\rho)+(1-\rho)\cos3(\theta+\pi/6)}{2 \rho}\\
\tensor{n} &= \frac{\frac{\tensor{S}}{p} - \tensor{\alpha}}{\sqrt{2/3}m}\\
\psi &= e - e_{c0} + \tilde{\lambda} (p/p_{at})^{a}
\end{aligned}
\right.,
\end{equation}
where $\rho,m,M,n^b,h,e_{c0},\tilde{\lambda},a$ are material parameters. 
\item[(L9)] Loading direction defined as \citep{pastor1990generalized, zienkiewicz1999computational}
\begin{equation}
\left\{
\begin{aligned}
n^{load}_v &= \frac{d_f}{\sqrt{1+d_f^2}}\\
n^{load}_s &= \frac{1}{\sqrt{1+d_f^2}}\\
d_f &= (1+\alpha)(M_f+q/p)
\end{aligned}
\right.,
\end{equation}
where $\alpha,M_f$ are material parameters. 
\item[(L10)] Loading direction defined as \citep{ling2006unified}
\begin{equation}
\left\{
\begin{aligned}
n^{load}_v &= \frac{d_f}{\sqrt{1+d_f^2}}\\
n^{load}_s &= \frac{1}{\sqrt{1+d_f^2}}\\
d_f &= (1+\alpha)(M_f \exp(m_f(1-e)) +q/p)\\
\end{aligned}
\right.,
\label{eq:nload_1}
\end{equation}
where $\alpha,M_f,m_f$ are material parameters. 
\item[(L11)] Loading direction given by a neural network trained with data inversely computed from experimental data (described later in the definition of plastic modulus edges). 
\end{description}

The edges ($\vec{\sigma}^{ivr}_n \rightarrow \tensor{m}^{flow}_{n}$ and $\vec{\xi}^{piv}_n \rightarrow \tensor{m}^{flow}_{n}$) represent the definition and evolution of the plastic flow direction.  
$\tensor{m}^{flow}_{n}$ can be either derived from an assumed plastic potential surface $g=0$ or defined explicitly in the space of stress invariants $\vec{\sigma}^{ivr}_n$.

The following common formulations of the plastic flow direction for granular materials are considered for model choices: 
\begin{description}
\item[(P1)] Plastic potential surface of J2 plasticity $g = q - c_g$ and $c_g$ is a parameter to ensure that the stress point is on the potential surface when the plastic deformation occurs.
\item[(P2)] Plastic potential surface of Drucker–Prager plasticity $g = q + \beta p - c_g$ and $\beta = \alpha - \beta_0$, where $\alpha$ can be defined through Eq. (\ref{eq:n_DP_0}) or (\ref{eq:n_DP_1}), and $\beta_0$ is an additional material parameter. 
\item[(P3)] Plastic potential surface of three-invariant Matsuoka–Nakai model \citep{borja2003numerical}
\begin{equation}
\left\{
\begin{aligned}
g &= (k_2 I_3)^{1/3} - (I_1 I_2)^{1/3}\\
k_2 &= c_0 + \kappa_2 (\frac{p_{at}}{I_1})^m\\
\kappa_2 &= \alpha \kappa_1
\end{aligned}
\right.,
\end{equation}
 where $\kappa_1$ can be defined through Eq. \ref{eq:n_MN_0} and $\beta_0$ is an additional material parameter.
\item[(P4)] Plastic potential surface of Nor-Sand \citep{jefferies1993nor, andrade2006capturing}
\begin{equation}
\left\{
\begin{aligned}
g &= \bar{\zeta} q + \bar{\eta} p\\
\bar{\zeta} &= \frac{(1+\bar{\rho})+(1-\bar{\rho})\cos3(\theta+\pi/6)}{2 \bar{\rho}}\\
\bar{\eta} &= \left\{\begin{aligned}
&M[1+\log(\bar{p}_i/p)]\ &\text{if}\ \bar{N}=0\\
&(M/\bar{N})[1-(1-\bar{N})(p/\bar{p}_i)^{\bar{N}/(1-\bar{N})}]\ &\text{if}\ \bar{N}>0\\
\end{aligned}\right.\\
\end{aligned}
\right.,
\end{equation}
where $\bar{\rho},\bar{N},M$ are material parameters and $\bar{p}_i$ is a free parameter to ensure $g=0$ when the material is undergoing plastic deformation. 
\item[(P5)] Plastic flow direction defined as \citep{dafalias2004simple}
\begin{equation}
\left\{
\begin{aligned}
\tensor{m}^{flow} &= B\tensor{n} -C(\tensor{n}^2-\frac{1}{3}\tensor{I}) + \frac{1}{3}D\tensor{I}\\
B &= 1+\frac{3}{2} \frac{1-c}{c \xi} \cos3(\theta+\pi/6)\\
C &= 3 \sqrt{3/2} \frac{1-c}{c \xi}\\
D &= A_d (\tensor{\alpha}_{\theta}^d - \tensor{\alpha}) : \tensor{n}\\
\tensor{\alpha}_{\theta}^d &= \sqrt{2/3} [\frac{1}{\zeta} M \exp(n^d \psi)-m] \tensor{n}
\end{aligned}
\right.,
\end{equation}
where $\rho,m,M,n^d,A_d,e_{c0},\tilde{\lambda},a$ are material parameters. 
\item[(P6)] Plastic flow direction defined as \citep{pastor1990generalized, zienkiewicz1999computational}
\begin{equation}
\left\{
\begin{aligned}
m^{flow}_v &= \frac{d_g}{\sqrt{1+d_g^2}}\\
m^{flow}_s &= \frac{1}{\sqrt{1+d_g^2}}\\
d_g &= (1+\alpha)(M_g+q/p)
\end{aligned}
\right.,
\end{equation}
where $\alpha,M_g$ are material parameters. 
\item[(P7)] Plastic flow direction defined as \citep{ling2006unified}
\begin{equation}
\left\{
\begin{aligned}
m^{flow}_v &= \frac{d_g}{\sqrt{1+d_g^2}}\\
m^{flow}_s &= \frac{1}{\sqrt{1+d_g^2}}\\
d_g &= (1+\alpha)(M_g \exp{m_g \psi} +q/p)\\
\psi &= e - e_{c0} + \tilde{\lambda} (p/p_{at})^{a}
\end{aligned}
\right.,
\label{eq:mflow_1}
\end{equation}
where $\alpha,M_g,m_g,e_{c0},\tilde{\lambda}, a$ are material parameters. 
\item[(P8)] Plastic flow direction given by a neural network trained with data inversely computed from experimental data (described later in the definition of plastic modulus edges). 
\end{description}

The edges ($\vec{\sigma}^{ivr}_n \rightarrow H_{n}$ and $\vec{\xi}^{piv}_n \rightarrow H_{n}$) represent the definition of the generalized hardening modulus.  
$H_{n}$ can be either derived from an assumed yield surface $f \leq 0$ or defined explicitly.

The following common formulations of hardening modulus for granular materials are considered for model choices: 
\begin{description}
\item[(H1)] Hardening modulus derived from classical yield surface $f(\tensor{\sigma}, \tensor{\epsilon}^p)$ and a chosen $\tensor{m}^{flow}$.
\begin{equation}
H = - \frac{\partial f/ \partial \tensor{\epsilon}^p : \tensor{m}^{flow}}{||\partial f / \partial \tensor{\sigma}||}.
\end{equation}
\item[(H2)] Hardening modulus defined as \citep{pastor1990generalized, zienkiewicz1999computational}
\begin{equation}
\left\{
\begin{aligned}
H &= H_0 (-p) H_f (H_v + H_s)\\
H_f &= (1+\frac{q}{p M_f} \frac{\alpha_f}{1+\alpha_f})^4\\
H_v &= 1+\frac{q}{p M_g}\\
H_s &= \beta_0 \beta_1 \exp(-\beta_0 \bar{\epsilon}^p_{s})
\end{aligned}
\right.,
\label{eq:hhard_1}
\end{equation}
where $\alpha_f,M_f,H_0,e_{c0},M_g,\beta_0,beta_1$ are material parameters. 
\item[(H3)] Hardening modulus defined as \citep{ling2006unified}
\begin{equation}
\left\{
\begin{aligned}
H &= H_0 \sqrt{p/p_{at}} H_f (1+\frac{q}{pM_b})\\
M_b &= M_g \exp(-m_b \psi)\\
H_0 &= H_{L0} \exp(m_0(1-e))\\
H_f &= (1+\frac{q}{p M_f} \frac{\alpha_f}{1+\alpha_f})^4\\
\end{aligned}
\right.,
\end{equation}
where $\alpha_f,M_f,H_{L0},m_0,M_g,m_b,e_{c0},\tilde{\lambda}, a$ are material parameters. 
\item[(H4)] Hardening modulus given by a neural network trained with data inversely computed from experimental data. 
\end{description}

The stress increment at each time step is known from the experimental data $\Delta \tensor{\sigma}^{data}_{n+1} =  \tensor{\sigma}^{data}_{n+1} -  \tensor{\sigma}^{data}_{n}$. 
For a chosen elasticity law $\tensor{C}^e_n(\vec{\sigma}^{ivr}_n, \vec{\xi}^{piv}_n)$, the data of incremental plastic strain at each time step is given by (using Eq. (\ref{eq:def_1}))
\begin{equation}
\Delta\tensor{\epsilon}^p_{n+1} = \Delta\tensor{\epsilon}_{n+1} - (\tensor{C}^e_n)^{-1} : \Delta \sigma^{data}_{n+1}.
\label{eq:rnn_1}
\end{equation}
Then the incremental plastic multiplier is $\Delta \lambda_{n+1} = ||\Delta\tensor{\epsilon}^p_{n+1}||$ and the plastic flow direction is obtained by $\tensor{m}^{flow}_{n} = \Delta\tensor{\epsilon}^p_{n+1}/\Delta \lambda_{n+1}$. 
Assuming associative flow rule, then $\tensor{n}^{load}_{n} = \tensor{m}^{flow}_{n}$. 
In this way, the plastic modulus can be uniquely inversely computed as
\begin{equation}
H_{n} = \frac{\tensor{n}^{load}_{n} : \tensor{C}^e_n : \Delta\tensor{\epsilon}_{n+1}}{\Delta \lambda_{n+1}} - \tensor{n}^{load}_{n} : \tensor{C}^e_n : \tensor{m}^{flow}_{n}.
\label{eq:rnn_2}
\end{equation}

\subsubsection{Game Choice alternatives: training neural network edges} \label{sec:gca}
In addition to the mathematical edges described above, we also consider the possibility of replacing any part 
of the elasto-plastic model with machine learning edges. In this framework, the machine learning models
are not used to directly map strain history to stress, but are used for each individual edge in the directed graph
to map the input vertices to the output vertices. 
For instance, the mapping of variables in the generalized plasticity framework can be obtained by training a recurrent neural network that represents the path-dependent constitutive relation between the history of input vertices of $\vec{\sigma}^{ivr}_n$ ($p,q,\theta$) and $\vec{\xi}^{piv}_n$ ($\bar{\epsilon}^p,\bar{\epsilon}_v^p,\bar{\epsilon}_s^p,e$) and the output vertices of $\tensor{n}^{load}_{n}$, $\tensor{m}^{flow}_{n}$ and $H_{n}$. 
The details of training data preparation, network design, training and testing are specified in the previous work on the meta-modeling framework for traction-separation models with data of microstructural features \citep{wang2019meta}.
In this framework, all neural network edges are generated using the same neural network architecture, i.e., 
two hidden layers of 64 GRU(Gated recurrent unit) neurons in each layer, and the output layer as a dense layer with linear activation function. 
All input and output data are pre-processed by standard scaling using mean values and standard deviations. 
Each input feature considers its current value and 19 history values prior to the current loading step. 
Each neural network is trained for 1000 epochs using the Adam optimization algorithm, with a batch size of 256. 
Finally, it should be noticed that one can further generalize the meta-modeling game by considering multiple 
neural network architectures as possible edges in the meta-modeling game. This generalization will be considered in 
the future but is out of the scope of the current study. 

\textit{Remarks on implementation} An elasto-plasticity model, once generated from AI, needs to be numerically integrated to compute the predicted stresses under different types of tests. Since the loading directions, plastic flow directions and hardening modulus can have a large number of options and may be exceedingly complex, we adopt a general-purpose explicit integration algorithm for all AI generated models, instead of using different implicit integration techniques necessary for different models. 
This algorithm is a combination of (1) the explicit integration with sub-stepping and automatic error control \citep{sloan1987substepping, sloan2001refined} (2) explicit integration of (potentially non-smooth) hardening laws \citep{tu2009return} (3) integration of generalized plasticity models \citep{de2002unified, mira2009generalized} (4) linearized integration for loading constraints \citep{bardet1991linearized}. The algorithm is detailed in Algorithm \ref{explicit_integration_algorithm}. 
This explicit scheme is versatile and stable, but not as accurate as fully implicit return mapping algorithms, hence for the evaluation of model accuracy scores, small time steps are required for the numerical integration.

\begin{algorithm}
	\caption{Explicit Integration Scheme}\label{explicit_integration_algorithm}
	\begin{algorithmic}[1]
		\Require AI-generated directed graph $\tuple{G}=(\set{V},\set{E})$ of the elasto-plasticity model.
		\Require Initial values of stress $\tensor{\sigma}_n$, strain $\tensor{\epsilon}_n$,  plastic internal variables $\vec{\xi}^{piv}_n$, and stress invariants $\vec{\sigma}^{ivr}_n$.
		
		\State Identify elasticity model $\tensor{C^e}(\vec{\sigma}^{ivr}, \vec{\xi}^{piv})$, loading direction $\tensor{n}^{load}(\vec{\sigma}^{ivr}, \vec{\xi}^{piv})$ (or yield surface $f(\vec{\sigma}^{ivr}, \vec{\xi}^{piv})$ if it exists), plastic flow direction $\tensor{m}^{flow}(\vec{\sigma}^{ivr}, \vec{\xi}^{piv})$ (or plastic potential $g(\vec{\sigma}^{ivr}, \vec{\xi}^{piv})$ if it exists), generalized hardening modulus $H(\vec{\sigma}^{ivr}, \vec{\xi}^{piv})$ from the directed graph $\tuple{G}$. 
		
		\State Define matrices of loading constraints $\tensor{E}$, $\tensor{S}$, and $d\vec{Y}$ \citep{bardet1991linearized}.
		\State Update the elastic stiffness tensor $\tensor{C}^e_n(\vec{\sigma}_n^{ivr}, \vec{\xi}_n^{piv})$, its Voigt form $\tensor{D^e}$, and the hardening modulus $H_n(\vec{\sigma}^{ivr}_n, \vec{\xi}^{piv}_n)$.
		\State Solve for the strain increment tensor $\Delta \tensor{\epsilon}_{n+1}$ from Voigt form equation $(\tensor{S} \tensor{D^e} + \tensor{E}) \cdot d \vec{\epsilon} = d\vec{Y}$.
		\State Compute the trial elastic stress increment $\Delta\tensor{\sigma}^e_{n+1} = \tensor{C}^e_n : \Delta\tensor{\epsilon}_{n+1}$ and the trial stress state $\tensor{\sigma}^e_{n+1} = \tensor{\sigma}_n + \Delta\tensor{\sigma}^e_{n+1}$.
		\State Determine the 'Elastic Loading' or 'Plastic Loading' condition from Eq. (\ref{eq:loadcond_1}) and (\ref{eq:loadcond_2}).
		
		\If {Elastic Loading}
		\State $\tensor{\sigma}_{n+1} = \tensor{\sigma}^e_{n+1}$, $\tensor{\epsilon}_{n+1} = \tensor{\epsilon}_n + \Delta\tensor{\epsilon}_{n+1}$, $\vec{\xi}^{piv}_{n+1} = \vec{\xi}^{piv}_{n}$.
		\State Exit
		\EndIf
		
		\If {Plastic Loading}
		
		\If {yield surface $f$ exists}
		\State Find $\alpha$ such that $f(\tensor{\sigma}_n + \alpha \Delta\tensor{\sigma}^e_{n+1}, \vec{\xi}_n^{piv}) = 0$ (Pegasus intersection scheme \citep{sloan2001refined}).
		\Else 
		\State Set $\alpha=0$.
		\EndIf
		
		\State Set $T=0$, $\Delta T=1$, $\Delta \lambda = 0$, $\tensor{\sigma}_T=\tensor{\sigma}_n + \alpha \Delta\tensor{\sigma}^e_{n+1}$, $\tensor{\epsilon}_T=\tensor{\epsilon}_n + \alpha \Delta\tensor{\epsilon}_{n+1}$.
		\While {$T < 1$}
		\State Compute $\tensor{n}^{load}(\tensor{\sigma}_T, \vec{\xi}_n^{piv})$, $\tensor{m}^{flow}(\tensor{\sigma}_T, \vec{\xi}_n^{piv})$.
		\State Update the elasto-plastic stiffness tensor $\tensor{C^{ep}} = \tensor{C^{e}} - \frac{1}{\chi} \tensor{C}^{e} : \tensor{m}^{flow} \otimes \tensor{n}^{load} : \tensor{C}^{e}$, 
		where $\chi = \tensor{n}^{load} : \tensor{C}^{e} : \tensor{m}^{flow} + H_n$ and its Voigt form $\tensor{D^{ep}}$.
		
		\State Solve for the strain increment tensor $\Delta \epsilon_T$ from Voigt form equation $(\tensor{S} \tensor{D^{ep}} + \tensor{E}) \cdot d \vec{\epsilon}_T = (1-\alpha) \cdot \Delta T \cdot d\vec{Y}$.
		
		\State Update $\Delta\tensor{\sigma}^e \leftarrow \tensor{C^{e}} : \Delta \epsilon_T$.
		
		\State Compute $\Delta \lambda_1 = \frac{1}{\chi} \tensor{n}^{load} : \Delta\tensor{\sigma}^e$, $\Delta \tensor{\sigma}_1 = \Delta\tensor{\sigma}^e - \tensor{C^{e}} : \Delta \lambda_1 \tensor{m}^{flow}$
		
		\State Compute $\tensor{n}^{load}(\tensor{\sigma}_T+\Delta \tensor{\sigma}_1, \vec{\xi}_n^{piv})$, $\tensor{m}^{flow}(\tensor{\sigma}_T+\Delta \tensor{\sigma}_1, \vec{\xi}_n^{piv})$.
		
		\State Compute $\Delta \lambda_2 = \frac{1}{\chi} \tensor{n}^{load} : \Delta\tensor{\sigma}^e$, $\Delta \tensor{\sigma}_2 = \Delta\tensor{\sigma}^e - \tensor{C^{e}} : \Delta \lambda_2 \tensor{m}^{flow}$
		
		\State Compute $\tilde{\tensor{\sigma}}_{T+\Delta T} = \tensor{\sigma}_T +0.5\cdot(\Delta \tensor{\sigma}_1+\Delta \tensor{\sigma}_2)$, and determine the relative error $R_{T + \Delta T} = \max(\frac{|| \Delta \tensor{\sigma}_2 - \Delta \tensor{\sigma}_1 ||}{2|| \tilde{\tensor{\sigma}}_{T+\Delta T} ||}, EPS)$
		
		\If {$R_{T + \Delta T} \leq STOL$}
		\State $\tensor{\sigma}_{T+\Delta T} = \tilde{\tensor{\sigma}}_{T+\Delta T}$, $\tensor{\epsilon}_{T+\Delta T} = \tensor{\epsilon}_{T} + \Delta \epsilon_T$, $\Delta \lambda \leftarrow \Delta \lambda + 0.5\cdot(\Delta \lambda_1 + \Delta \lambda_2)$.
		\State $T \leftarrow T + \Delta T$
		\State Determine new $\Delta T$ for the next substep \citep{sloan2001refined}.
		\Else
		\State Determine new $\Delta T$ for this failed substep \citep{sloan2001refined}.
		\EndIf
		\EndWhile
		
		\State $\tensor{\sigma}_{n+1} = \tensor{\sigma}_T$, $\tensor{\epsilon}_{n+1} = \tensor{\epsilon}_T$, update $\vec{\xi}_{n+1}^{piv}$ from Eq. (\ref{eq:piv_def}). 
		
		\State Exit
		
		\EndIf
	\end{algorithmic}
\end{algorithm}

\section{Deep reinforcement learning for the two-player meta-modeling game} \label{sec:drl_algorithm}
\begin{figure}[h!]\center
	\includegraphics[width=1.0\textwidth]{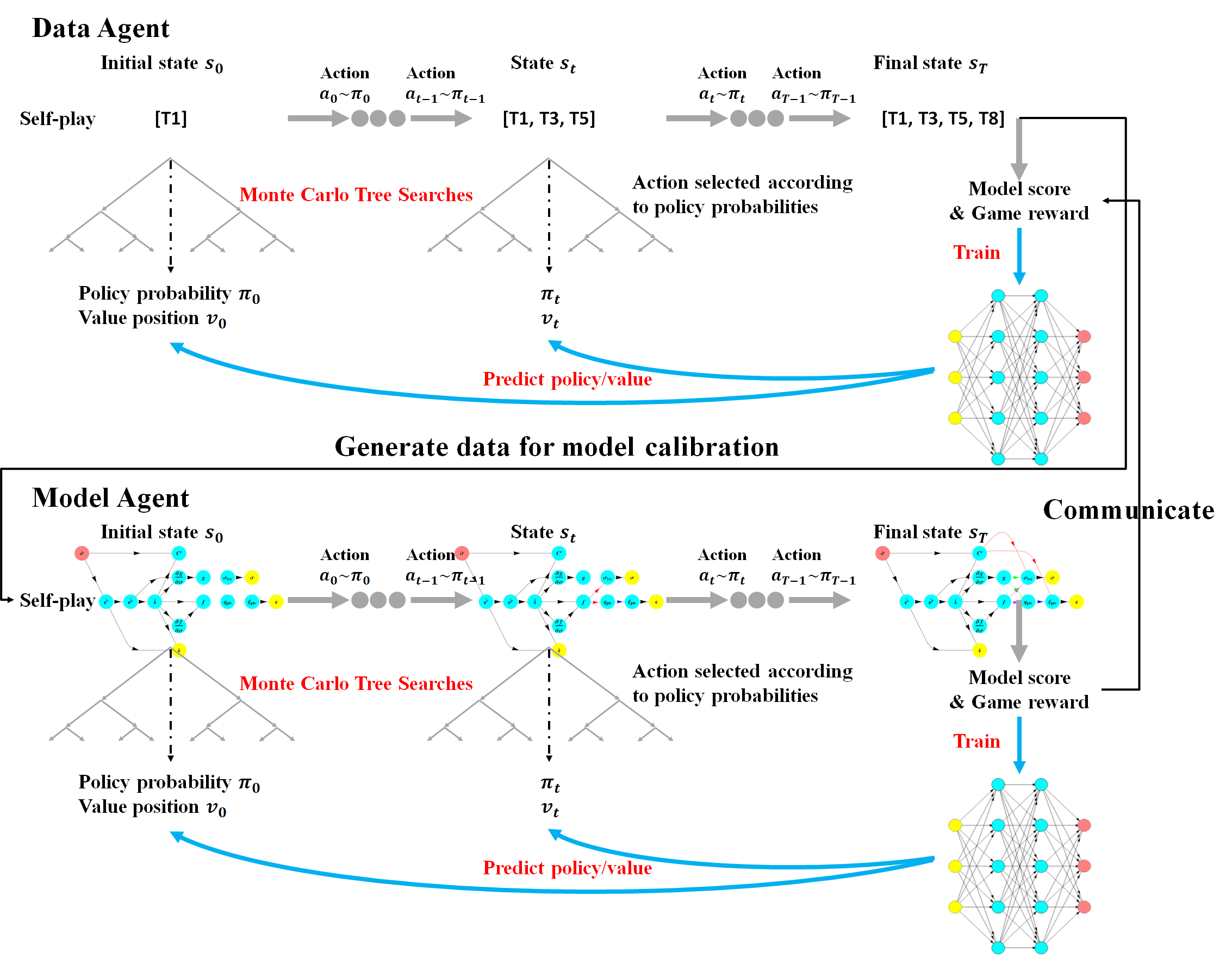}
	\caption{Multi-player interactive deep reinforcement learning for data-driven discovery of elasto-plasticity mechanisms. In this framework,
	two deep neural networks are used to make decisions the mastermind model and data agents, while the model agents may also employ different strategies, including neural networks, mathematical expressions or other forms of mapping operators to compete a constitutive law.}
	\label{fig:selfplay_learn}
\end{figure}

With the two-player game completely defined in the previous section, a deep reinforcement learning (DRL) algorithm is employed as a guidance of taking actions of both experimentalist and modeler in the game to maximize the final model score (Figure \ref{fig:selfplay_learn}). 
The learning is completely free of human interventions after the game settings. 
This tactic is considered one of the key ideas leading to the major breakthrough in AI playing 
the game of Go (AlphaGo Zero) \citep{silver2017mastering}, Chess and shogi (Alpha Zero) \citep{silver2017masteringchess} and many other games. 
In \citep{wang2019meta}, the key ingredients (Policy/Value network, confidence bound for Q-value, Monte Carlo Tree Search) of the DRL technique are detailed and applied to a meta-modeling game for modeler agent only, focusing on finding the optimal topology of physical relations from fixed training/testing datasets. 
In this work, the game design is further extended that (1) the modeling game also involves the "component selection" from a set of candidate edge choices having the same source and target nodes (derive a directed graph from a directed multigraph) and (2) the choice of training dataset is carried out by an additional experimentalist agent. 
Since DRL needs to figure out the optimal strategies for two agents, the algorithm is extended to multi-agent multi-objective DRL \citep{tan1993multi,foerster2016learning,tampuu2017multiagent}. 
The AI for experimentalist and modeler agents are separate, each has its own Policy/Value network and decision tree search. 
But their intelligence are improved simultaneously during the self-plays of the entire Data collection/Meta-modeling game, according to the individual rewards they receive from the game environment and the communications between themselves (Figure \ref{fig:selfplay_learn}). 
The strategies of both agents can be cooperative or competitive of different degrees, depending on the design of the game reward system (for example, the video game of Pong in \citep{tampuu2017multiagent}). 
In this work, we consider only the learning of fully cooperative strategies, as shown in the game reward system designed in Sections \ref{subsec:datagame} and \ref{subsec:modelgame}. 

The pseudocode of the reinforcement learning algorithm to play the two-player meta-modeling game is presented  in Algorithm \ref{mcts_algorithm}. 
This is an extension of the algorithm in \citep{wang2019meta}. 
As demonstrated in Algorithm \ref{mcts_algorithm}, 
each complete DRL procedure involves $numIters$ number of training iterations and one final iteration for generating the final converged digraph model. Each iteration involves $numEpisodes$ number of game episodes that construct the training example set $trainExamples$ for the training of the policy/value network $f_{\theta}$. For decision makings in each game episode, the action probabilities are estimated from $numMCTSSims$ times of MCTS simulations. 

\begin{algorithm}
	\caption{Self-play reinforcement learning of the cooperative data collection/meta-modeling game}\label{mcts_algorithm}
	\begin{algorithmic}[1]
		\Require The definitions of the cooperative data collection/meta-modeling game: game boards (Sections \ref{sec:gbe} and \ref{sec:gbm}), states (Sections \ref{sec:gse} and \ref{sec:gsm}), actions (Sections \ref{sec:gae} and \ref{sec:gam}), game rewards (Sections \ref{sec:gree} and \ref{sec:grem}), game rules (Sections \ref{sec:grue} and \ref{sec:grum}).
		\State Initialize the policy/value networks $f_{\theta}^{\text{Data}}$ and $f_{\theta}^{\text{Model}}$ for both data agent and model agent. For fresh learning, the networks are randomly initialized. For transfer learning, load pre-trained networks instead.
		\State Initialize empty sets of the training examples for both data agent and model agent $trainExamples^{\text{Data}} \leftarrow []$, $trainExamples^{\text{Model}} \leftarrow []$.
		\For{i in [0,..., $numIters$ (number of training iterations)-1]}
		\For{j in [1,..., $numEpisodes$ (number of game episodes)]}
		\State Initialize the starting game state $s$.
		\State Initialize empty tree of the Monte Carlo Tree search (MCTS), set the temperature parameter $\tau = 1$ for "exploration and exploitation".
		\While{True}
		\State Check for all legal actions at current state $s$ according to the game rules.
		\State Get the action probabilities $\pi(s,\cdot)$ for all legal actions by performing $numMCTSSims$ times of MCTS simulations.
		\State Sample action $a$ from the probabilities $\pi(s,\cdot)$
		\State Modify the current game state to a new state $s$ by taking the action $a$.
		\If{$s$ is the end state of a game}
		\State Evaluate the score of the constructed digraph.
		\State Evaluate the reward $r$ of this game episode according to the model score.
		\State \textbf{Break}.
		\EndIf
		\EndWhile
		\State Append the history in this game episode $[s,a,\pi(s,\cdot),r]$ to $trainExamples^{\text{Data}}$ and $trainExamples^{\text{Model}}$.
		\EndFor
		\State Train the policy/value networks $f_{\theta}^{\text{Data}}$ and $f_{\theta}^{\text{Model}}$ with $trainExamples^{\text{Data}}$ and $trainExamples^{\text{Model}}$.
		\EndFor
		
		\State Use the final trained networks $f_{\theta}^{\text{Data}}$ and $f_{\theta}^{\text{Model}}$ in MCTS for one more iteration of "competitive gameplays" ($numEpisodes$ games) to generate the final converged digraph model.
		\State Exit
	\end{algorithmic}
\end{algorithm}


Remark: \textbf{Non-cooperative meta-modeling game and Nash equilibrium}. 
In the case of the cooperative game where both agents share the same goal or score system, there is no need 
to determine the Nash equilibrium as the joint actions of the experimentalist/modeler group takes a collective of payoffs. 
However, in many realistic situations in modern-day research, it is possible that the data
and modeler agents may have different or even conflicting goals and hence finding the best strategies the two agents take is equivalent to finding the Nash equilibrium. The meta-modeling model, in this case, is not only helpful for generating models but also helps us understand the relationships among objectives between the data and modeler agents, the resultant 
actions taken by both players, and the outcomes, assuming each player is acting in a rational manner. 

\section{Numerical Experiments} \label{sec:numericalExp}
In this section, we present two cooperative modeling games with different data to 
demonstrate the intelligence, robustness and efficiency of the deep reinforcement learning algorithm on improving
the accuracy and consistency of the generated elasto-plasticity models through self-plays. 
In the first example, synthetic data computed from selected J2, Drucker-Prager and Matsuoka–Nakai plasticity are used to train the data and model agents, to validate the meta-modeling framework and show that the AI has the ability to react appropriately such that the correct interpretation  (i.e. the model itself if the data is generated from that model) can be recovered from the data. 
In the second example, sub-scale discrete element simulations (DEM) are used to generate synthetic benchmark data for model calibrations and blind prediction evaluations to mimic data from real-world granular materials. 


\subsection{Numerical Experiment 1: Testing the ability of AI for reverse engineering constitutive laws}
The correctness of the proposed meta-modeling framework is first verified through a series of tests on "virtual materials" having exact elasto-plastic constitutive behaviors. 
The goal of this example is to show that the framework can \textit{exactly} recover all edge components of a pre-selected directed graph for elasto-plastic constitutive model, based on the data from AI-selected experiments. 
In other words, the purpose of this numerical experiment is a verification  exercise that tests
 whether both agents can automatically derive the right strategies to recover the models from data 
 without explicit input of human intervention during the training, under the idealized condition that 
 the data does not contain any noise. 
The test models for the AI to recover are, respectively, 
\begin{enumerate}
	\item J2 model with Von Mises yield function and an isotropic hardening with power law.
	\item Drucker-Pager model with frictional hardening.
	\item Three-invariant Matsuoka–Nakai model.
\end{enumerate}

In the reverse engineering numerical experiments, we first implement three implicit return mapping algorithm for the aforementioned models. 
The experimentalist agent is then given the executable files of the return mapping algorithms of these models and run these executable files 
to generate data. Meanwhile, the modeler agent uses the data generated from the experimentalist agent as input. In each iteration of a training session, the experimentalist agent can decide to terminate the numerical tests at any time. 
The experimental test choices available for the experimentalist agent consist of
\begin{description}
	\item[T1:] One-dimensional extension test ($\dot{\epsilon}_{11}>0$,  $\dot{\epsilon}_{22}=\dot{\epsilon}_{33}=\dot{\epsilon}_{12}=\dot{\epsilon}_{23}=\dot{\epsilon}_{13}=0$, $p0=-200kPa$)
	\item[T2:] One-dimensional compression test ($\dot{\epsilon}_{11}<0$,  $\dot{\epsilon}_{22}=\dot{\epsilon}_{33}=\dot{\epsilon}_{12}=\dot{\epsilon}_{23}=\dot{\epsilon}_{13}=0$, $p0=-200kPa$).
	\item[T3:] Drained triaxial extension test ($\dot{\epsilon}_{11} > 0$, $\dot{\sigma}_{22}=\dot{\sigma}_{33}=\dot{\sigma}_{12}=\dot{\sigma}_{23}=\dot{\sigma}_{13}=0$, $p0=-200kPa$).
	\item[T4:] Drained triaxial compression test ($\dot{\epsilon}_{11} < 0$, $\dot{\sigma}_{22}=\dot{\sigma}_{33}=\dot{\sigma}_{12}=\dot{\sigma}_{23}=\dot{\sigma}_{13}=0$, $p0=-200kPa$).
	\item[T5:] Undrained triaxial extension test ($\dot{\epsilon}_{11} > 0$, $\dot{\epsilon}_{11}+\dot{\epsilon}_{22}+\dot{\epsilon}_{33}=0$, $\dot{\sigma}_{22}=\dot{\sigma}_{33}$, $\dot{\sigma}_{12}=\dot{\sigma}_{23}=\dot{\sigma}_{13}=0$, $p0=-200kPa$).
	\item[T6:] Undrained triaxial compression test ($\dot{\epsilon}_{11} < 0$, $\dot{\epsilon}_{11}+\dot{\epsilon}_{22}+\dot{\epsilon}_{33}=0$, $\dot{\sigma}_{22}=\dot{\sigma}_{33}$, $\dot{\sigma}_{12}=\dot{\sigma}_{23}=\dot{\sigma}_{13}=0$, $p0=-200kPa$).
	\item[T7:] Simple shear test ($\dot{\epsilon}_{12}>0$, $\dot{\sigma}_{11}=\dot{\sigma}_{22}=\dot{\epsilon}_{33}=\dot{\epsilon}_{23}=\dot{\epsilon}_{13}=0$, $p0=-200kPa$).
\end{description}

The modeling choices available for the modeler agent are specified in the \textit{Game Choices} of the Section \ref{subsec:modelgame}. 
The model score is defined as:
\begin{equation}
\text{SCORE} = 0.5*A^{\text{calibration}}_{\text{accuracy}} + 0.5*A^{\text{prediction}}_{\text{accuracy}},
\label{eq:game1_score}
\end{equation}
where $P\%=80\%$ and $\varepsilon_{\text{crit}}=1e^{-5}$ for Eq. (\ref{eq:acc_indicator}) of accuracy evaluations. 
The DRL meta-modeling procedure (Algorithm \ref{mcts_algorithm}) contains $numIters=10$ training iterations of "exploration and exploitation" of game strategies, by setting the temperature parameter $\tau$ to 1. 
Then an iteration of "competitive gameplay" ($\tau=0.01$) is conducted to showcase the performance of the final trained AI agent. 
Each iteration consists of $numEpisodes=30$ self-play episodes of the game. 
Hence one execution of the entire DRL procedure contains $numIters * numEpisodes = 10 * 30 = 300$ game episodes for training the policy/value neural network. 
Each game starts with a randomly initialized neural network for the policy/value predictions, and each play step requires $numMCTSSims=30$ MCTS simulations. 
Then the play steps and corresponding final game rewards are appended to the set of training examples for the training of the policy/value network. 

\begin{figure}[h!]\center
	\includegraphics[width=0.95\textwidth]{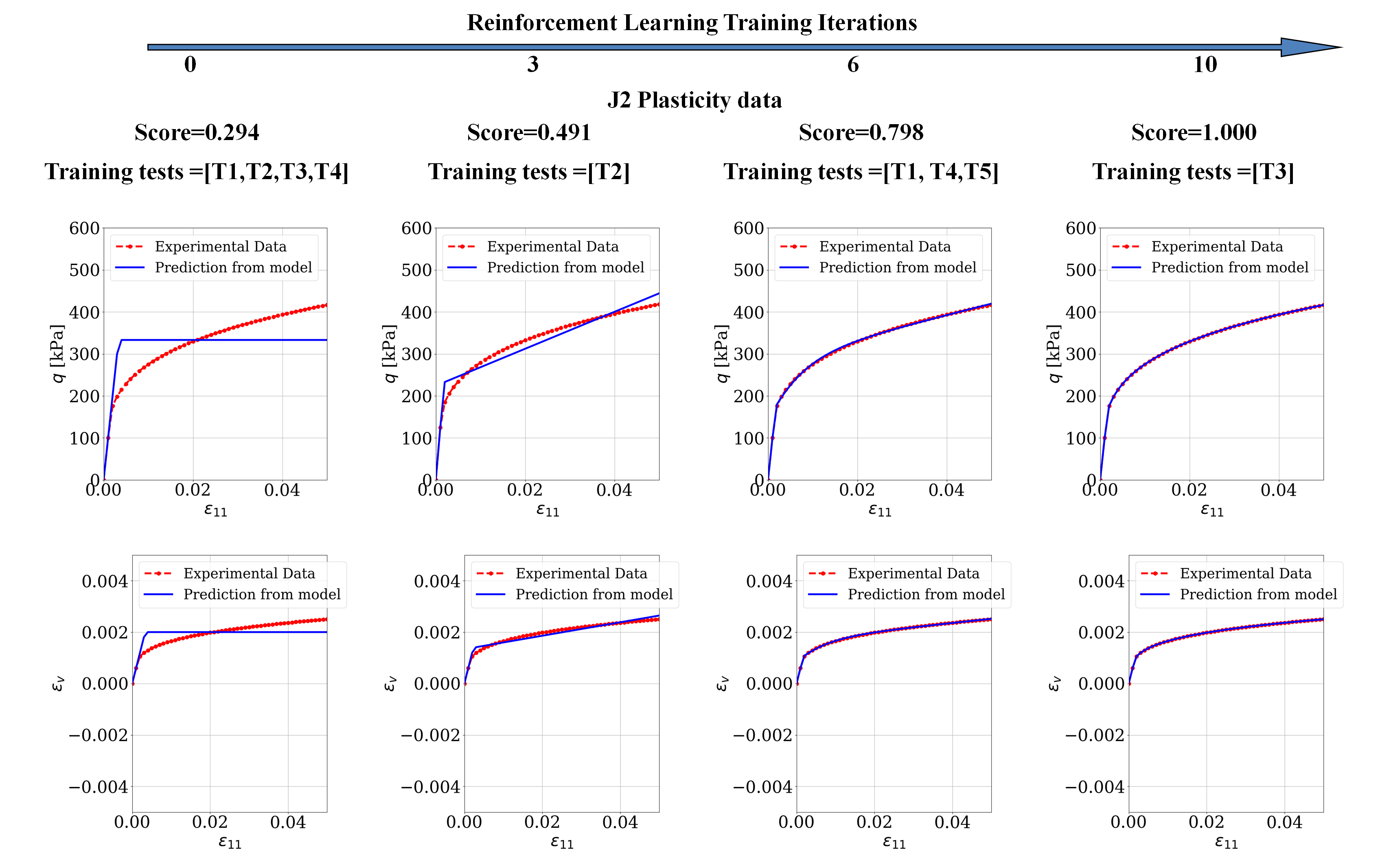}
	\caption{Knowledge of elasto-plastic models learned by deep reinforcement learning in Numerical Experiment 1 using synthetic data from J2 plasticity. Four representative games played during the DRL iterations and their prediction accuracy against synthetic data are presented.}
	\label{fig:example1_improve_J2}
\end{figure}

\begin{figure}[h!]\center
	\includegraphics[width=0.95\textwidth]{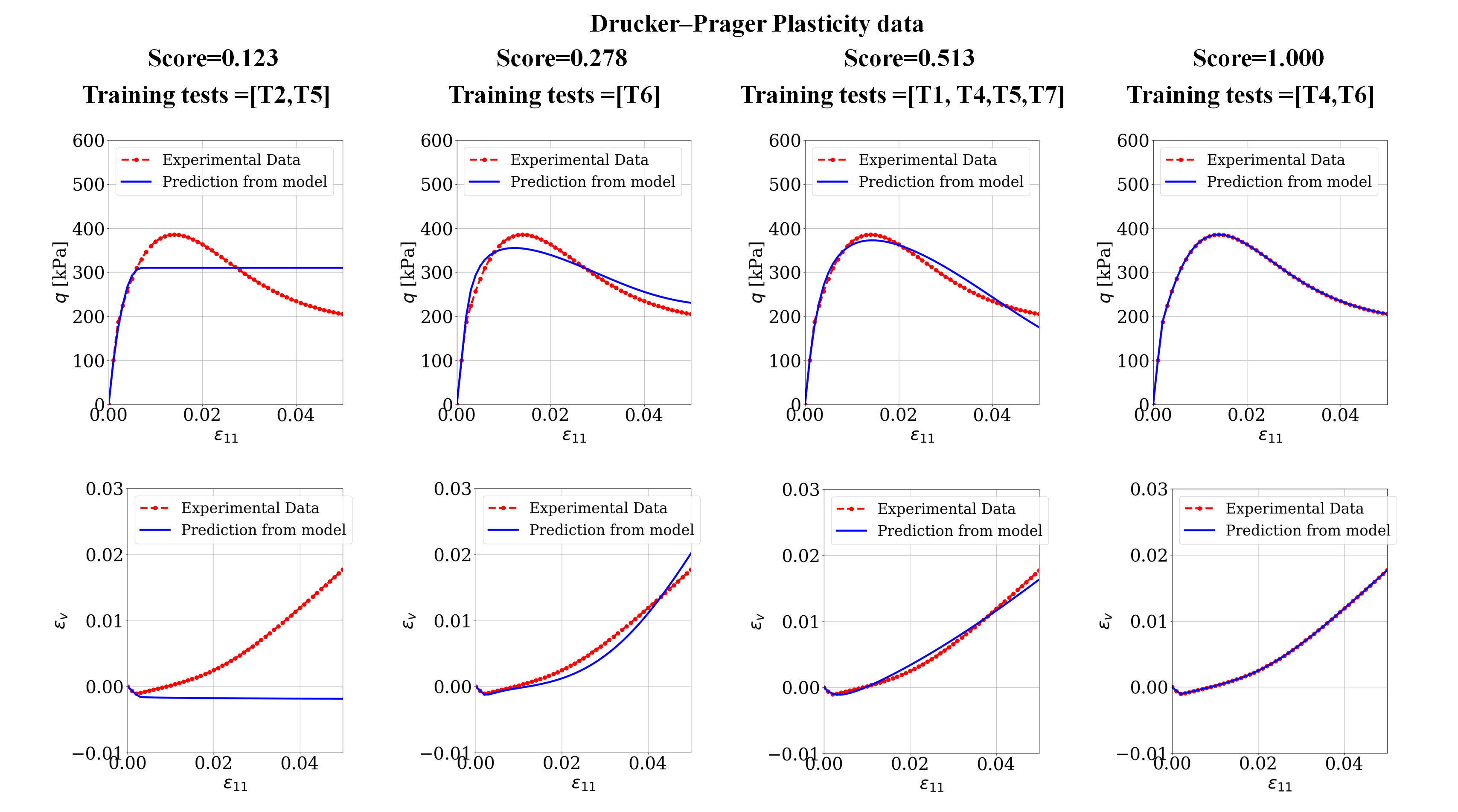}
	\caption{Knowledge of elasto-plastic models learned by deep reinforcement learning in Numerical Experiment 1 using synthetic data from Drucker-Pager plasticity. Four representative games played during the DRL iterations and their prediction accuracy against synthetic data are presented.}
	\label{fig:example1_improve_DP}
\end{figure}

\begin{figure}[h!]\center
	\includegraphics[width=0.95\textwidth]{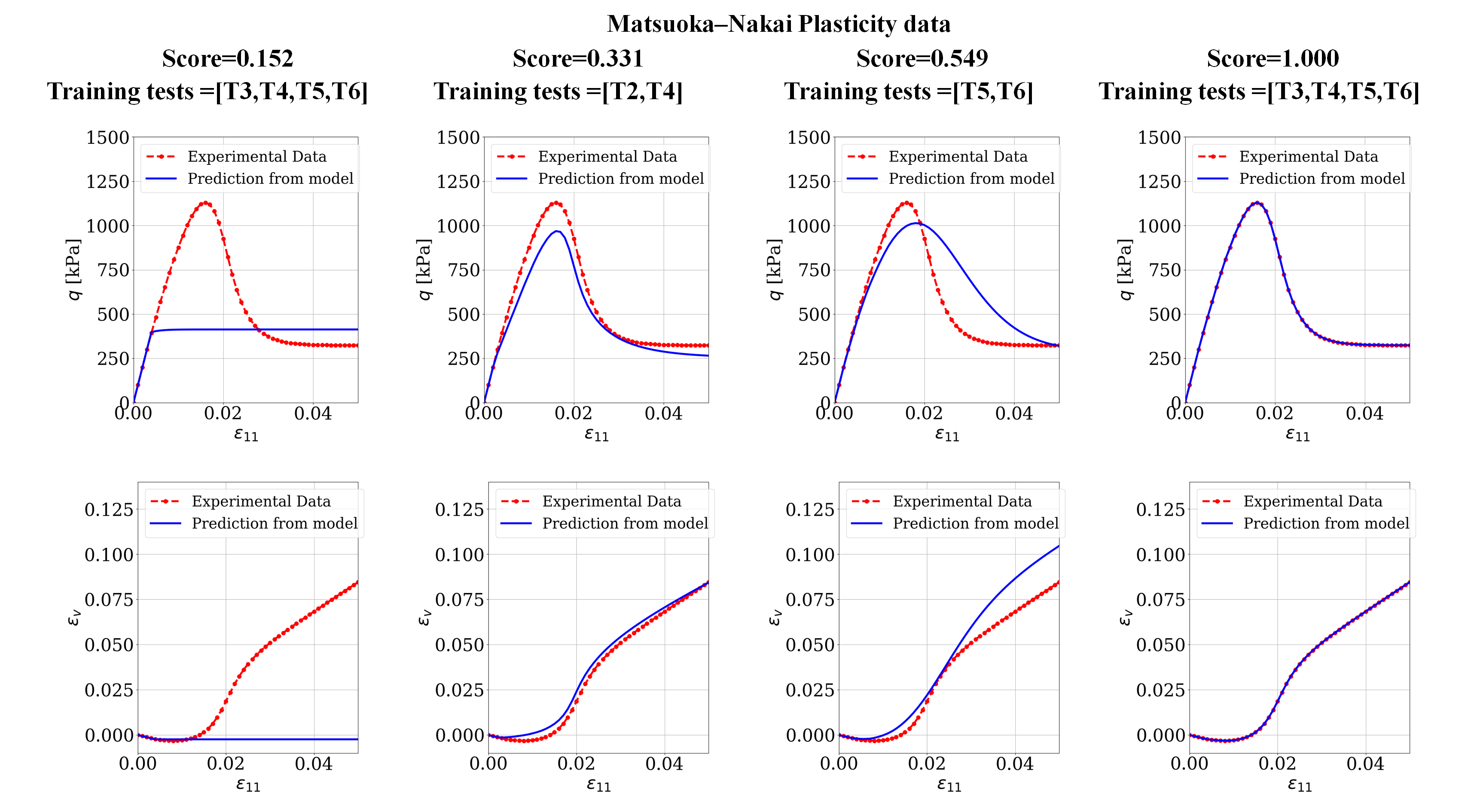}
	\caption{Knowledge of elasto-plastic models learned by deep reinforcement learning in Numerical Experiment 1 using synthetic data from Matsuoka–Nakai plasticity. Four representative games played during the DRL iterations and their prediction accuracy against synthetic data are presented.}
	\label{fig:example1_improve_MN}
\end{figure}

\begin{figure}[h!]\center
	\subfigure[J2 Plasticity]{
		\includegraphics[width=0.31\textwidth]{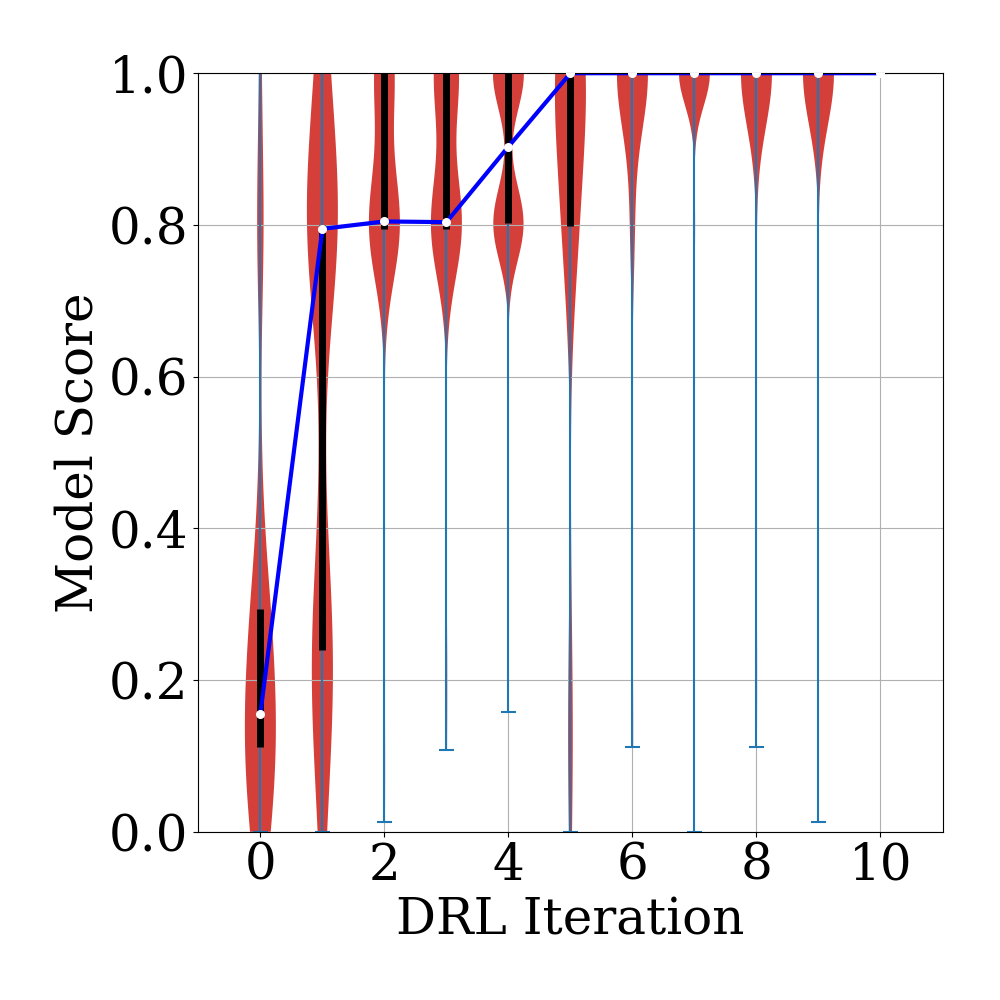}
	}
	\subfigure[Drucker–Prager Plasticity]{
		\includegraphics[width=0.31\textwidth]{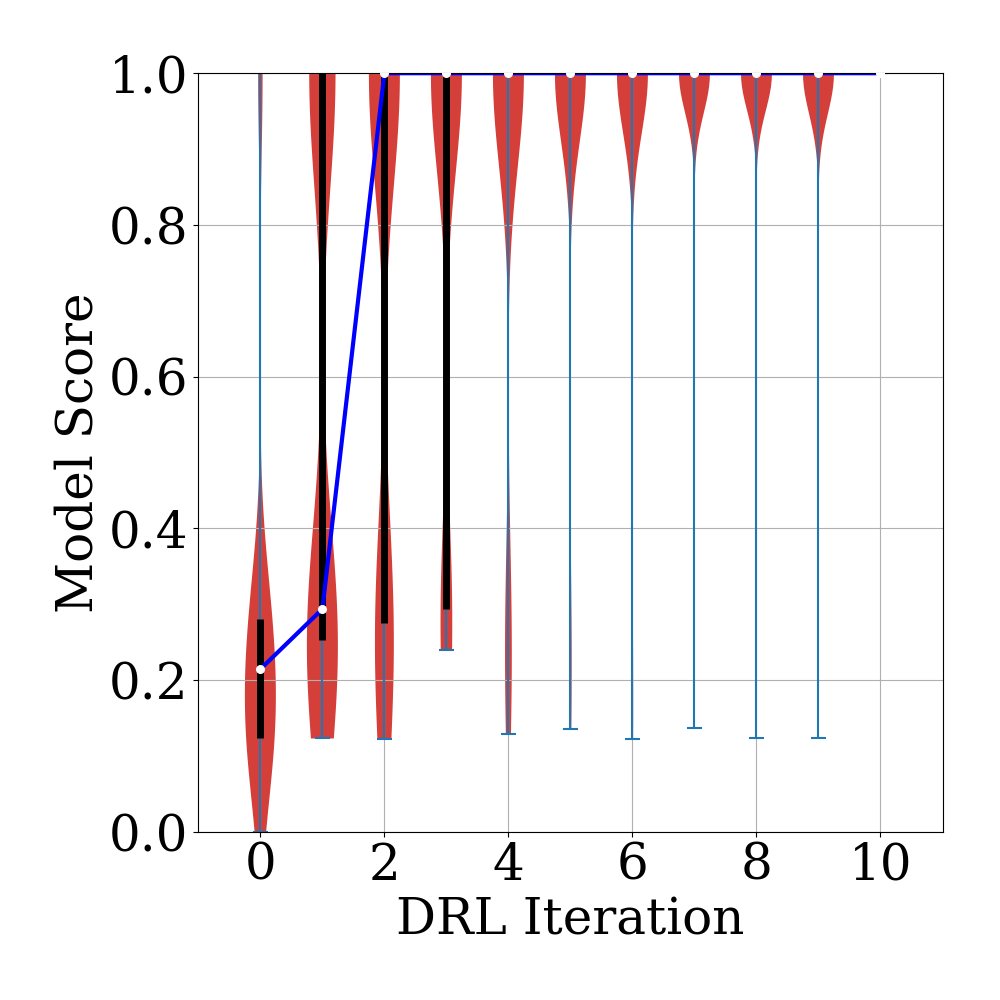}
	}
	\subfigure[Matsuoka–Nakai Plasticity]{
		\includegraphics[width=0.31\textwidth]{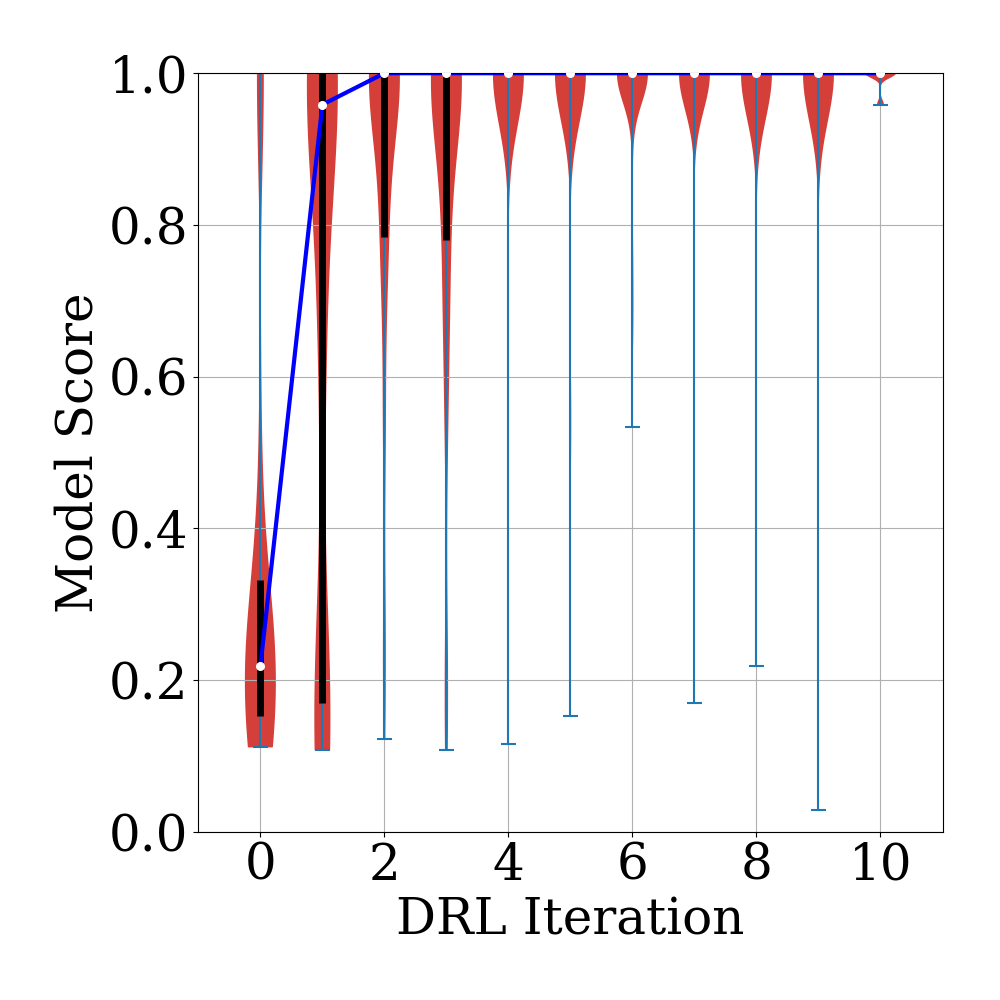}
	}
	\caption{Violin plots of the density distributions of model scores in each DRL iteration in Numerical Experiment 1. Statistics of the model scores in deep reinforcement learning iterations from 20 separate runs of the DRL procedure for Numerical Experiment 1. Each DRL procedure contains ten iterations 0-9 of "exploration and exploitation" (by setting the temperature parameter $\tau=1.0$) and a final iteration 10 of "competitive gameplay" ($\tau=0.01$). Each iteration consists of 30 games. The shaded area represents the density distribution of scores. The white point represents the median. The thick black bar represents the inter-quartile range between 25\% quantile and 75\% quantile. The maximum and minimum scores played in each iteration are marked by horizontal lines.}
	\label{fig:example1_learn_violinplot}
\end{figure}

\begin{figure}[h!]\center
	\subfigure[J2 Plasticity]{
		\includegraphics[width=0.31\textwidth]{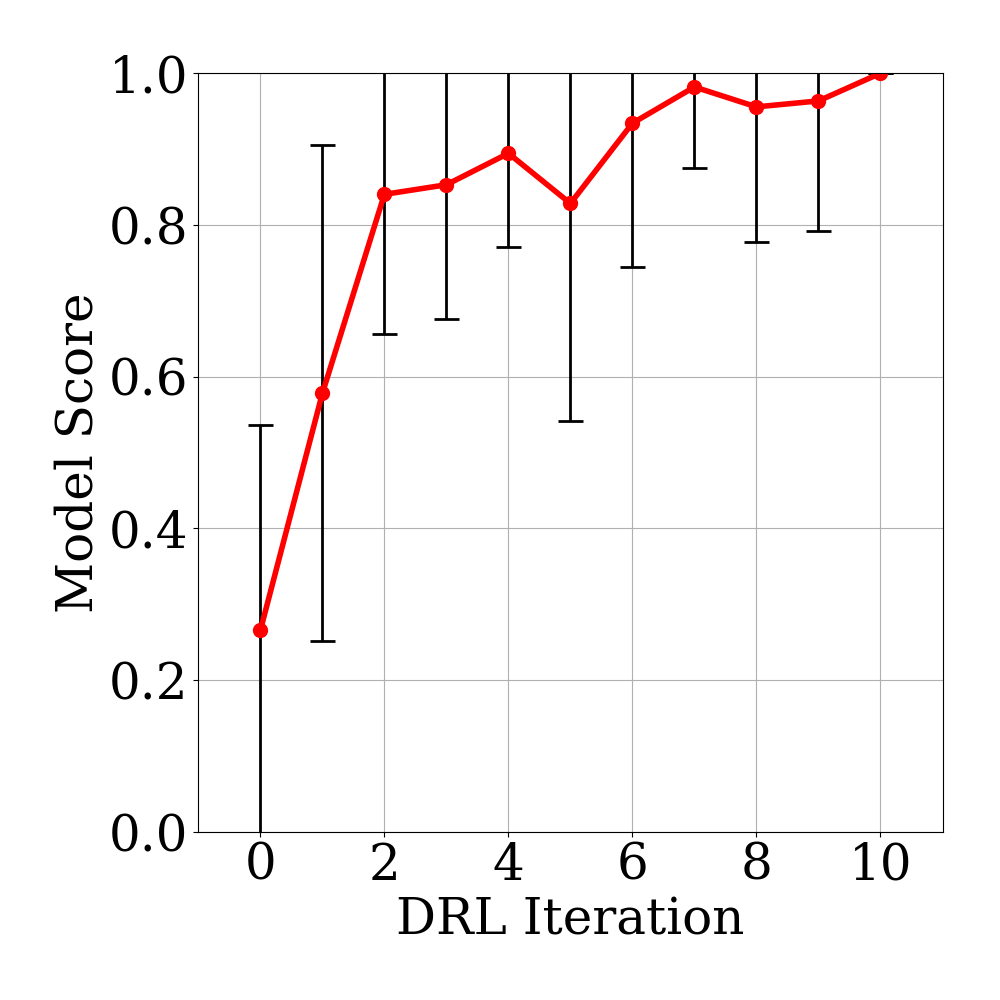}
	}
	\subfigure[Drucker–Prager Plasticity]{
		\includegraphics[width=0.31\textwidth]{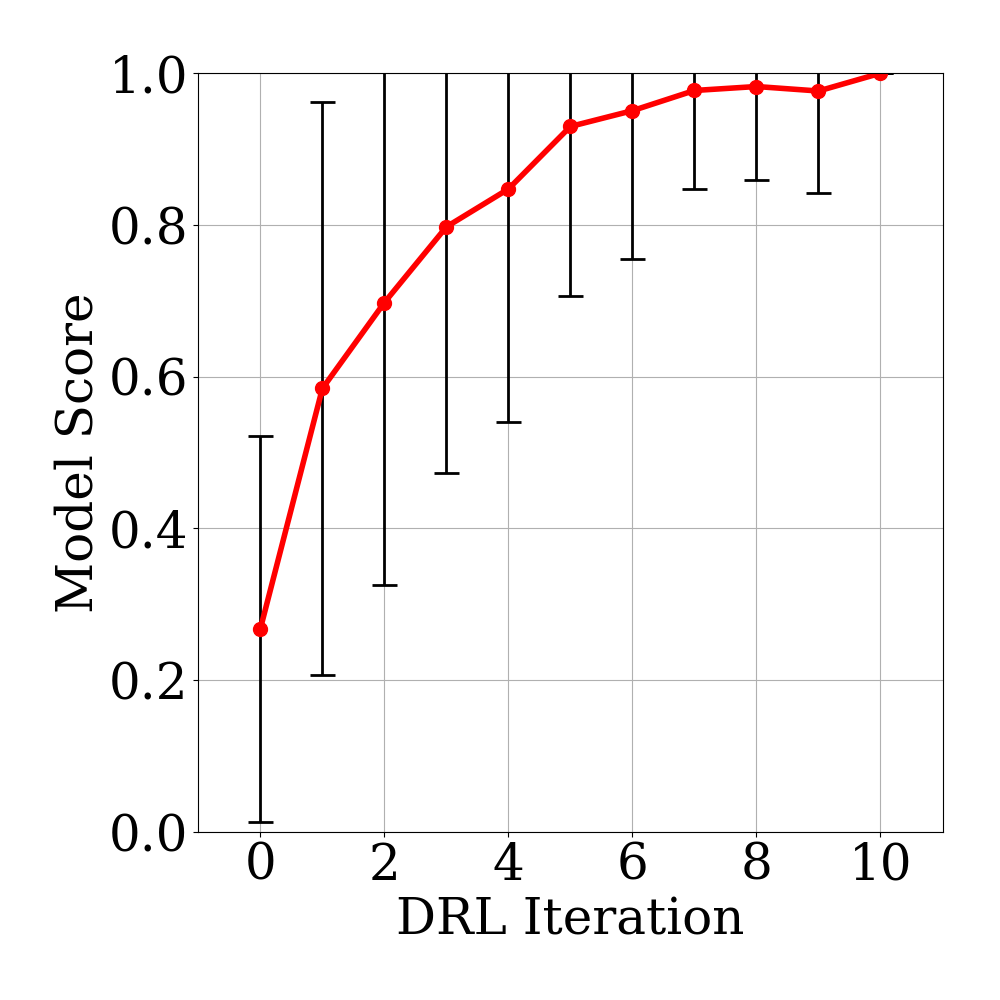}
	}
	\subfigure[Matsuoka–Nakai Plasticity]{
		\includegraphics[width=0.31\textwidth]{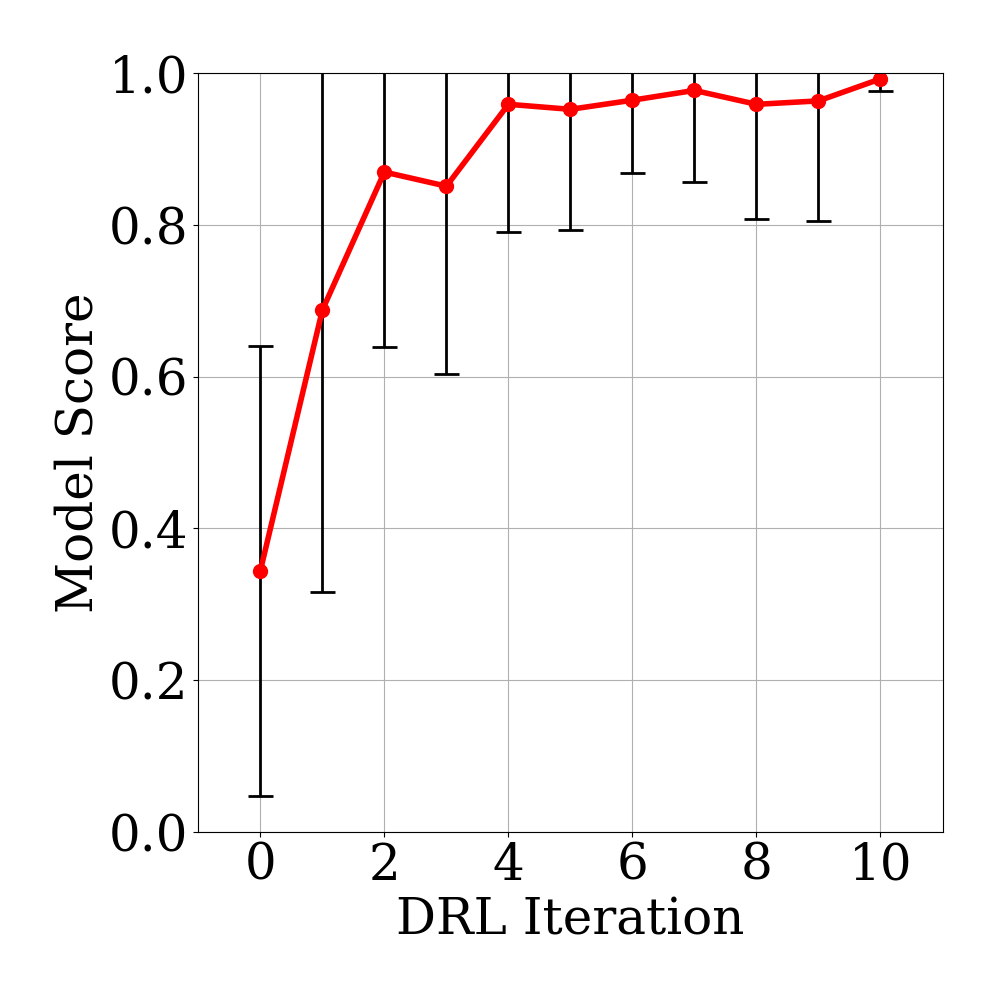}
	}
	\caption{Mean value (red dots) and $\pm$ standard deviation (error bars) of model score in each DRL iteration in Numerical Experiment 1.}
	\label{fig:example1_learn_errorbar}
\end{figure}

Figures \ref{fig:example1_improve_J2}, \ref{fig:example1_improve_DP} and \ref{fig:example1_improve_MN} present the example model predictions and calibration tests during the DRL improvement of the experimentalist and modeler agents. 
Both agents try out different combinations of calibration data and model choices, and evaluate their model scores and individual game rewards. 
The agents learn from all the gameplay results that they have experienced and converge their individual strategies to the optimal ones that eventually generate the optimal set of experiment tests for model calibration and exactly recover the plasticity model used to generate the synthetic data. 
The "cooperative" convergence of the strategies of both agents is of crucial importance, since the calibration dataset and the selected model must be simultaneously optimal for the final model score to be maximum. 
Although the gameplays could be different in each separate run of the two-player DRL algorithm due to the randomness in initial Policy/Value networks and the action possibilities involved in Monte Carlo Tree Search, the optimal strategies are always recovered if the exploration is sufficient. This is confirmed in Figure \ref{fig:example1_learn_violinplot} and \ref{fig:example1_learn_errorbar} in which the statistics of the gameplay scores of 20 separate executions of the two-player DRL algorithm are analyzed. 

The fact that the two-player meta-modeling game is able to reach a perfect score in blind prediction indicates that it 
has successfully reverse engineered the constitutive law. The ability to automatically reverse-engineering a constitutive model could be of potential commercial value, as it allows one to understand attributes of legacy or proprietary software even when only the executable is available. Even in the case when reverse engineering fails to recover the constitutive responses perfectly, 
the score can indicate how close the DRL-generated model replicates the constitutive law in the legacy or proprietary codes. 

Furthermore, the fact that the training is able to recover the model also enables us to use a different 
architecture for computational mechanics software in which the material model library does not necessarily contain 
multiple constitutive laws categorized by labels or model names. Instead, any new model in the literature that contains 
new "action" not available in the previous constitutive law can be decomposed into directed graphs and subsequently 
be merged with the existing pool of actions such that the modeler agent can have more tools to
generate new models that optimize objective functions. Since (1) new actions that complete the model will only be picked 
by the modeler agent if they can help it achieve a higher score,and (2) should this happens, the improvement in prediction 
quality is quantified by the increase in the score, the meta-modeling game can be used as a tool to evaluate 
the true benefit of any new action that departs from the state-of-the-art. 

\subsection{Numerical Experiments 2: Testing the ability of AI for forward predictions} \label{sec:ne2}
In this numerical experiment, we examine the ability of the proposed meta-modeling agents
to (1) generate the knowledge and model represented by the directed graph from given data, 
(2) decide the set of experiments that aids data-driven discovery and (3) terminate 
the learning process when further experiments no longer benefit predictions. 

In this test, we consider an idealized situation in which the data is generated from discrete element simulations
 for granular materials \citep{cundall1979discrete, kuhn2015stress, wang2019updated}. 
 While the constitutive responses from the discrete element simulations may contain 
 fluctuation, we do not introduce any contaminated noise on purpose to test how the meta-modeling 
 procedure  might be affected by noise. While this could be addressed using dropout layers as shown in \citet{wang2018multiscale}, 
 a comprehensive study on learning with noisy data  is out of the scope in this study but will be considered in the future. 
The data for calibration and evaluation of prediction accuracy of the deep-reinforcement-learned constitutive models are generated by numerical simulations on a representative volume element (RVE) of densely-packed spherical DEM particles. 
The open-source discrete element simulation software YADE for DEM is used by the experimentalist agent to generate 
data, including the homogenized stress and strain measures and the geometrical and microstructural attributes such as 
coordination number, fabric tensor, porosity
 \citep{vsmilauer2010yade, sun2014micromechanical}. 
The discrete element particles in the RVE have radii between $1 \pm 0.3$ mm with a uniform distribution. 
The Cundall's elastic-frictional contact model (\citep{cundall1979discrete}) is used for the inter-particle constitutive law. 
The material parameters are: interparticle elastic modulus $E_{eq}=0.5$ GPa, ratio between shear and normal stiffness $k_s/k_n=0.3$, frictional angle $\varphi=$ \ang{30}, density $\rho=2600$ $kg/m^3$, Cundall damping coefficient $\alpha_{damp}=0.6$. 

The test data constitute of triaxial tests on DEM samples with different initial confining pressure and void ratio $\dot{\sigma}_{33}=\dot{\sigma}_{12}=\dot{\sigma}_{23}=\dot{\sigma}_{13}=0, b = \frac{\sigma_{22}-\sigma_{33}}{\sigma_{11}-\sigma_{33}}$. 
\begin{description}
	\item[T1:] $\dot{\epsilon}_{11}<0$, $b=0$, $p_0 = -300 kPa$, $e_0 = 0.539$.
	\item[T2:] $\dot{\epsilon}_{11}<0$, $b=0$, $p_0 = -400 kPa$, $e_0 = 0.536$.
	\item[T3:] $\dot{\epsilon}_{11}<0$, $b=0$, $p_0 = -500 kPa$, $e_0 = 0.534$.
	\item[T4:] $\dot{\epsilon}_{11}>0$, $b=0$, $p_0 = -300 kPa$, $e_0 = 0.539$.
	\item[T5:] $\dot{\epsilon}_{11}>0$, $b=0$, $p_0 = -400 kPa$, $e_0 = 0.536$.
	\item[T6:] $\dot{\epsilon}_{11}>0$, $b=0$, $p_0 = -500 kPa$, $e_0 = 0.534$.
	\item[T7:] $\dot{\epsilon}_{11}<0$, $b=0.5$, $p_0 = -300 kPa$, $e_0 = 0.539$.
	\item[T8:] $\dot{\epsilon}_{11}<0$, $b=0.5$, $p_0 = -400 kPa$, $e_0 = 0.536$.
	\item[T9:] $\dot{\epsilon}_{11}<0$, $b=0.5$, $p_0 = -500 kPa$, $e_0 = 0.534$.
	\item[T10:] $\dot{\epsilon}_{11}>0$, $b=0.5$, $p_0 = -300 kPa$, $e_0 = 0.539$.
	\item[T11:] $\dot{\epsilon}_{11}>0$, $b=0.5$, $p_0 = -400 kPa$, $e_0 = 0.536$.
	\item[T12:] $\dot{\epsilon}_{11}>0$, $b=0.5$, $p_0 = -500 kPa$, $e_0 = 0.534$.
	\item[T13:] $\dot{\epsilon}_{11}<0$, $b=0.1$, $p_0 = -300 kPa$, $e_0 = 0.539$.
	\item[T14:] $\dot{\epsilon}_{11}<0$, $b=0.1$, $p_0 = -400 kPa$, $e_0 = 0.536$.
	\item[T15:] $\dot{\epsilon}_{11}<0$, $b=0.1$, $p_0 = -500 kPa$, $e_0 = 0.534$.
	\item[T16:] $\dot{\epsilon}_{11}>0$, $b=0.1$, $p_0 = -300 kPa$, $e_0 = 0.539$.
	\item[T17:] $\dot{\epsilon}_{11}>0$, $b=0.1$, $p_0 = -400 kPa$, $e_0 = 0.536$.
	\item[T18:] $\dot{\epsilon}_{11}>0$, $b=0.1$, $p_0 = -500 kPa$, $e_0 = 0.534$.
	\item[T19:] $\dot{\epsilon}_{11}<0$, $b=0.25$, $p_0 = -300 kPa$, $e_0 = 0.539$.
	\item[T20:] $\dot{\epsilon}_{11}<0$, $b=0.25$, $p_0 = -400 kPa$, $e_0 = 0.536$.
	\item[T21:] $\dot{\epsilon}_{11}<0$, $b=0.25$, $p_0 = -500 kPa$, $e_0 = 0.534$.
	\item[T22:] $\dot{\epsilon}_{11}>0$, $b=0.25$, $p_0 = -300 kPa$, $e_0 = 0.539$.
	\item[T23:] $\dot{\epsilon}_{11}>0$, $b=0.25$, $p_0 = -400 kPa$, $e_0 = 0.536$.
	\item[T24:] $\dot{\epsilon}_{11}>0$, $b=0.25$, $p_0 = -500 kPa$, $e_0 = 0.534$.
	\item[T25:] $\dot{\epsilon}_{11}<0$, $b=0.75$, $p_0 = -300 kPa$, $e_0 = 0.539$.
	\item[T26:] $\dot{\epsilon}_{11}<0$, $b=0.75$, $p_0 = -400 kPa$, $e_0 = 0.536$.
	\item[T27:] $\dot{\epsilon}_{11}<0$, $b=0.75$, $p_0 = -500 kPa$, $e_0 = 0.534$.
	\item[T28:] $\dot{\epsilon}_{11}>0$, $b=0.75$, $p_0 = -300 kPa$, $e_0 = 0.539$.
	\item[T29:] $\dot{\epsilon}_{11}>0$, $b=0.75$, $p_0 = -400 kPa$, $e_0 = 0.536$.
	\item[T30:] $\dot{\epsilon}_{11}>0$, $b=0.75$, $p_0 = -500 kPa$, $e_0 = 0.534$.
\end{description}

\begin{figure}[h!]\center
	\subfigure[Violin plots of the density distribution of model scores in each DRL iteration]{
		\includegraphics[width=0.45\textwidth]{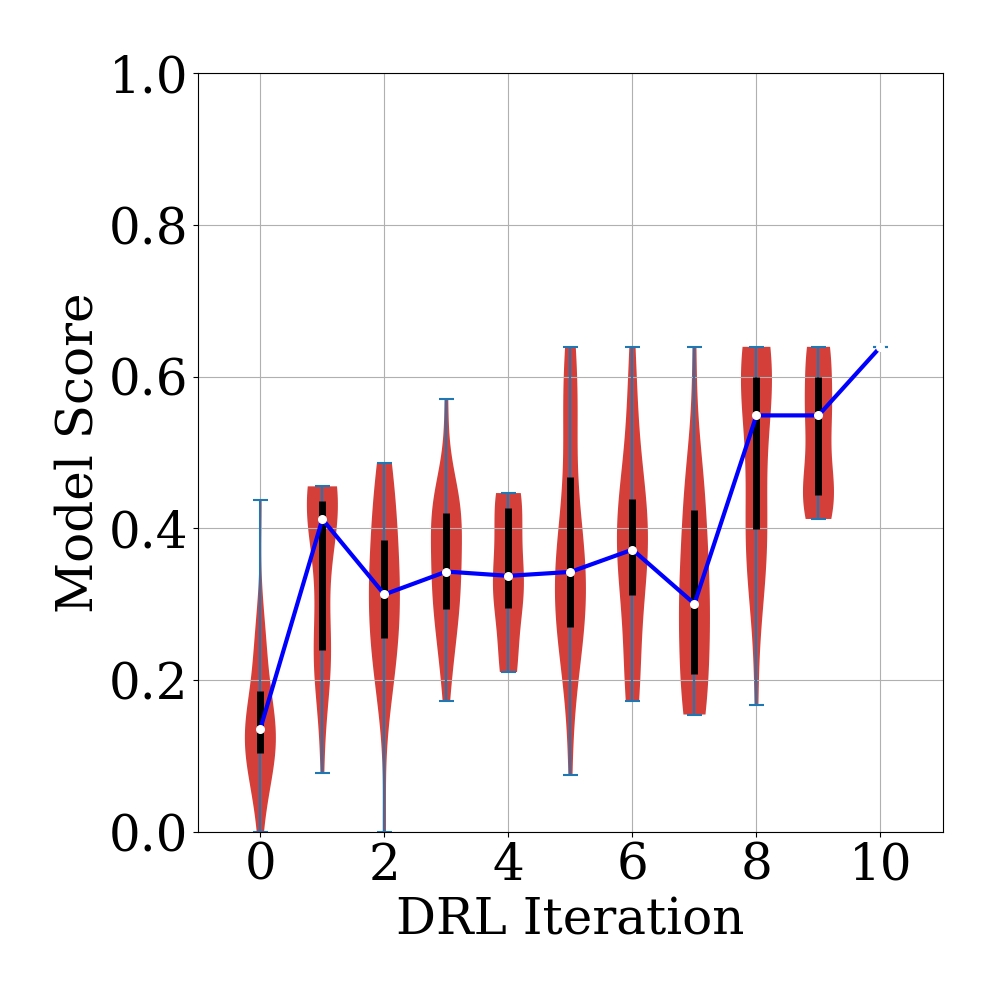}
	}
	\subfigure[Mean value and $\pm$ standard deviation of model score in each DRL iteration]{
		\includegraphics[width=0.45\textwidth]{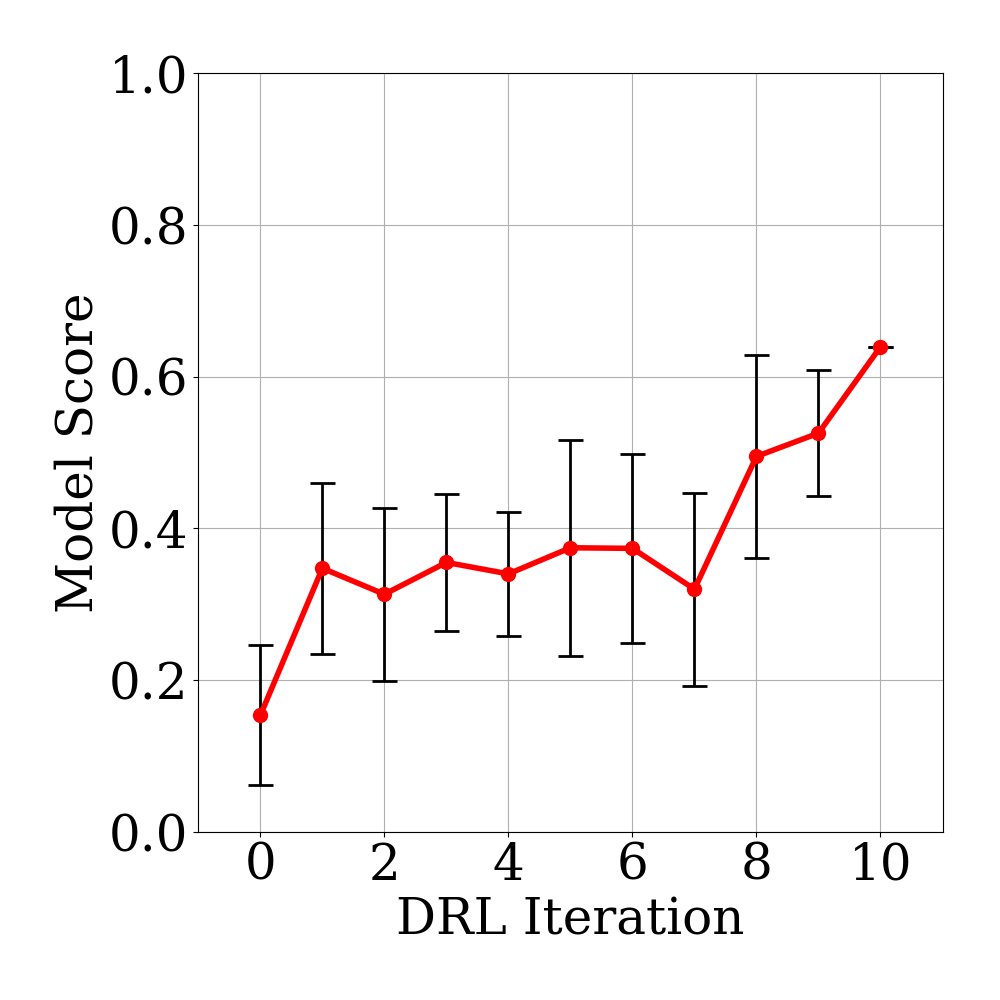}
	}
	\caption{Statistics of the model scores in deep reinforcement learning iterations from 5 separate runs of the DRL procedure for Numerical Experiment 2. Each DRL procedure contains ten iterations 0-9 of "exploration and exploitation" (by setting the temperature parameter $\tau=1.0$) and a final iteration 10 of "competitive gameplay" ($\tau=0.01$). Each iteration consists of 30 games. (a) Violin Plot of model scores played in each DRL iteration. The shaded area represents the density distribution of scores. The white point represents the median. The thick black bar represents the interquartile range between 25\% quantile and 75\% quantile. The maximum and minimum scores played in each iteration are marked by horizontal lines. (b) Mean model score in each iteration and the error bars mark $\pm$ standard deviation.}
	\label{fig:game2_statistics}
\end{figure}

The candidate tests for the calibration data generation include $\textbf{T}_c^0 = \{T1,T2,T3,...,T11,T12\}$ and the validation tests are $\textbf{T}_v^0 = \{T13,T14,T15,...,T19,T30\}$. 
As explained in Section \ref{subsec:datagame}, the tests not selected in the final calibration set by the experimentalist agent will be moved to the final validation set to evaluate the blind prediction performance. 
The parameters for the DRL procedure are identical to the settings in Example 1.
The statistics of the gameplay results from 5 separate runs of the DRL procedure are presented in Figure \ref{fig:game2_statistics}. 
We observe efficient improvements in the generated elasto-plastic models over the DRL training iterations with the discrete element simulation data. 

\begin{figure}[h!]\center
	\includegraphics[width=0.95\textwidth]{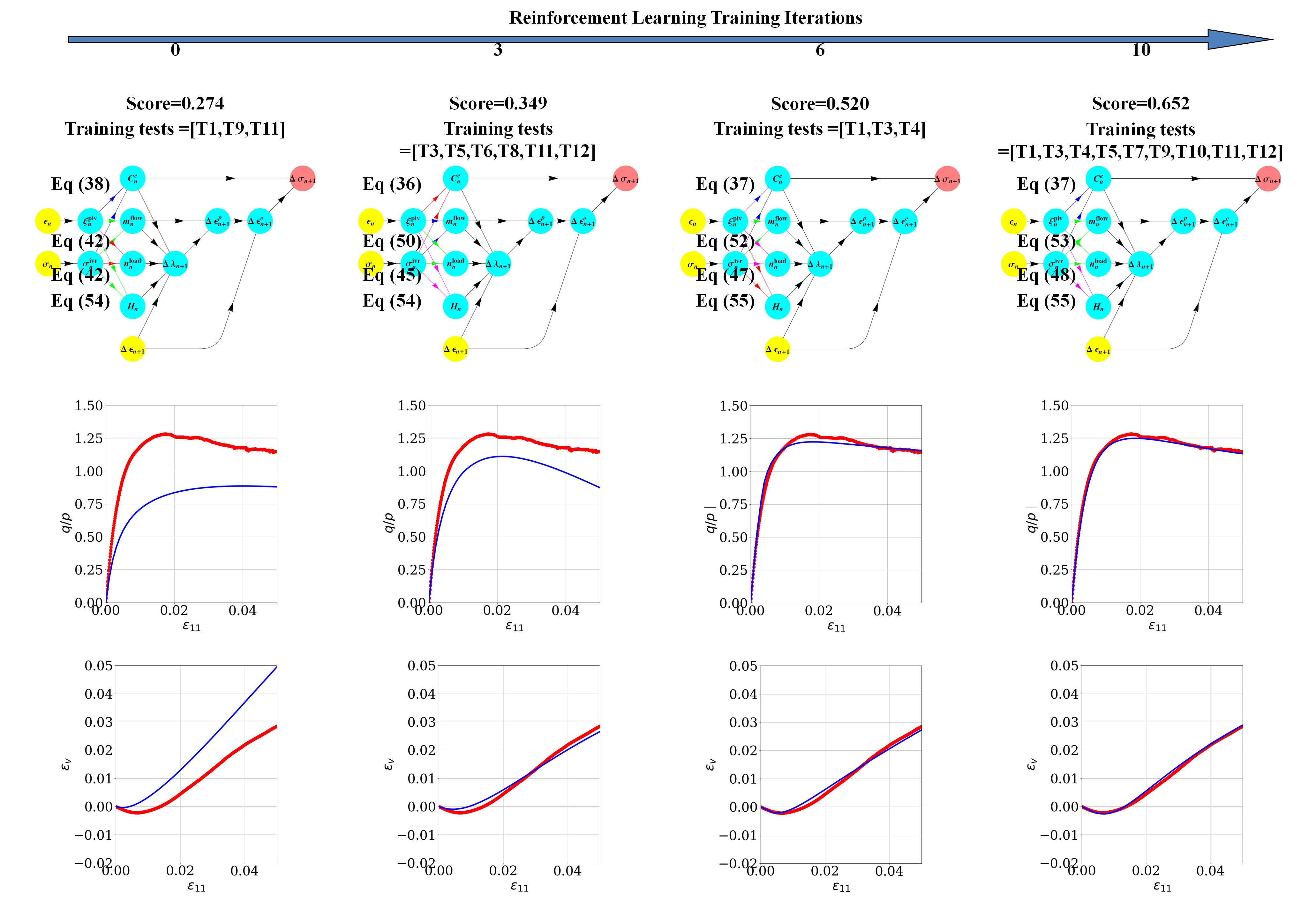}
	\caption{Knowledge of elasto-plastic models learned by deep reinforcement learning in Numerical Experiment 2 using data from drained triaxial tests. Four representative games played during the DRL iterations and their prediction accuracy against synthetic data are presented. The color edges illustrate different labeled edges selected in the constitutive models. The labels are represented by equation numbers and these equations are detailed in \textit{Game Choices} in Section \ref{subsec:modelgame}.}
	\label{fig:Test2_Improve}
\end{figure}

\begin{figure}[h!]\center
	\includegraphics[width=0.95\textwidth]{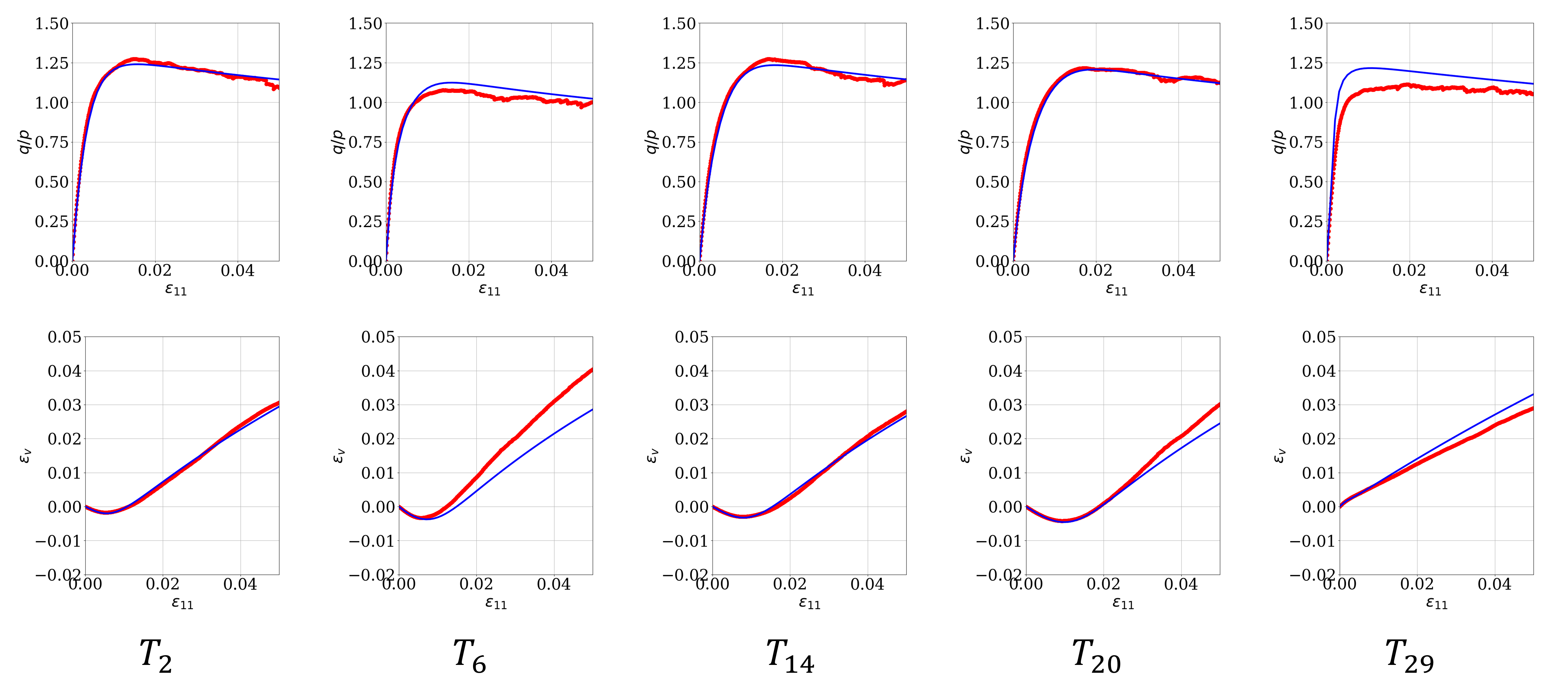}
	\caption{Five examples of blind predictions from the optimal digraph configuration (The 4th digraph in Figure \ref{fig:Test2_Improve}) against data from the tests.}
	\label{fig:Example2_Predict}
\end{figure}

Figure \ref{fig:Test2_Improve} presents the example model predictions and calibration tests 
during the DRL improvement of the experimentalist and modeler agents. 
The final converged calibration test set chosen by the AI experimentalist after the DRL 
procedure consists of the triaxial extension and compression tests with $b=0$ and $b=0.5$ 
under initial pressures of -300 kPa and -500 kPa.
Accordingly, the final converged elasto-plastic model generated by the AI modeler after the DRL
 procedure is composed of the non-linear elasticity of Eq. (\ref{eq:elast_2}), the loading direction defined as Eq. (\ref{eq:nload_1}), the plastic flow direction defined as Eq. (\ref{eq:mflow_1}), and the hardening modulus
 defined as Eq. (\ref{eq:hhard_1}). The resultant model is a generalized plasticity model 
 (without explicitly defined yield surface and plastic potential) combined with the critical state
 plasticity theory (dependence on the $p,q,\theta$ stress invariants and the void ratio $e$). 
 Figure \ref{fig:Example2_Predict} presents five representative examples of blind predictions
 of this selected model and the selected calibration data. This optimal model for the given action space
 is generated from data obtained from 9 experiments in the following order: [T1, T3, T4, T5, T7, T9, T10, T11, T12]. 

One interesting aspect revealed in this numerical experiment is the potential of using the meta-modeling game 
as a tool to evaluate and analyze of relative policy values of the ingredients of constitutive laws  in a prediction task. 
For instance, this numerical experiment reveals that the optimal configuration of the constitutive model for predicting the behavior of monotonic loading triaxial compression test should not contain any neural network edge
 (Eq. (\ref{eq:rnn_1}), (\ref{eq:rnn_2}) in Section \ref{subsec:modelgame}) 
This could be attributed to the 
facts that the training data of the loading directions, plastic flow directions and hardening moduli from the DEM
experimental data contain high-frequency fluctuations and that our testing data, which are used 
to evaluate the forward prediction performance, contain only monotonic stress paths. 
Since the high-frequency fluctuation makes the neural network easily to exhibit overfitting responses, 
and the relatively simple stress paths make it less advantageous to use a high-dimensional universal approximator 
like a neural network in any component of the constitutive models, 
the edges that map input from the output vertices through mathematical expressions 
are revealed to have higher policy values as the game progresses and ultimately become the 
selected models.  

Note that this result is in sharp contrast with the meta-modeling game results of the traction-separation law 
in which the neural network edges become dominant and yield a consistently good forward predictions \citep{wang2018multiscale, wang2019meta}.  
Comparing the choices the agents made in the two games reveal that the autonomous agents are capable of 
adjusting their decisions based on the availability of the data and the type of the forward prediction tasks. In other words, the agents are able to make judgments such that it employs 
edges that contain low-dimensional mathematical expression when the regularization (avoiding the curse of high dimensionality) is more critical than 
high-dimensionality afforded by the large numbers of neural network nodes (in this case), but also able to 
select the high-dimensional neural network options when the advantages of the options outweigh the drawbacks 
(in the traction-separation law game in \citep{wang2019meta}). 
Note that this optimal configuration sought by the meta-modeling game is sensitive to the available actions. 
 For instance, the improvements of the neural network
 could be achieved by introducing techniques to filter out the 
noisy data and employing advanced neural networks with noise-resistant architecture \citep{sun2018stochastic}. 
These changes can impose adjustments in the policy values and therefore affect the optimal configuration. 
The incorporation of de-noising mechanisms and the investigation of the influence of data quality
on the meta-modeling game framework will be conducted in the future study. 

This \textbf{automated} strategy change by the AI agents is significant as it demonstrates that the agent system is able to adapt to the environment (availability of calibration data and the types of testing data) to make rational choices like a human modeler should when given different prediction tasks of different complexities.  

Another important implication of the meta-modeling game is its ability to quantitatively analyze the performance of families of models currently (or historically, if possible to be inferred from reverse engineering) available in the literature for an intended prediction task. 
Table \ref{tab:ScoreDistributions} shows the post-game analysis of the performance of the 112 models automatically generated from the two-player game. The resultant models are grouped into five different classes based on the types of the edges used  in the game. The interesting aspect of the data in Table \ref{tab:ScoreDistributions} is that it 
provides users a quantitative measure that configurations based on generalized plasticity and critical state outperform 
all the other 90 configurations. This result is consistent with the convention understanding from soil mechanics in which 
 the classical critical state plasticity theory and the resultant plastic dilatancy/contraction predictions is  regarded as the key ingredient for predictive constitutive laws.  
Examinations on models in Class 1 also reveals that three-invariant generalized plasticity with critical state perform the best in the blind predictions, especially when the material states of the granular materials in the calibration tests (e.g. confining pressure, initial void ratio, stress path) are significantly different than the ones in blind predictions.  

\begin{table}[t]
	\centering
	\begin{tabular}{| p{1.1cm} | p{1.5cm} | p{1cm} | p{1.5cm} | p{1.8cm} | p{1cm} | p{2.8cm} | p{1cm} |}
		\hline
		Model Class & Number of Models & Mean Score & Standard deviation & Generalized Plasticity 'GP' & Critical State 'CS' & Classical pressure dependent elasto-plasticity 'DP' & Others 'O'\\ \hline
		1 & 22 & 0.603 & 0.054 & \checkmark & \checkmark & & \\ \hline
		2 & 25 & 0.565 & 0.051 & \checkmark & & & \\ \hline
		3 & 13 & 0.295 & 0.028 & & \checkmark &\checkmark  & \\ \hline
		4 & 19 & 0.450 & 0.086 & & & \checkmark & \\ \hline
		5 & 33 & 0.163 & 0.063 & & & & \checkmark\\ \hline
	\end{tabular}
	\caption{Five classes of the constitutive models generated during the deep reinforcement learning.}
	\label{tab:ScoreDistributions}
\end{table}

\begin{figure}[h!]\center
	\includegraphics[width=0.55\textwidth]{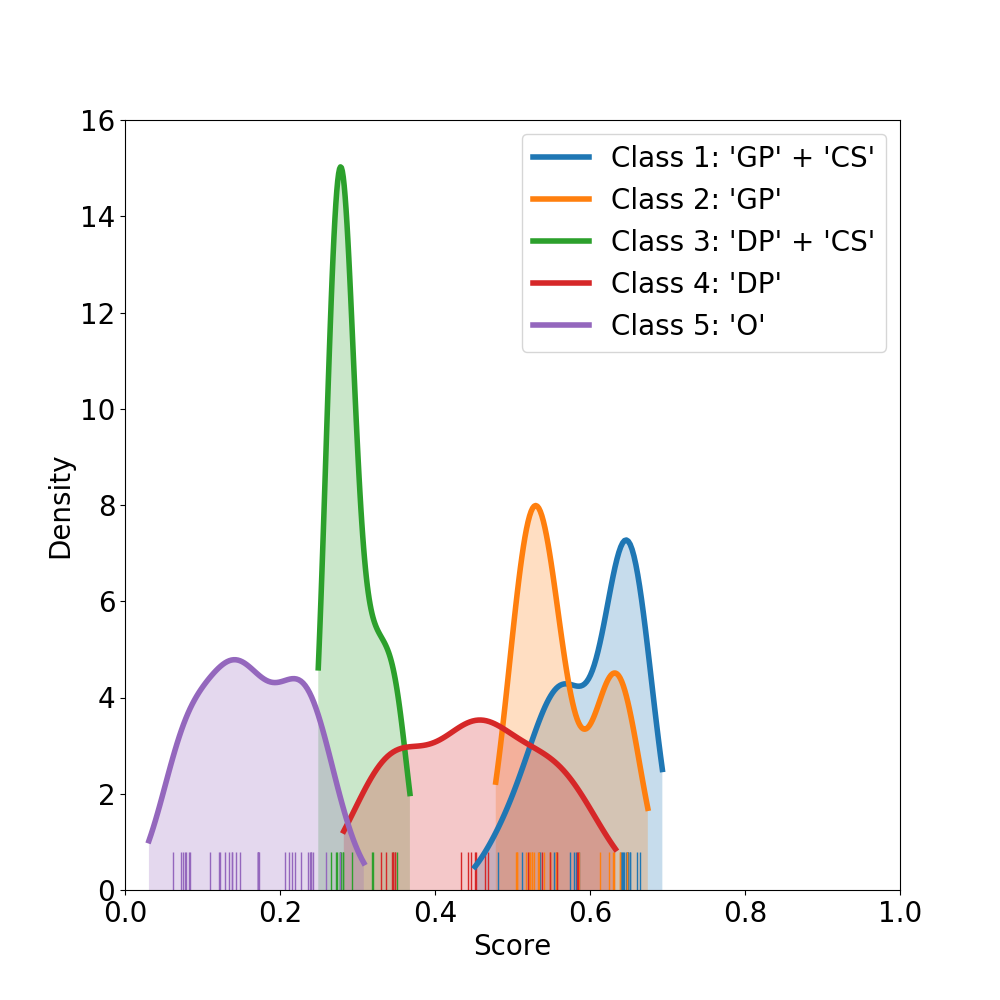}
	\caption{Distribution of the scores of the models generated during the deep reinforcement learning. The models are grouped into five families (see Table \ref{tab:ScoreDistributions}). The curves present the Gaussian kernel density estimation of the model score distributions (The estimated function $f_h(x) = \frac{1}{nh} \sum_{i=1}^n K(\frac{x-x_i}{h})$ for score data $(x_1, x_2, ..., x_n)$, $K(x)$ is the standard normal distribution function, $h$ is the bandwidth parameter determined by Scott's rule $h = \frac{3.5 \hat{\sigma}}{n^{1/3}}$, where $\hat{\sigma}$ is the standard deviation).}
	\label{fig:ScoreDistributions}
\end{figure}

However, comparisons of the results in Classes 1, 2,3 and 4 shown in Figure \ref{fig:ScoreDistributions} reveal 
a somewhat surprising conclusion in which 
the generalized plasticity seems to be consistently the more important ingredient than the critical state theory 
for yielding predictive models. Although it is important to stress that this conclusion must be interpreted 
with respect to the types and amount of data available and the intended prediction task, 
this result does provide further evidence to support the speculations that 
the generalized plasticity, if calibrated properly, does likely to improve the accuracy 
of blind predictions of the responses of granular materials in  the monotonic triaxial compression tests \citep{zienkiewicz1984generalized, pastor1990generalized, ling2003pressure, sanchez2005double}.


In conclusion, this numerical experiment shows that the meta-modeling game can provide three important types of  knowledge, the knowledge on the hierarchy of information flow, the estimation on the amount of data required to reach the state-of-the-art performance for a given action space and specified objective, and the relative values/benefits/importance of each model/theoretical/data-driven components revealed in the post-game analysis. 

Remark. Note that applying the meta-modeling game  to predicting responses of granular materials under different water drainage conditions may likely yield a very different conclusion where machine learning edges could be more widely used in the optimal configuration. This is because of the lack of a constitutive model that can quantitatively capture the constitutive responses of granular materials in drained and undrained conditions \citep{gens1988critical, manzari1997critical, zienkiewicz1999computational, pestana2002evaluation, sun2013unified}. 
The creation of models for more generic purposes and the estimation of the trusted range of application are both important issues, which  will be considered in future studies but are out of the scope of this paper.

%

\subsection{Numerical Experiment 3: AI-generated material models in  finite element simulations}
\begin{figure}[h!]\center
	\includegraphics[width=0.4\textwidth]{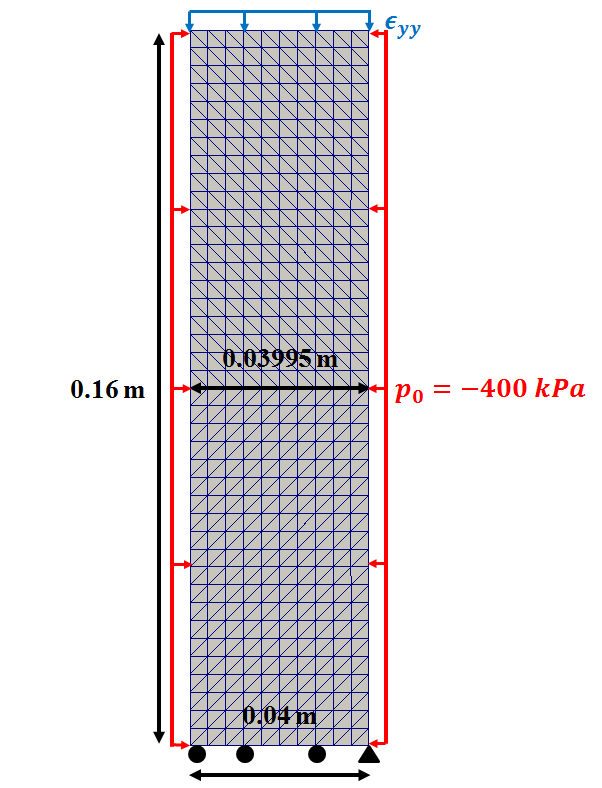}
	\caption{Description of the geometry, mesh, boundary, and loading conditions of the plane strain compression
		problem.}
	\label{fig:example3_mesh}
\end{figure}

To demonstrate the applicability of the AI-generated models from the plays of the data collection/meta-modeling game presented in Numerical Experiment 2, we conduct finite element simulations of a plane strain compression test on a rectangular specimen. The geometry, mesh and the boundary conditions of the simulations are given in Figure \ref{fig:example3_mesh}. 
The specimen is initially consolidated to isotropic pressure of $p_0$ =-400kPa, and have a uniform initial void ratio of 0.536. 
The specimen is compressed from the top surface, while the constant pressure $p_0$ are maintained on the lateral surfaces. 
Slight imperfection is introduced at the middle of the specimen to trigger heterogeneous deformation and shear bands. 
Three simulations are performed with the material properties given by the three example models generated during the DRL in Numerical Experiment 2 (1st, 3rd and 4th digraphs in Figure \ref{fig:Test2_Improve}). 

\begin{figure}[h!]\center
	\subfigure[Model 1]{
		\includegraphics[width=0.23\textwidth]{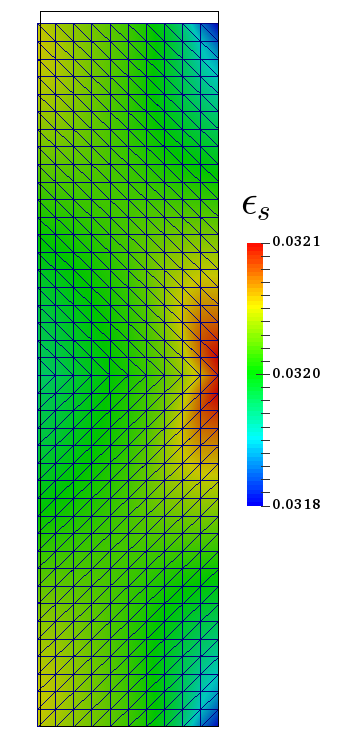}
	}
	\subfigure[Model 2]{
		\includegraphics[width=0.23\textwidth]{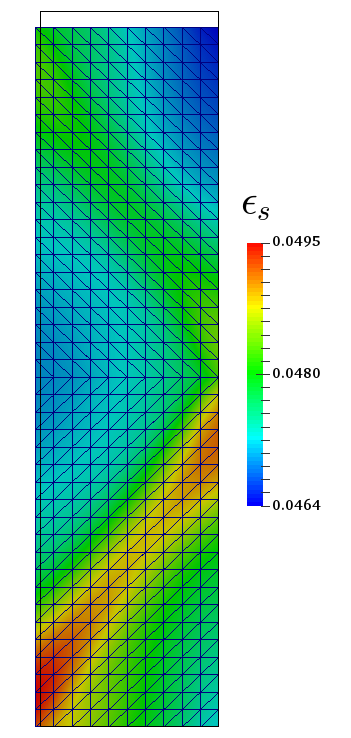}
	}
	\subfigure[Model 3]{
		\includegraphics[width=0.225\textwidth]{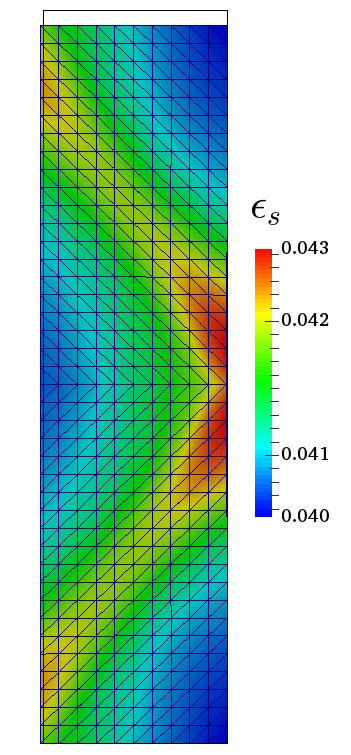}
	}
	\subfigure[Model 4]{
	\includegraphics[width=0.23\textwidth]{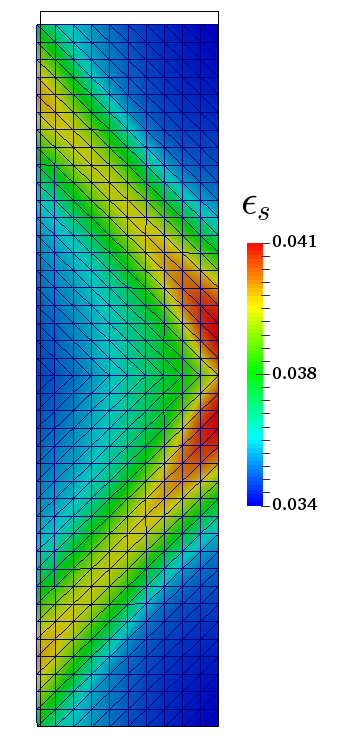}
	}
	\caption{Distribution of local deviatoric strain $\epsilon_s$ within the specimen at the end of the plane strain compression loadings. The finite element solutions using three models generated during the deep reinforcement learning of the meta-modeling game are compared (Model 1 is generated in the 1st DRL iteration in Figure \ref{fig:Test2_Improve}, Model 2 in the 5th iteration,  Model 3 in the 8th iteration, Model 4 in the 10th iteration).}
	\label{fig:example3_eps_s}
\end{figure}

\begin{figure}[h!]\center
	\subfigure[Model 1]{
		\includegraphics[width=0.23\textwidth]{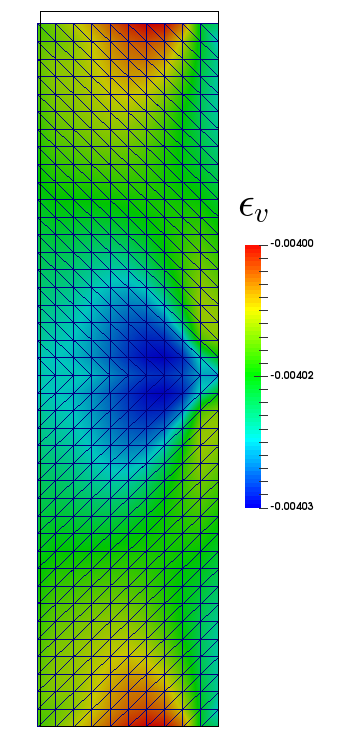}
	}
	\subfigure[Model 2]{
		\includegraphics[width=0.23\textwidth]{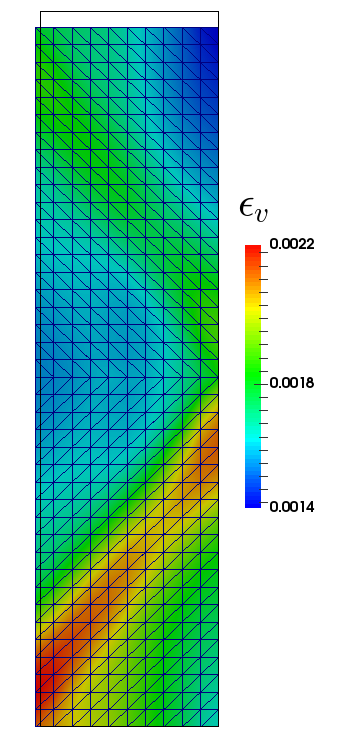}
	}
	\subfigure[Model 3]{
		\includegraphics[width=0.225\textwidth]{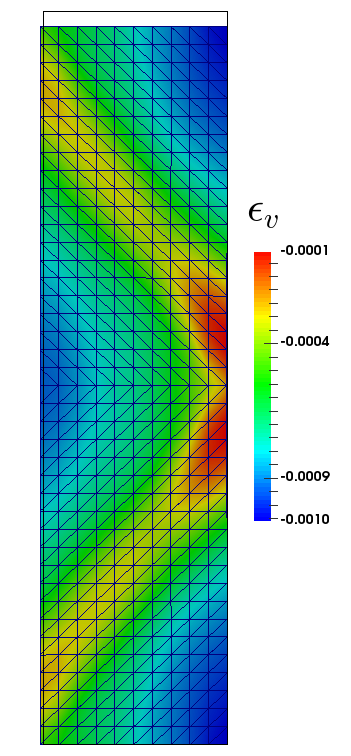}
	}
	\subfigure[Model 4]{
		\includegraphics[width=0.23\textwidth]{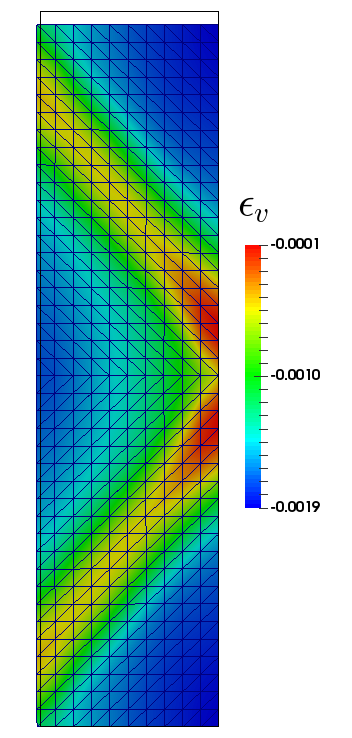}
	}
	\caption{Distribution of local volumetric strain $\epsilon_v$ within the specimen at the end of the plane strain compression loadings. The finite element solutions using three models generated during the deep reinforcement learning of the meta-modeling game are compared (Model 1 is generated in the 1st DRL iteration in Figure \ref{fig:Test2_Improve}, Model 2 in the 5th iteration,  Model 3 in the 8th iteration, Model 4 in the 10th iteration).}
	\label{fig:example3_eps_v}
\end{figure}

\begin{figure}[h!]\center
	\subfigure[Global diffrential stress $q$]{
		\includegraphics[width=0.45\textwidth]{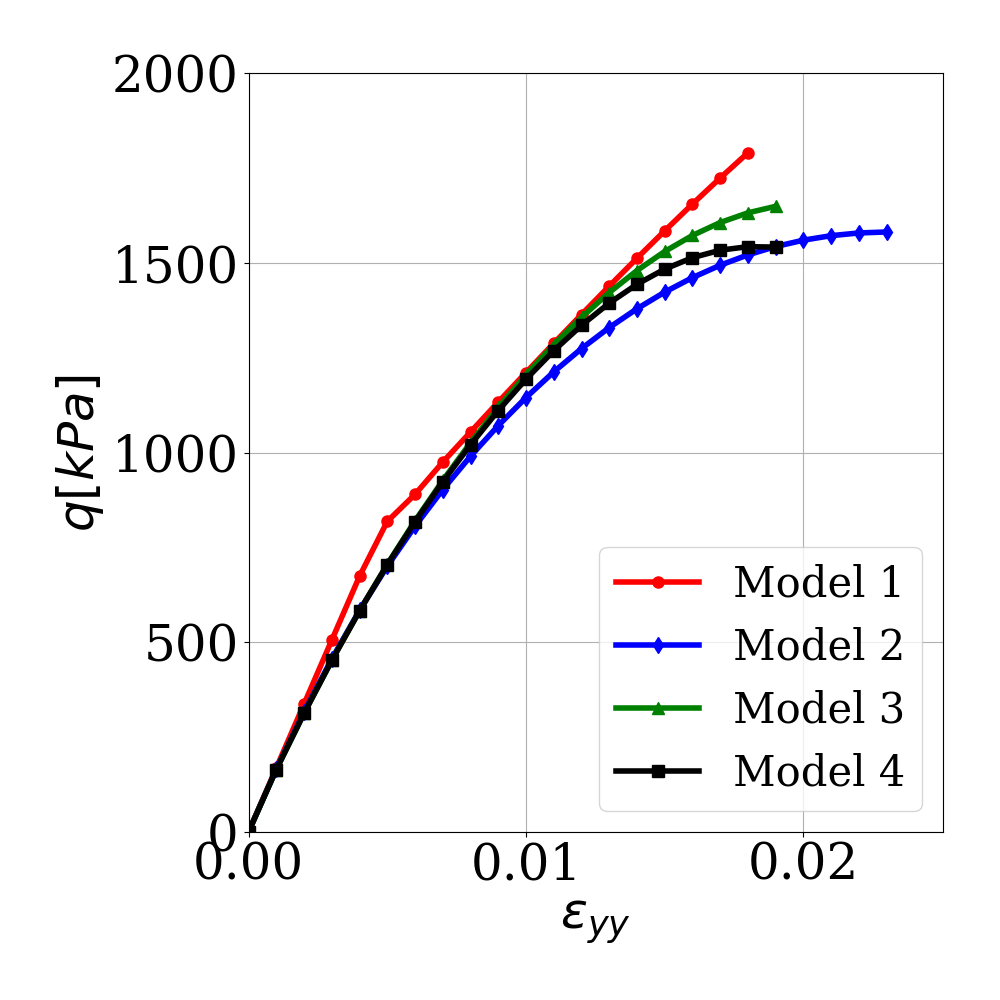}
	}
	\subfigure[Global volumetric strain $\epsilon_v$]{
		\includegraphics[width=0.45\textwidth]{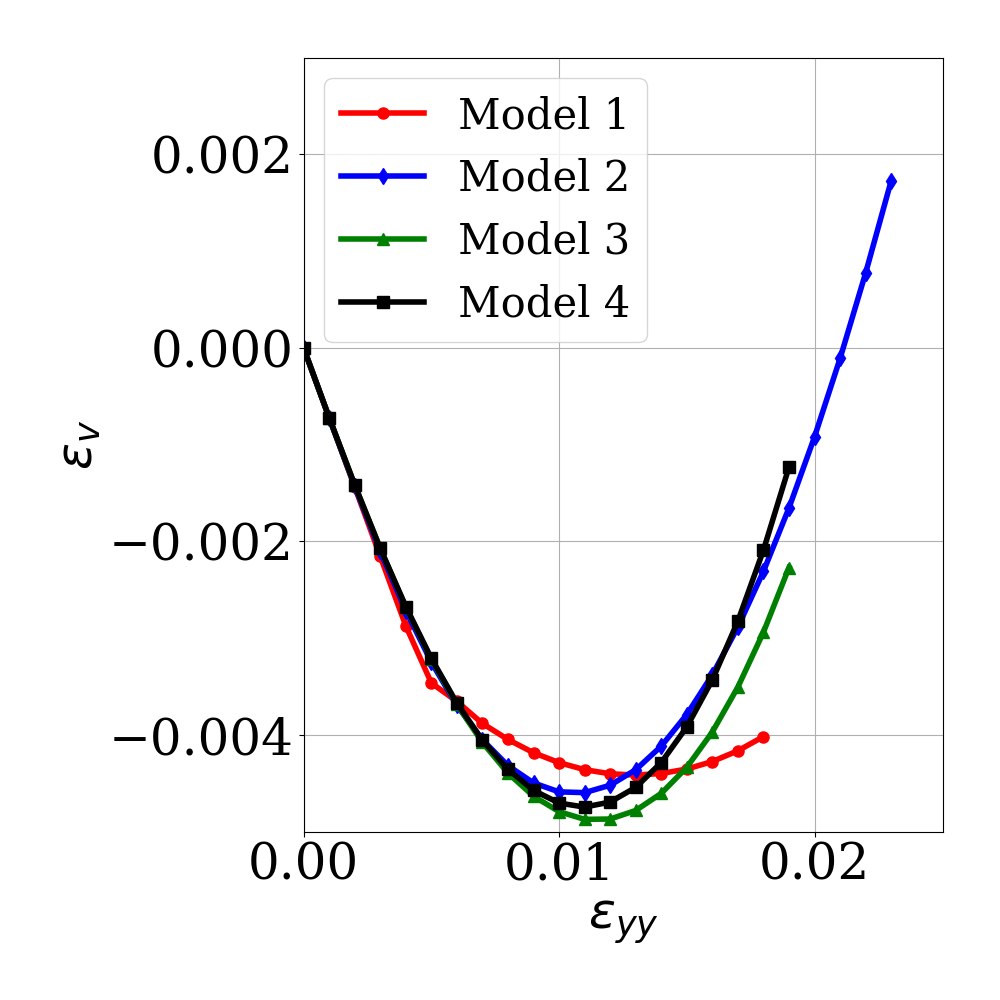}
	}
	\caption{Evolution of the global differential stress ($q = |\sigma_{yy} - p_0|$, $p_0 = -400 kPa$, and $\sigma_{yy}$ is the stress on the top surface) and the global volumetric strain $\epsilon_v$ of the specimen with respect to the axial strain $\epsilon_{yy}$ computed using three models generated during the deep reinforcement learning of the meta-modeling game (Model 1 is generated in the 1st DRL iteration in Figure \ref{fig:Test2_Improve}, Model 2 in the 5th iteration,  Model 3 in the 8th iteration, Model 4 in the 10th iteration). Each simulation terminates when the finite element solutions start to diverge.}
	\label{fig:example3_curve}
\end{figure}

The finite element implementation of the AI generated digraph-based model is simple and convenient. 
All modeling choices (Section \ref{sec:metamodeling}) and the general purpose integration scheme (Algorithm \ref{mcts_algorithm}) are already implemented in a single material model class. 
The FEM program is free to switch to other models simply by loading the digraphs and the corresponding calibrated parameters from the gameplay into this material class. 
The local distribution of the deviatoric strain $\epsilon_s$ and the volumetric strain $\epsilon_v$ in the specimen from the three models are compared in Figure \ref{fig:example3_eps_s} and Figure \ref{fig:example3_eps_v}, respectively. 
The global differential stress - axial strain and volumetric strain - axial strain curves are compared in Figure \ref{fig:example3_curve}.  
The results demonstrate that all the local constitutive models, regardless of the quality, can all be implemented in the finite element solver. As mentioned previous, this meta-modeling game can be easily incorporated in a new finite element solver architecture in which material library commonly used in the current paradigm is replaced by one single labeled directed multigraph and the conventional material identification process is replaced by the meta-modeling game such that both the optimal \textbf{combination of model components and material parameters} are simultaneously selected. 
Furthermore, the results also indicate that the qualities of the constitutive laws are continuously improved in each iteration of the meta-modeling game. In particular, we see that the correct type of shear band for dense granular assembles (dilatant shear band) is reproduced in the numerical specimens after 5th iterations (cf. \citet{aydin2006geological, sun2013unified}), and the shear band mode converges in the 8th iteration. 

\section{Conclusion and future Perspectives}
We introduce a new multi-agent meta-modeling game 
in which the experimental task, i.e. the generation of data,  and the modeling task, 
the interpretation of data, are handled by two artificial intelligence agents. 
Mincing the collaboration of a pair of experimentalist and modeler collaborating to derive, implement, calibrate 
and validate a model to explain a path-dependent process, 
these two agents interact with each other sequentially and exchange information until either 
the model and data they reach the objectives or when further action does not generate a further reward. 
The major contribution of this research is as follows. 
First, we introduce the idea of using labeled directed multi-graph to mathematically represent the action space 
of the modeling agent. This action space can be expanded by adding plausible actions invented by previous 
human modelers or by generated new actions from deep neural networks or other machine learning methods. 
This invention therefore enables 
us to idealize the process of writing constitutive models as a continuous decision-making process in an action space of very high dimensions such that a pre-defined objective function can be maximized. 
As shown previously in work such as AlphaGo \citet{silver2017masteringb, silver2017masteringchess}, 
using deep neural networks in a deep reinforcement learning framework to search proper actions 
from a very large number of possible moves  is shown to achieve superior performance. 
To the best knowledge of the authors, this is the first time the ideas of using deep reinforcement learning
applied on generating the knowledge graph and constitutive laws for history-dependent responses of materials. 

The introductions of the graph, directed graph and labeled directed multigraph in the meta-modeling game 
enables us to derive a meta-modeling game more closely resemble a more human-like iterative cyclical scientific process through which information is continuously gathered, hypotheses are continuously tested and the plausible understanding  is continually revised. 
The major elements of scientific methods used by human, including  
characterization (observation and measurement stored in vertices, definition stored in edges), 
hypotheses (selection of a particular form of edges and edge sets), predictions (the information flow from 
root to leave of the directed graph obtained from the meta-modeling game) and experiments are all 
incorporated and automated. 
This new approach produces a forecast engine that can make predictions, but more importantly has the ability to generate human-interpretable knowledge on the relationships amount different measurable physical quantities. 
This feature is significantly unique among other neural network approaches which often produce black-box models 
with no easy way to interpret the rationale of the predictions. It should be pointed out that 
models generated from the meta-modeling game do not discriminate the types of the edges used. They 
can be any operator that links the input to output, including but not limited to regression, support vector machine, neural network, mathematical expression or a bootstrapped version of them. These edges are only being formed by the AI
when they are estimated to have higher policy value according to a specific objective function. 

By introducing a gateway to merge existing and new models and introduce  a
 seamless integration of data generation and data-driven discovery. 
Since the meta-modeling game stops when further action does not yield reward, this framework enables 
one to determine the best \textbf{configuration} of model one can possibly obtain within a well-defined action space 
for a given set of data. As shown in Section \ref{sec:ne2}, this ability is not only important in 
making better predictions, but also help us identify the limitation of the action space. Given that modern
constitutive laws have become increasingly complex and are often combined products of multiple material theories, concepts and assumptions created by different schools or theoretical backgrounds, the quantifiable policy values of 
the edges in the edge label set, if used properly, may
enable us to pin down the relative values of each \textbf{component} of the 
constitutive laws while avoiding any potential implicit bias. 
As a result, the cooperative game enables us to make forward predictions while controlling the cost of generating the experimental data. Unlike other AI field which is largely driven by the exponential growth of available data, extracting an adequate amount of reliable experimental data remains a challenging task for the field of mechanics. The cooperative game designed in this paper does not only provide a tool to optimize the collaborations of the AI agents, but also shed lights on how to make productive scientific discovery through emulating the research progress in a setting where data generation can be costly.  

In this work, we assume that the data obtained from experiments are perfect and without any significant noise. 
Furthermore, the meta-modeling game is also operated in a setting where the vertex set and the corresponding label 
are fixed. Future work will consider how to introduce quantifiable assurance of the meta-modeling game, incorporate 
sensitivity analysis in the validation and predictions, and quantify different types of uncertainties. 
For instance, one trains Bayesian neural network to generate edges that deliver not only deterministic predictions
but also perform variational inference. 
By quantifying the sensitivity of the predictions, one may explore the weakness of the existing action space for 
both the modeler and experimentalist agents and use this knowledge to generate new actions. 
Work in this area is currently in progress.


\section{Acknowledgments}
The work of KW and WCS is supported by the Earth Materials and Processes
program from the US Army Research Office under grant contract 
W911NF-18-2-0306,
the Dynamic Materials and Interactions Program 
from the Air Force Office of Scientific Research under 
grant contract FA9550-17-1-0169, the nuclear energy university program 
from Department of Energy under grant contract 
DE-NE0008534, the Mechanics of Material program
at National Science Foundation under grant 
contract CMMI-1462760, and the Columbia SEAS Interdisciplinary Research Seed Grant.  
The work of QD is supported in part by
NSF  CCF-1704833, DMS-1719699, DMR-1534910, 
and  ARO  MURI  W911NF-15-1-0562. 
 These supports are gratefully 
acknowledged. 
The views and conclusions contained in this document are those of the authors, 
and should not be interpreted as representing the official policies, either expressed or implied, 
of the sponsors, including the Army Research Laboratory or the U.S. Government. 
The U.S. Government is authorized to reproduce and distribute reprints for 
Government purposes notwithstanding any copyright notation herein.

\bibliographystyle{plainnat}
\bibliography{main}

\end{document}